\documentclass[pdflatex,sn-apa]{sn-jnl}

\usepackage{graphicx}%
\usepackage{multirow}%
\usepackage{amsmath,amssymb,amsfonts}%
\usepackage{amsthm}%
\usepackage{mathrsfs}%
\usepackage[title]{appendix}%
\usepackage[table, xcdraw, dvipsnames]{xcolor}%
\usepackage{textcomp}%
\usepackage{manyfoot}%
\usepackage{booktabs}%
\usepackage{listings}%

 \usepackage{anyfontsize}
\usepackage{amsmath}
\usepackage{amsfonts}
\usepackage{bbm}
\usepackage[ruled]{algorithm2e}
\usepackage{algpseudocode}
\usepackage{hyperref}
\usepackage{stfloats}
\usepackage{subfig}

\usepackage{booktabs}\let\cline\cmidrule
\usepackage{tablefootnote}
\usepackage{threeparttable}

\usepackage{mathtools}
\usepackage{adjustbox}
\usepackage{tabularray}
\usepackage{pifont}
\usepackage[misc]{ifsym}
\usepackage{hhline}
\usepackage{makecell}

\theoremstyle{thmstyleone}%
\theoremstyle{thmstyletwo}%

\theoremstyle{thmstylethree}%

\newcommand{\boldrm}[1]{\textbf{\textrm{#1}}}

\definecolor{bluegray}{RGB}{221,228,247}
\definecolor{orggray}{RGB}{255, 250, 218}
\definecolor{impblue}{RGB}{68,114,196}
\newcommand{\improve}[1]{\textbf{\textcolor{MidnightBlue}{+#1}}}
\newcommand{\decrease}[1]{\textbf{\textcolor{Orange}{--#1}}}

\newcommand{\eg}{e.g.}
\newcommand{\ie}{i.e.}

\DeclareMathOperator*{\argmaxA}{arg\,max}
\newcommand{\cmark}{\ding{51}}%
\newcommand{\xmark}{\ding{55}}%
\newcommand\Tstrut{\rule{0pt}{2.2ex}}         
\newcommand\Bstrut{\rule[-0.5ex]{0pt}{0pt}}   
\newlength\replength
\newcommand\repfrac{.33}

\newcommand\tdashfill[1][\repfrac]{\cleaders\hbox to \replength{%
  \smash{\rule[\arraystretch\ht\strutbox]{\repfrac\replength}{\rulewidth}}}\hfill}
\newcommand\tabdashline{%
  \makebox[0pt][r]{\makebox[\tabcolsep]{\tdashfill\hfil}}\tdashfill\hfil%
  \makebox[0pt][l]{\makebox[\tabcolsep]{\tdashfill\hfil}}%
  \\[-\arraystretch\dimexpr\ht\strutbox+\dp\strutbox\relax]%
}
\newcommand\tdotfill[1][\repfrac]{\cleaders\hbox to \replength{%
  \smash{\raisebox{\arraystretch\dimexpr\ht\strutbox-.1ex\relax}{.}}}\hfill}

\newcommand{\repeatdashline}[1]{%
    \ifnum#1>0
        \tabdashline & \repeatdashline{\numexpr#1-1\relax}
    \fi
}

\begin{document}

\title[Interactive Test-Time Adaptation with Reliable Spatial-Temporal Voxels for Multi-Modal Segmentation]{Interactive Test-Time Adaptation with Reliable Spatial-Temporal Voxels for Multi-Modal Segmentation}


\author[1]{\fnm{Haozhi} \sur{Cao}}\email{haozhi002@ntu.edu.sg}

\author[2]{\fnm{Yuecong} \sur{Xu}}\email{yc.xu@nus.edu.sg}

\author[1]{\fnm{Pengyu} \sur{Yin}}\email{pengyu001@ntu.edu.sg}

\author[1]{\fnm{Xingyu} \sur{Ji}}\email{xingyu001@ntu.edu.sg}

\author[1]{\fnm{Shenghai} \sur{Yuan}}\email{shyuan@ntu.edu.sg}

\author[3,4]{\fnm{Jianfei} \sur{Yang}}\email{jianfei.yang@ntu.edu.sg}

\author[1]{\fnm{Lihua} \sur{Xie}}\email{elhxie@ntu.edu.sg}

\affil[1]{\orgdiv{Centre for Advanced Robotics Technology Innovation (CARTIN)}, \orgname{Nanyang Technological University}}

\affil[2]{\orgdiv{Department of Electrical and Computer Engineering}, \orgname{National University of Singapore}}

\affil[3]{\orgdiv{School of Electrical and Electronic Engineering}, \orgname{Nanyang Technological University}}

\affil[4]{\orgdiv{School of Mechanical and Aerospace Engineering}, \orgname{Nanyang Technological University}}


\abstract{Multi-modal test-time adaptation (MM-TTA) adapts models to an unlabeled target domain by leveraging the complementary multi-modal inputs in an online manner. While previous MM-TTA methods for 3D segmentation offer a promising solution by leveraging self-refinement per frame, they suffer from two major limitations: 1) unstable frame-wise predictions caused by temporal inconsistency, and 2) consistently incorrect predictions that violate the assumption of reliable modality guidance. To address these limitations, this work introduces a comprehensive two-fold framework. Firstly, building upon our previous work Re\textbf{L}iable Sp\textbf{at}ial-\textbf{t}emporal Vox\textbf{e}ls (Latte), we propose Latte++ that better suppresses the unstable frame-wise predictions with more informative geometric correspondences. Instead of utilizing a universal sliding window, Latte++ employs multi-window aggregation to capture more reliable correspondences to better evaluate the local prediction consistency of different semantic categories. Secondly, to tackle the consistently incorrect predictions, we propose Interactive Test-Time Adaptation (ITTA), a flexible add-on to empower effortless human feedback with existing MM-TTA methods. ITTA introduces a novel human-in-the-loop approach that efficiently integrates minimal human feedback through interactive segmentation, requiring only simple point clicks and bounding box annotations on one object in $1\%$ of frames. Instead of using independent interactive networks, ITTA employs a lightweight promptable branch with a momentum gradient module to capture and reuse knowledge from scarce human feedback during online inference. Extensive experiments across five MM-TTA benchmarks demonstrate that ITTA achieves consistent and notable improvements with robust performance gains for target classes of interest in challenging imbalanced scenarios, while Latte++ provides complementary benefits for temporal stability. Code will be available at \url{https://github.com/AronCao49/Latte-plusplus}.}




\maketitle

\section{Introduction}
\label{sec:intro}
3D semantic segmentation~\citep{hong2024unified,ding2024lowis3d} plays an essential role in various autonomous applications, such as autonomous driving and robotic navigation~\citep{guo2020deep, feng2020deep, deng2024plgslam}. The increasing demand for robust sensing has led to the widespread adoption of multi-modal sensors (\eg, cameras and LiDARs) in autonomous systems, where complementary information from different modalities enhances the overall perception capabilities. However, most existing advances in multi-modal 3D semantic segmentation rely on the expensive fully supervised learning, which requires laborious point-wise annotations and therefore suffers from poor generalizability when facing domain shifts~\citep{qi2017pointnet,choy20194d,tang2020searching}. 

\begin{figure}[t]
    \centering
    \includegraphics[width=\textwidth]{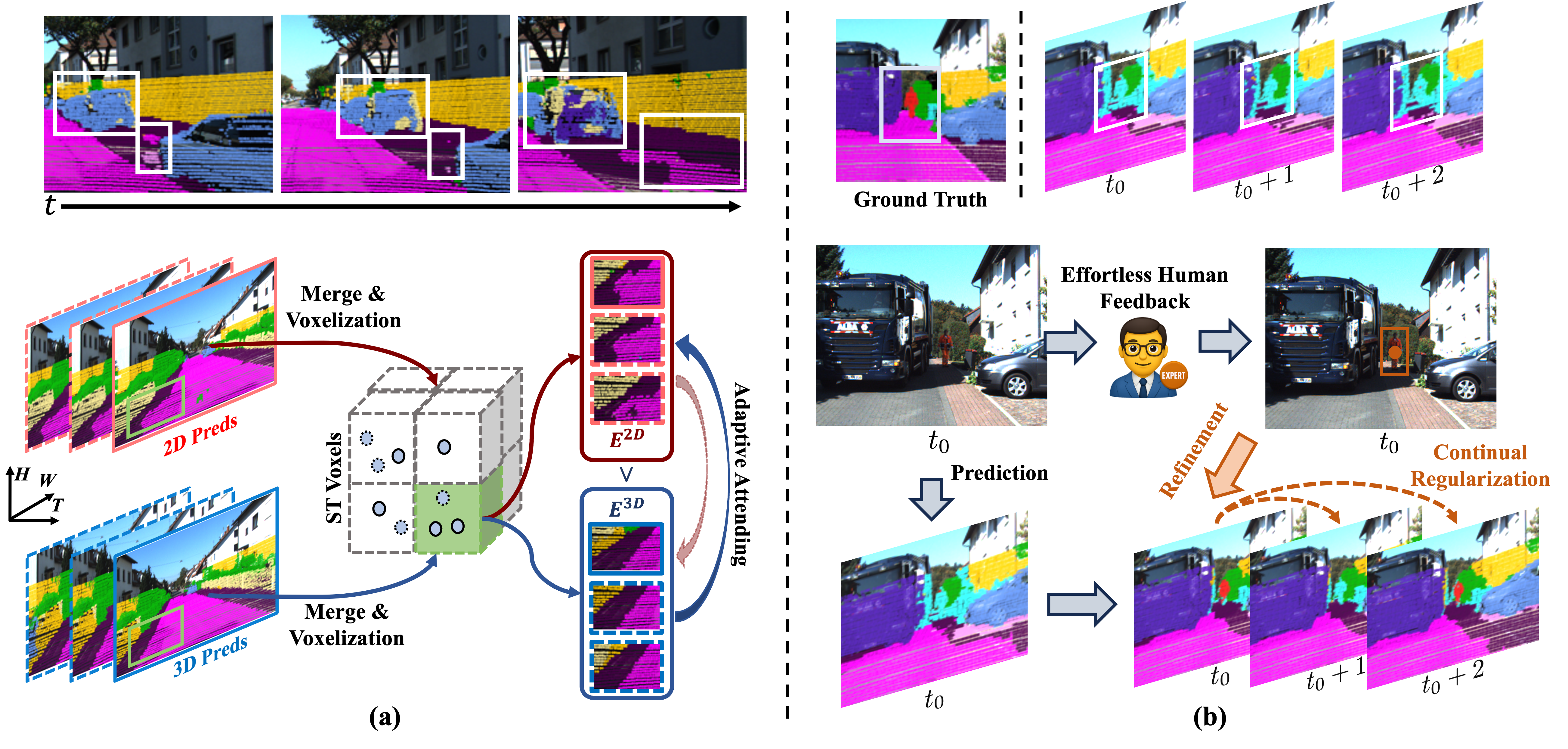}
    \caption{Illustration of two types of challenging noisy prediction and our proposed methods. (a) demonstrates the unstable frame-wise predictions between consecutive frames from the state-of-the-art MM-TTA method~\citep{shin2022mm} highlighted in white boxes. To suppress this frame-wise instability, our Latte and Latte++ achieve reliable cross-modal attending by estimating Spatial-Temporal (ST) entropy (\ie, \color{red}$E^{\mathrm{2D}}$ \color{black}and \color{blue}$E^{\mathrm{3D}}$\color{black}) within each ST voxel. In terms of (b) consistently incorrect predictions (\eg, the consistently misclassified pedestrian in white boxes), we propose Interactive Test-Time Adaptation (ITTA) by leveraging effortless human feedback for instant refinement and continual regularization.}
    \label{fig:Intro}
\end{figure}

To overcome this limitation, recent efforts have been devoted to Multi-Modal Unsupervised Domain Adaptation (MM-UDA)~\citep{jaritz2022cross,peng2021sparse,li2022cross, xing2023cross} that leverages multi-modal information to transfer knowledge from the labeled source domain to the unlabeled target domain. Despite its effectiveness, MM-UDA requires full access to the source domain dataset and multiple training epochs in an offline manner, which is infeasible when tackling distribution shifts observed during online inference. Motivated by the recent progress of Test-Time Adaptation (TTA)~\citep{niu2022efficient, niu2023towards} for single-modal input, the first Multi-Modal Test-Time Adaptation (MM-TTA) method has been proposed for 3D segmentation~\citep{shin2022mm}. MM-TTA inherits the fundamental premise of TTA, which utilizes a pre-trained model and prohibits direct access to samples from the source domain. As MM-TTA is formulated as a \textit{quick adaptation} scenario (\ie, one epoch for training only as in~\citep{shin2022mm, su2023revisiting}), it necessitates a reliable source of supervision signals to ensure stable optimization in an online manner. 

Despite existing efforts in prediction refinement and modal-wise reliability estimation~\citep{shin2022mm}, previous methods focus on frame-wise optimization, which falls short when addressing unstable single-frame predictions caused by the non-trivial domain shift between the source and target domains. As illustrated in Fig.~\ref{fig:Intro}(a), the single-frame refinement strategy~\citep{shin2022mm} suffers from noisy single-frame predictions on the car in the white rectangle across time, degrading the segmentation performance and aggravating the denoising burden in downstream tasks (\eg, semantic-based retrieval~\citep{yin2023outram} and obstacle recognition). Worse still, these temporally unstable single-frame predictions can be wrongly regarded as reliable, propagating noise to the other modality and causing error accumulation and catastrophic forgetting~\citep{niu2022efficient, niu2023towards}. While previous methods~\citep{wang2022continual, Cao_2023_ICCV} alleviate this instability by utilizing the average predictions of multiple augmented frames, they are computationally expensive for online adaptation since the inference time grows linearly with the increasing number of augmented frames.

Beyond unstable frame-wise predictions, consistently incorrect predictions across frames pose a more significant challenge. Existing MM-TTA methods, mostly based on modality-attention or prediction-refinement mechanisms~\citep{shin2022mm,Cao_2023_ICCV,cao2025reliable}, rely on the foundation premise of previous MM-TTA methods: at least one modality provides noiseless guidance to construct reliable self-supervision signals. Unfortunately, this assumption frequently fails in practice, particularly for imbalanced classes and out-of-distribution samples where the online predictions are much noisier. This results in \textbf{consistently incorrect predictions} (\eg, incorrect predictions of pedestrian as in Fig.~\ref{fig:Intro}(b)) that models can not effectively rectify by themselves, causing more severe error accumulation compared to other semantic classes and deteriorating the overall imbalanced performance. The severity of this challenge has motivated the exploration of incorporating human feedback in prediction rectification, while existing methods either target image classification rather than segmentation~\citep{gui2024active,wang2025effortless} or require tedious pixel-wise human correction~\citep{hu2024towards}, making them impractical for online 3D segmentation that requires real-time performance with minimal human intervention.

Building upon these premises, this work aims to improve the existing MM-TTA methods when tackling the two aforementioned types of noisy predictions: (1) unstable frame-wise predictions, and (2) consistently incorrect predictions. Firstly, to tackle noisy predictions of type (1), we propose \textbf{Latte++} which extends our previous work Re\textbf{L}iable Sp\textbf{at}ial-\textbf{t}emporal Vox\textbf{e}ls (Latte)~\citep{cao2025reliable}, to better evaluate the spatial-temporal prediction consistency of different semantic categories. Illustrated as in Fig.~\ref{fig:Intro}(a), the baseline Latte mitigates the unstable frame-wise predictions by estimating the modal-wise prediction consistency within each Spatial-Temporal (ST) Voxel extracted using a pre-defined sliding aggregation window. Considering the fact that prediction consistency of different semantic categories can be better evaluated under different window sizes (\eg, dynamic objects are better evaluated in smaller windows to reduce the shadow-like effect~\citep{chen2019suma++} while static backgrounds could benefit from a larger windows to introduce more spatial-temporal correspondences), Latte++ enhances the consistency evaluation by leveraging various aggregation windows to achieve a better reliability estimation. 

Secondly, to tackle the challenging type (2) predictions, we propose Interactive Test-Time Adaptation (ITTA) for multi-modal 3D semantic segmentation, which efficiently integrates human feedback online in the form of interactive segmentation~\citep{kirillov2023segment,huang2023interformer} as shown in Fig.~\ref{fig:Intro}(b). Leveraging the paradigm of interactive segmentation, ITTA acts as a flexible and efficient add-on that captures effortless human prompts (\eg, points and boxes that simple clicks and draws can effortlessly generate) for online rectification. Instead of utilizing an independent interactive segmentation network, we introduce and warm up a lightweight and promptable branch on the 2D network as a human prompt interface. A momentum gradient module is further introduced to capture and reuse knowledge from occasionally available human feedback in a continuous manner. The proposed ITTA is a flexible add-on that can be combined with existing MM-TTA methods, and we demonstrate that it achieves significant improvements with limited human prompts (points/boxes of one object in $1\%$ of frames) across different benchmarks and different MM-TTA methods.

Our validation is extensively conducted across five different benchmarks under two different MM-TTA schemes (\ie, a stationary target domain~\citep{shin2022mm}, namely MM-TTA, or a \textit{continually} changing target domain, namely MM-\textit{C}TTA~\citep{Cao_2023_ICCV}), which demonstrates the superiority of Latte++ compared to existing MM-TTA methods and the significant improvement brought by ITTA. To sum up, our contributions are listed as follows:
\begin{itemize}
    \item For MM-TTA without human feedback, we extend our Latte to Latte++ by using various sliding windows to achieve better class-wise prediction accuracy.
    \item We design ITTA, a flexible MM-TTA add-on that can effectively capture effortless human prompts as feedback to rectify the online predictions. Momentum gradient is additionally proposed to preserve and reuse the human knowledge captured as a continual regularization.
    \item Extensive experiments are conducted on ITTA with Latte++ and three other MM-TTA methods across five different benchmarks, which demonstrate its effectiveness on rectifying online imbalanced performance.
\end{itemize}

\section{Related Works}
\textbf{Test-Time Adaptation (TTA)}. 
TTA is proposed to mitigate the domain shift between the inaccessible training source domain and the testing target domain. Different from source-free adaptation~\citep{liang2020we, simons2023summit}, which also discards access to the source domain yet requires multiple training epochs, the adaptation process of TTA is completely online, following the ``one-pass'' protocol as in~\citep{su2023revisiting}. Existing TTA methods can be categorized into two types based on their adaptation strategies~\citep{liang2024comprehensive}, including offline TTA methods (\eg, source-free domain adaptation~\citep{kundu2020universal,liang2021source,chen2022contrastive, zheng2024360sfuda++} and test-time training) and online TTA methods\footnote{As our discussion mainly lies in online adaptation, ``TTA'' refers to online TTA methods in this work for simplicity}. Due to its practical and challenging setting, TTA is attracting more and more attention. Previously proposed TTA methods, which mainly target image classification, can be divided into two categories: i) backpropagation(BP)-free TTA, and ii) BP-based TTA. BP-free TTA methods consider a challenging scenario where computational resources are limited, which therefore restricts updating learnable parameters through backpropagation. They therefore propose various solutions without updating most of the network parameters, such as batch normalization adaptation~\citep{gong2022note,zhao2023delta}, output logit adaptation~\citep{boudiaf2022parameter}, and prompt adaptation using forward-pass only~\citep{niu2024test}. Despite its efficiency, the prohibition of updating parameters usually limits the adaptation capacity of BP-free TTA methods, leading to inferior performance compared to BP-based TTA methods.

BP-based TTA methods, on the other hand, attempt to address this task by proposing unsupervised online objectives from different perspectives, such as entropy minimization~\citep{wang2020tent, niu2022efficient, niu2023towards}, self-training with pseudo-labels~\citep{goyal2022test,wang2022towards}, and augmentation invariance~\citep{wang2022continual, zhang2022memo}. In addition to regularizing predictions, some methods propose to adapt networks from the feature level instead~\citep{liu2021ttt++, su2023revisiting}, while more recent methods begin to consider different variants of TTA scenarios, such as continual TTA~\citep{wang2022continual, niu2022efficient, Cao_2023_ICCV}, non-i.i.d TTA~\citep{gong2022robust}, or the mix of aforementioned cases~\citep{yuan2023robust}. While both existing BP-free and BP-based TTA methods apply to multi-modal 3D semantic segmentation, they omit the existing spatial-temporal correlations within consecutive scans and therefore are incapable of addressing inconsistency issues as in Fig.~\ref{fig:Intro}(a). In this work, we argue that temporal information in 3D segmentation can be effectively leveraged for TTA, as demonstrated by Latte in this work.

\textbf{Multi-modal domain adaptation for 3D segmentation}. 
To avoid expensive annotation costs and overcome the poor generalizability of fully supervised solutions, various multi-modal domain adaptation methods have been investigated. Existing methods can be roughly divided into three types based on their adaptation settings: (i) MM-UDA, which requires full access to both source and target domains and offline training, (ii) Multi-Modal Source-Free Domain Adaptation (MM-SFDA), which relies on offline training with the target domain only, and (iii) MM-TTA, which is performed online on the target domain during the inference process. For MM-UDA, xMUDA~\citep{jaritz2022cross} is the primary work that incorporates cross-modal learning with MM-UDA. It regards cross-modal prediction consistency and pseudo-labels as its supervision signals in the unlabeled target domain. Most subsequent works propose solutions to address its limitations from different perspectives, such as more diverse point-pixel correspondence~\citep{peng2021sparse, xing2023cross}, procedures to mitigate domain gaps~\citep{li2022cross, liu2021adversarial}, and alleviating the class-imbalanced problem~\citep{cao2023mopa}. Inspired by the recent success of the Vision Foundation Model (VFM)~\citep{kirillov2023segment,awais2025foundation} and Language-Image Pre-Training (CLIP)~\citep{radford2021learning}, some recent MM-UDA methods~\citep{cao2023mopa,wu2024clip2uda,wuunidseg,peng2025learning} attempt to leverage VFM or CLIP as the domain invariant feature to achieve better transfer of usable knowledge to the target domains. Specifically, MoPA~\citep{cao2023mopa} proposes to encourage consistent predictions within the filtered unlabeled semantic masks provided by SAM~\citep{kirillov2023segment}, while the following works~\citep{wuunidseg,wu2024clip2uda,peng2025learning} further extend these knowledge transferring techniques from a feature-level perspective. Specifically, UniDSeg~\citep{wuunidseg} introduces additional lightweight learnable modules between layers of VFM to facilitate the transitional prompting between modalities, whereas a simpler alternative~\citep{peng2025learning} is proposed to align the 3D features to the ones from VFM directly. CLIP2UDA~\citep{wu2024clip2uda} leverages the pre-aligned textual embedding from CLIP as the middle ground for multi-modal feature alignment.

Despite the effectiveness of MM-UDA methods, they inevitably require full access to the source domain, which could be infeasible due to different factors, such as privacy or license concerns. To overcome this limitation, some primary works propose MM-TTA and MM-SFDA, which perform adaptation using backbones pre-trained on the source domain only. Specifically, SUMMIT~\citep{simons2023summit} proposes the first MM-SFDA method by estimating cross-modal prediction agreement, yet it requires multiple offline training epochs. MM-TTA~\citep{shin2022mm} is a more flexible and efficient pipeline, which conducts online adaptation as the existing online TTA methods. MMTTA~\citep{shin2022mm} proposes the first MM-TTA method, which adaptively attends to the modality with less prediction entropy in a point-wise manner. However, all existing MM-UDA, MM-SFDA, and MM-TTA methods focus on the refinement of single-scan prediction while neglecting the correlations between geometric nearest neighbors. In this work, we propose Latte/Latte++ to leverage these informative correspondences that naturally exist for better modal-wise reliability estimation and more effective cross-modal attending. 

\textbf{Temporal processing of point clouds}. 
Since point clouds are naturally in the form of temporally consecutive frames, previous methods have widely explored how to leverage such temporal information for different tasks. Specifically, scene flow~\citep{vedula1999three, vogel20113d, vogel2013piecewise, huang2022dynamic} contains the movement of every point in the 3D world. Considering that point-wise computation is expensive, previous works on different tasks (\eg, segmentation~\citep{choy20194d,saltori2022gipso,chen2023clip2scene} and object detection~\citep{piergiovanni20214d}) usually interpret the temporal information from a simpler perspective. In terms of fully supervised learning, temporal information is primarily leveraged by 4D segmentation method~\citep{choy20194d} with either generalized sparse convolution kernels~\citep{graham2015sparse} or fully connected MLPs~\citep{fan2022pstnet}, while the others~\citep{xu2022int} extract long-term temporal information by an efficient memory bank. On the other hand, considering scenarios with insufficient or zero annotations, some works propose different strategies to couple spatial-temporal information without ground-truth labels. For instance, \citep{chen2023clip2scene} proposes to guide voxelized spatial-temporal representation with strong text embedding from CLIP~\citep{radford2021learning} instead of labels in a multi-modal manner. In contrast, \citep{saltori2022gipso} interprets such representation by nearest neighbor search in a frame-to-frame manner. In this work, we design sliding-window aggregation and voxelization rather than aggregating all frames as in~\citep{chen2023clip2scene}, which better indicates the point-wise prediction reliability. 

\section{Methodology}
\noindent
\textbf{Problem definition}. During MM-TTA for 3D segmentation in the target domain $\mathcal{T}$, both 2D RGB images and 3D point clouds are consecutively observed, represented as $\boldrm{x}^{\textrm{2D}}_{\mathcal{T},t} \in \mathbb{R}^{3\times H\times W}$ and $\boldrm{x}^{\textrm{3D}}_{\mathcal{T},t} \in \mathbb{R}^{N\times 4}$, respectively, where $t$ indicates the order of frames and $H, W$ denote the height and width of images. The input of each 3D point consists of its 3D coordinate $\{x, y, z\}$ and feature, such as intensity and reflectivity. Both 2D and 3D networks are pre-trained on the labeled source domain prior to MM-TTA, denoted as $\phi_{\mathcal{T},t}^{\textrm{m}}(\cdot)$ where $\textrm{m} \in \{\textrm{2D}, \textrm{3D}\}$. During MM-TTA, both networks are initialized from the parameters pre-trained on the source domain. We follow previous works~\citep{jaritz2022cross,shin2022mm} to obtain point-pixel correspondences by projective 3D points to the 2D image based on the relative projection matrix for each frame, which leads to the modal-wise predictions $\boldrm{p}^\textrm{m}_{\mathcal{T},t} = \phi_{\mathcal{T},t}^{\textrm{m}}(\boldrm{x}^{\textrm{m}}_{\mathcal{T},t}), \boldrm{p}^\textrm{m}_{\mathcal{T},t} \in \mathbb{R}^{N\times K}$ where $K$ denotes the number of semantic classes. Accessing raw data from the labeled source domain is strictly prohibited during MM-TTA, and the subscript $\mathcal{T}$ indicating the target domain is omitted by default.

We propose a two-pronged approach targeting both types of prediction instability in MM-TTA. To alleviate the unstable frame-wise predictions, we first present Latte++, an enhanced version of our previous work Latte~\citep{cao2025reliable} that incorporates more diverse aggregation strategies for different semantic categories. We then introduce Interactive Test-Time Adaptation (ITTA), which addresses consistently incorrect predictions by incorporating human feedback during online adaptation.

\begin{figure*}[t]
    \centering
    \includegraphics[width=\textwidth]{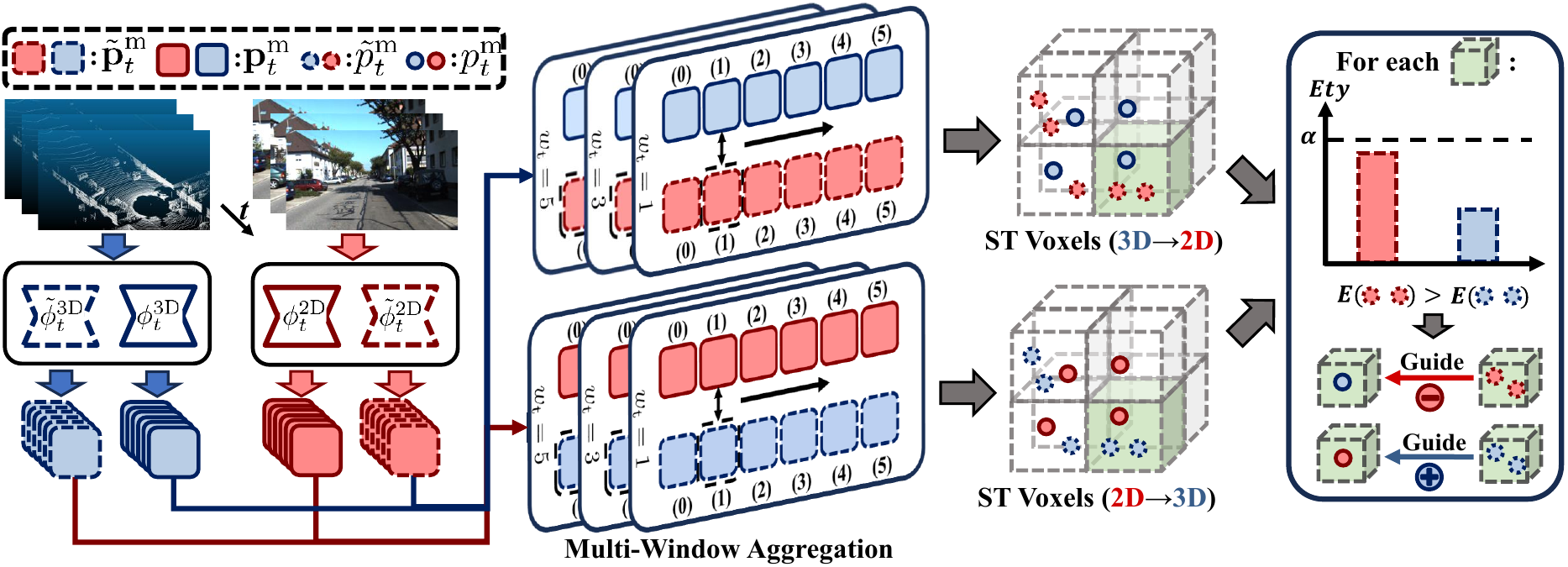}
    \caption{Overall structure of Latte++. Taking a student prediction frame of one modality as the query input, our sliding-window aggregation searches its spatial-temporal correspondences through voxelization within a time window to establish the temporally local prediction consistency. ST voxels are then generated, where those with high ST entropy (larger than $\alpha$-quantile) are discarded as unreliable correspondences, while the others are leveraged for adaptive cross-modal learning by attending to the modality with lower ST entropy in a voxel-wise manner.}
    \label{fig:Overall Method}
\end{figure*}

\subsection{Latte and Latte++}
In this section, we introduce the overall pipeline of Latte++ and its foundation, Latte. As shown in Fig.~\ref{fig:Overall Method}, given consecutive frames of images and point clouds as multi-modal input, both Latte and Latte++ first extract multi-modal frame-wise predictions utilizing student and teacher networks (Section~\ref{sec:Frame-wise Preds}). The predictions are then passed to the sliding-window aggregation and voxelization (Section~\ref{sec:Slide Window}). Latte subsequently computes ST voxels and ST entropy that indicate the prediction consistency and certainty (Section~\ref{sec:ST Voxels}), while Latte++ leverages different corresponding under various aggregated frames to generate more informative ST voxels and entropy for reliability estimation (Section~\ref{sec: Multi-Window}). The final ST entropy is leveraged for cross-modal attending to alleviate the noise from unreliable modality-specific predictions (Section~\ref{sec:Cross-Modal}).

\subsubsection{Frame-wise Predictions from Students and Teachers}\label{sec:Frame-wise Preds}
Identical to Latte~\citep{cao2025reliable}, Latte++ adopts slow teacher models and fast student models in a modal-wise manner, denoted as $\Tilde{\phi}^{\textrm{m}}_t(\cdot)$ and $\phi_t^{\textrm{m}}(\cdot)$, respectively. Given the input $\boldrm{x}^{\textrm{m}}_{t}$, the student and teacher predictions $\boldrm{p}^{\textrm{m}}_{t}$, $\Tilde{\boldrm{p}}^{\textrm{m}}_{t}$ are computed as:
\begin{equation}
    \Tilde{\boldrm{p}}^{\textrm{m}}_{t} = \Tilde{\phi}_t^{\textrm{m}}(\boldrm{x}^{\textrm{m}}_{t}), \ \
    \boldrm{p}^{\textrm{m}}_{t} = \phi_t^{\textrm{m}}(\boldrm{x}^{\textrm{m}}_{t}).\label{Eq: Student Teacher}
\end{equation}
During the online adaptation, teacher predictions serve as the source of generating supervision signals to directly optimize student models, while the parameters of teacher models are updated as the moving average of those from student models.

\subsubsection{Sliding-Window Aggregation and Voxelization}\label{sec:Slide Window}
The goal of Latte and Latte++ is to efficiently capture spatial-temporal correspondences to estimate the prediction reliability through consistency within correspondences. Establishing spatial-temporal correspondences in 3D space has been explored in other tasks, where they mainly consider temporally global relationships of all input frames~\citep{saltori2022gipso,chen2023clip2scene} or frame-to-frame correspondences~\citep{saltori2022gipso}.  
However, they exhibit limitations in terms of consistency evaluation: the former can not effectively emphasize the risk of temporally local inconsistency (\ie, inconsistent predictions within a short window), while the latter captures a relatively limited number of correspondences.

To effectively leverage spatial-temporal information, Latte~\citep{cao2025reliable} proposes sliding-window aggregation that utilizes a fixed and pre-defined time window to evaluate local consistency. 
Specifically, given a student prediction frame $\boldrm{p}^{\textrm{m}}_{i}$ with the temporal index $i$ as the query and the pre-defined window size $w_\mathrm{t}$, the correspondence search is conducted within a time window of consecutive frames denoted as $\{j \vert |j-i| \leq w_\mathrm{t}\}$. The merged point cloud $\hat{\boldrm{x}}_{i}^{\textrm{3D}} \in \mathbb{R}^{\hat{N}\times 3}$ (point features are omitted for simplicity) for correspondence search is then formulated as:
\begin{equation}
    \hat{\boldrm{x}}_{i}^{\textrm{3D}} = \textrm{cat}(\{\boldrm{T}_{j \rightarrow i}*\boldrm{x}_{j}^{\textrm{3D}} \vert |j-i| \leq w_\mathrm{t}\}),\ \ \boldrm{T}_{j \rightarrow i} = \boldrm{T}_i^{-1}\boldrm{T}_j,\label{Eq: Slide Window}
\end{equation}
where $\boldrm{T}_i$ is the estimated pose at frame $i$ and so as $\boldrm{T}_j$, which can be easily obtained online through off-the-shelf SLAM algorithms~\citep{vizzo2023kiss, ji2024lio}. $\textrm{cat}(\cdot)$ is the concatenation operation along the point number dimension while $*$ denotes the pose transformation. The voxelization is then performed on $\hat{\boldrm{x}}_{i}^{\textrm{3D}}$, formulated as:
\begin{align}
    \boldrm{v}_{i}^{\textrm{3D}} & = \mathcal{V}_\boldrm{s}(\hat{\boldrm{x}}_{i}^{\textrm{3D}}),\ \ \boldrm{v}_{i}^{\textrm{3D}} \in \mathbb{R}^{N_\textrm{v}\times 3}, \nonumber \\
    \boldrm{M}_{i}^{\textrm{3D}} & = \{k|\forall g \in [1, \hat{N}], \lfloor \hat{\boldrm{x}}_{i,g}^{\textrm{3D}} / \boldrm{s} \rfloor = \boldrm{v}_{i,k}^{\textrm{3D}} \}, \ \ \boldrm{M}_{i}^{\textrm{3D}}  \in \mathbb{R}^{\hat{N}},\label{Eq: Voxel Inverse}
\end{align}
where $\boldrm{v}_{i}^{\textrm{3D}}$ is the extracted voxels and $\mathcal{V}_\boldrm{s}$ is the voxelization operation with a voxel size $\boldrm{s}$. Points located in the same voxel are then regarded as correspondences. $\boldrm{M}_{i}^{\textrm{3D}}$ denotes that hash table that maps each voxel back to its containing points (\eg, $\boldrm{v}_{i,k}^{\textrm{3D}}$ to $\hat{\boldrm{x}}_{i,g}^{\textrm{3D}}$ as in Equation~(\ref{Eq: Voxel Inverse})), which is preserved for the following correspondence mapping. Compared to globally merging all frames in a single step, the sliding-window aggregation provides various correspondences through neighborhood selection strategies for consistency evaluation. In practice, both aggregation and voxelization are conducted iteratively for each frame in the input batch. 

\subsubsection{Spatial-Temporal Voxels and Entropy}\label{sec:ST Voxels}
Given the geometric correspondences captured by voxelization, our next step is to emphasize the predictions with higher correspondence consistency within each voxel. The aggregation and voxelization procedures inevitably involve some unreliable correspondences of two different types: (i) voxels that involve points of different semantic objects (\eg, those located on the edge shared by two objects) and (ii) voxels with highly inconsistent predictions due to uncertain predictions across time. To suppress the noisy contribution from such unreliable correspondences, we utilize ST voxels that leverage ST entropy for voxel-wise reliability evaluation and cross-modal attending. 

Specifically, an ST voxel in Latte consists of a query and a reference, where the query is encouraged to be consistent with references. Since the teacher models generate more stable predictions compared to the student models~\citep{wang2022continual,Cao_2023_ICCV}, we regard the single-frame student predictions as the query and multi-frame teacher predictions as our references, forming ST voxels denoted as $\boldrm{v}^{\textrm{ST}}_{i}$. Without losing generality, here we take a single ST voxel indexed by $k$ in frame $i$ as an example, denoted as  $\boldrm{v}^{\textrm{ST}}_{i,k}=\{ \boldrm{p}_\textrm{q}^\textrm{m}, \boldrm{p}^\mathrm{m}_{\mathrm{r}} \}$, where $\boldrm{p}_\textrm{q}^\textrm{m} \in \mathbb{R}^{N_\textrm{q} \times K}$ and $\boldrm{p}_{\textrm{r}}^\textrm{m} \in \mathbb{R}^{N_\textrm{r} \times K}$ are the point-wise student predictions at the query frame $i$ and teacher predictions in the frame searching range $j$ aggregated with sliding window $w_\mathrm{t}$, respectively, formulated as:
\allowdisplaybreaks
\begin{align}
    \boldrm{p}_\textrm{q}^\textrm{m} = &\{ {\boldrm{p}^{\textrm{m}}_{i, g}} \vert \lfloor \boldrm{x}_{i, g}^{\textrm{3D}} / \boldrm{s} \rfloor = \boldrm{v}_{i, k}^{\textrm{3D}}\},\nonumber \\
    \boldrm{p}_{\textrm{r}}^\textrm{m} = &\{ {\Tilde{\boldrm{p}}^{\textrm{m}}_{j, g}} \vert \lfloor \boldrm{x}_{j, g}^{\textrm{3D}} / \boldrm{s} \rfloor = \boldrm{v}_{i, k}^{\textrm{3D}}, |j-i| \leq w_\mathrm{t}\}.\label{Eq: ST Voxel Reference}
\end{align}

The reliability of each voxel is then evaluated as Shannon's entropy~\cite{wyner1974recent} of the average teacher predictions, where the unreliable ST voxels are then filtered out given a pre-defined quantile $\alpha$ as in Fig.~\ref{fig:Overall Method}, denoted as:
\allowdisplaybreaks
\begin{align}
    E^{\textrm{m}}_{i, k} &= -\sum_c^K \bar{\boldrm{p}}^{\textrm{m}}_{\textrm{r},c} \log \bar{\boldrm{p}}^{\textrm{m}}_{\textrm{r},c}, \ \ 
    \bar{\boldrm{p}}^{\textrm{m}}_\textrm{r} = \sum_{n=1}^{N_\textrm{r}} \psi(\boldrm{p}_{\textrm{r}, n}^\textrm{m}) / N_\textrm{r},\label{Eq: Average ST Preds}, \ \ h_{i, k} = \begin{cases}
                    0,\ \textrm{if} \ \ E^{m}_{i, k} > Q^m(\alpha)\\
                    1,\ \textrm{if} \ \ E^{m}_{i, k} \leq Q^m(\alpha)
                    \end{cases},
\end{align}
where $E^{\textrm{m}}_{i, k}$ is regarded as the ST entropy of voxel $k$ for modality $\textrm{m}$, indicating the modal-specific prediction reliability within this voxel. $Q^m(\alpha)$ is the $\alpha$-quantile of all ST entropy in modality $\textrm{m}$ in the same batch and $\psi(\cdot)$ denotes the Softmax function along the class dimension. 

\subsubsection{Latte++: Multi-Window Aggregation and Intra-Modal Attending}\label{sec: Multi-Window}
As illustrated in Section~\ref{sec:Slide Window} and Section~\ref{sec:ST Voxels}, the vanilla Latte evaluates the local prediction consistency within a pre-defined universal window size $w_\mathrm{t}$. However, this one-size-fits-all approach yields suboptimal consistency evaluation for semantic classes exhibiting different temporal behaviors. For instance, highly dynamic objects (\eg, moving motorcycles) require a shorter window to avoid incorrect correspondences due to rapid motion, while static backgrounds (\eg, vegetation) can benefit from a larger window with more correspondences. To address this limitation, we enhance the consistency evaluation for different semantic classes by collecting correspondences across various sliding window sizes.

Specifically, instead of using a single sliding window size $w_\mathrm{t}$, we define a set of $N_\mathrm{t}$ available window sizes denoted as $\mathbf{w}_\mathrm{t}=\{w_d\}_{d=1}^{N_\mathrm{t}}$. Aggregation and voxelization (Equation~(\ref{Eq: Slide Window}-\ref{Eq: Voxel Inverse})) are conducted with each $w_d \in \mathbf{w}_\mathrm{t}$ in the same voxelization grid. As a result, each voxel corresponds to a set of reference predictions 
$\{\boldrm{p}^\mathrm{m}_{\mathrm{r},d}\}_{d=1}^{N_\mathrm{t}}$ (Equation~(\ref{Eq: ST Voxel Reference})) and their corresponding ST entropy $\{E^{\textrm{m}}_{i, k, d}\}_{d=1}^{N_\mathrm{t}}$ (Equation~(\ref{Eq: Average ST Preds})), where $d$ corresponds to the time window index. The resulting set of reference predictions contains various spatial-temporal correspondences lying within different time windows. Thus, its corresponding ST entropy provides a more informative reliability estimation compared to using a single sliding window as in Latte. To gather the estimation from different time windows, we designed an intra-modal weighting mechanism before cross-modal interaction to attend to the reference predictions with less ST entropy, formulated as:
\begin{gather}
    E^{\textrm{m}}_{i, k} = \sum_d^{N_\mathrm{t}} \tau^{\mathrm{m}}_{d} E^{\textrm{m}}_{i, k, d}, \ \ \bar{\boldrm{p}}^{\textrm{m}}_\textrm{r} = \sum_d^{N_\mathrm{t}} \tau^{\mathrm{m}}_{d} \bar{\boldrm{p}}^{\mathrm{m}}_{\mathrm{r}, d}, \ \ \tau^{\mathrm{m}}_{d} = \frac{h_{i, k, d}\tau^{\mathrm{b}}_{d} \mathrm{exp}(E^{\textrm{m}}_{i, k, d})}{\sum_d^{N_\mathrm{t}} h_{i, k, d} \mathrm{exp}(E^{\textrm{m}}_{i, k, d})},\label{Eq: multi-window init weights}
\end{gather}
where $\tau^{\mathrm{m}}_{d}$ is the window weight computed based on the predefined initial weight $\tau^{\mathrm{b}}_{d}$ and its corresponding ST entropy $E^{\textrm{m}}_{i, k, d}$. Since the long-term prediction consistency exhibits more confidence in prediction consistency than the short-term ones, we introduce an initial weight $\tau^{\mathrm{b}}_{d}$ (here we empirically set $\tau^{\mathrm{b}}_{d} = \log_2(2+w_d)$) which is positively related to the window size. The $\alpha$-quantile filtering is also introduced in Equation~(11) to filter out the extremely noisy candidates before our intra-modal attending. Note that $Q^\mathrm{m}(\alpha)$ here is computed based on all ST entropy from different window sizes. The valid index in Equation~(8) is subsequently re-formulated as:
\begin{equation}
    h_{i, k, d} = 
        \begin{cases}
            0,\ \ \textrm{if} \ E^{m}_{i, k, d} > Q^m(\alpha)\\
            1,\ \ \textrm{if} \ E^{m}_{i, k, d} \leq Q^m(\alpha)
        \end{cases},\ \ 
    h_{i, k} = 
        \begin{cases}
            0,\ \ \textrm{if} \ \forall d, h_{i, k, d} = 0\\
            1,\ \ \textrm{if} \ \exists d, h_{i, k, d} = 1
        \end{cases}.\label{Eq: multi-window conf perc}
\end{equation}

\subsubsection{ST Voxel Aided Cross-Modal Learning}\label{sec:Cross-Modal}
Each modality has its pros and cons under different conditions. Therefore, online cross-modal learning should be able to adaptively attend to the reliable modality while suppressing the unreliable one. To achieve this, Latte++ inherits the cross-modal attending mechanism of Latte, which amplifies the contribution of the modality with more consistency and certainty for each ST voxel. Specifically, taking ST voxel $\boldrm{v}^{\textrm{ST}}_{i,k}$ and its corresponding ST entropy $E^{\textrm{m}}_{i, k}$ as an example, the voxel-wise cross-modal attending weights and the weighted cross-modal consistency loss $\mathcal{L}^{\textrm{xM}}_{i, k}$ are computed as:
\begin{align}
    w_\textrm{v}^{\textrm{2D}} &= \frac{\textrm{exp}(E^{\textrm{2D}}_{i, k})}{\textrm{exp}(E^{\textrm{2D}}_{i, k}) + \textrm{exp}(E^{\textrm{3D}}_{i, k})}, \ \ 
    w_\textrm{v}^{\textrm{3D}} = 1 - w_\textrm{v}^{\textrm{2D}}, \label{Eq: ST voxel weights}\\
    \mathcal{L}^{\textrm{xM}}_{i, k} &= w_\textrm{v}^{\textrm{2D}} D_{\textrm{KL}}(\bar{\boldrm{p}}^{\textrm{3D}}_\textrm{q} \| \bar{\boldrm{p}}^{\textrm{2D}}_\textrm{r}) + 
    w_\textrm{v}^{\textrm{3D}} D_{\textrm{KL}}(\bar{\boldrm{p}}^{\textrm{2D}}_\textrm{q} \| \bar{\boldrm{p}}^{\textrm{3D}}_\textrm{r}), \label{Eq: ST xM Cons}
\end{align}
where $D_{\textrm{KL}}(\cdot)$ represents the KL-divergence between two probability. $\bar{\boldrm{p}}^{\textrm{3D}}_\textrm{q}$ is the average query predictions in the ST voxel, similarly computed as in Equation~(\ref{Eq: Average ST Preds}).

The voxel-wise cross-modal attending is further extended to the point level for better online predictions and cross-modal pseudo-label generation. Specifically, the ST entropy is propagated from the ST voxel to its reference point-wise prediction, indicating its confidence and consistency within its spatial-temporal neighborhoods. If one's ST entropy has been filtered out as in Equation~(8), its ST entropy would fall back to the point-level entropy. For an arbitrary point predictions $\boldrm{p}^\textrm{m} \in \boldrm{p}_\textrm{r}^\textrm{m}$, the cross-modal predictions $y^{\textrm{xM}}$ are formulated as:
\allowdisplaybreaks
\begin{align}
    \hat{E}_\textrm{r}^{\textrm{m}} &= h_{i, k}E^{\textrm{m}}_{i, k} - (1 - h_{i, k})\sum_c^K \boldrm{p}^{\textrm{m}}_{c} \log \boldrm{p}^{\textrm{m}}_{c},\label{Eq: Point Ety}\\
    w_\textrm{p}^{\textrm{2D}} &= \frac{\textrm{exp}(\hat{E}^{\textrm{2D}})}{\textrm{exp}(\hat{E}^{\textrm{2D}}) + \textrm{exp}(\hat{E}^{\textrm{3D}})}, \ \ 
    w_\textrm{p}^{\textrm{3D}} = 1 - w_\textrm{p}^{\textrm{2D}},\label{Eq: ST Point Weight}\\
    \boldrm{p}^{\textrm{xM}} &= w_\textrm{p}^{\textrm{2D}}\boldrm{p}^{\textrm{2D}} + w_\textrm{p}^{\textrm{3D}}\boldrm{p}^{\textrm{3D}}, \ \
    y^{\textrm{xM}} = \argmaxA_c \boldrm{p}^{\textrm{xM}} \label{Eq: Pseudo-Labels}.
\end{align}

\setlength{\textfloatsep}{5pt}
\begin{algorithm}[t]
\small
\SetKwInput{kwModelInit}{Init Model}
\SetKwInput{kwTagetDomainInit}{Target Domain}
\DontPrintSemicolon
\SetInd{5pt}{5pt}
\kwTagetDomainInit{Multi-modal input $\mathcal{X}_{\mathcal{T}} = \{(\boldrm{x}^{\textrm{2D}}_{t}, \boldrm{x}^{\textrm{3D}}_{t})\}$.}

\kwModelInit{$\Tilde{\phi}_t^\text{m}$, $\phi_t^\text{m}$, $\text{m}\in\{\text{2D},\text{3D}\}$, window sizes $\boldrm{w}_\mathrm{t}$.} 

\For{$\boldrm{x}=\{(\boldrm{x}_t^{\textrm{2D}}, \boldrm{x}_t^{\textrm{3D}})\}_{t=t_0+1}^{B} \in \mathcal{X}_{\mathcal{T}}$}
{
\textit{(a)} Compute $\Tilde{\boldrm{p}}^{\textrm{m}}_{t}, 
\boldrm{p}^{\textrm{m}}_{t}$ by Equation~(\ref{Eq: Student Teacher}).\;
\For{$i \in [t_0+1, t_0+B]$}
    {
    \eIf{$|\boldrm{w}_\mathrm{t}| == 1$}
        {
        \textit{(b)} Compute $\hat{\boldrm{x}}_{i}^{\textrm{3D}}, \boldrm{v}^{\textrm{3D}}_i, \boldrm{M}^{\textrm{3D}}_i$ within $w_\mathrm{t}$ by Equation~(\ref{Eq: Slide Window})-(\ref{Eq: Voxel Inverse}).\;
        \textit{(c)} Compute $\boldrm{v}_i^{\textrm{ST}}$ and $E^{\textrm{m}}_i$ by Equation~(\ref{Eq: ST Voxel Reference})-(8).\;
        }
        {
        \For{$w_d \in \boldrm{w}_\mathrm{t}$}
            {
                \textit{(d)} Compute $E^{\textrm{m}}_{i, k, d}, \boldrm{p}^\mathrm{m}_{\mathrm{r},d}$ in $w_d$ by Equation~(\ref{Eq: Slide Window})-(\ref{Eq: ST Voxel Reference}).\;
            }
            
        \textit{(e)} Compute $E^{\textrm{m}}_i, \bar{\boldrm{p}}^\mathrm{m}_\boldrm{r}, h_{i,k}$ by Equation~(\ref{Eq: multi-window init weights})-(\ref{Eq: multi-window conf perc}).\;
        }
    }
\textit{(f)} Compute $\mathcal{L}^{\textrm{xM}}_{i}$ by Equation~(\ref{Eq: ST voxel weights})-(\ref{Eq: ST xM Cons}).\;
\textit{(g)} Compute $\hat{E}_\textrm{r}^{\textrm{m}}$ and $\boldrm{y}^{\textrm{xM}}_t$ by Equation~(\ref{Eq: Point Ety})-(\ref{Eq: Pseudo-Labels})\;
\textit{(h)} Compute $\mathcal{L}$ by Equation~(\ref{Eq: Overall Loss}) and update with Equation~(\ref{Eq: Teacher Update})\;
}
\caption{Adaptation and Online Prediction Process of \textbf{Latte} (\textit{(a-c), (f-h)}) and \textbf{Latte++} (\textit{(a), (d-h)})}\label{Algo: PsCode}
\end{algorithm}

\subsubsection{Online Predictions and Optimization}\label{sec:Overall}
As shown in Fig.~\ref{fig:Overall Method}, given a batch of $B$ consecutive frames $\{\boldrm{x}_t^\textrm{m}\}_{t=t_0+1}^{t_0+B}$ as input, we first extract the student and teacher predictions through Equation~(\ref{Eq: Student Teacher}). Subsequently, frames are aggregated and voxelized as in Equation~(\ref{Eq: Slide Window}), resulting in a batch of merged point cloud $\{\hat{\boldrm{x}}_t^\textrm{3D}\}_{t=t_0+1}^{t_0+B}$ and voxels $\{\boldrm{v}_t^\textrm{3D}\}_{t=t_0+1}^{t_0+B}$. The ST voxels and ST entropy computation process are then applied to each frame of the merged point cloud and voxels. Due to the overlap of the sliding window, multiple ST entropy values could be propagated to the same point prediction, where we empirically take the average value of all ST entropy received to proceed Equation~(\ref{Eq: Point Ety}). The overall loss function and the updating scheme can be formulated as:
\allowdisplaybreaks
\begin{align}
    \mathcal{L} &= \sum_t \mathcal{F}(\boldrm{p}^{\textrm{m}}_t, \boldrm{y}^{\textrm{xM}}_t) + \frac{\lambda_{\textrm{xM}}}{B}\sum_t \sum_k \mathcal{L}^{\textrm{xM}}_{t,k} \label{Eq: Overall Loss},\\
    \Tilde{\theta}^{\textrm{m}}_t &= \lambda_s \Tilde{\theta}^{\textrm{m}}_{t-1} + (1 - \lambda_s)\theta^{\textrm{m}}_t, \label{Eq: Teacher Update}
\end{align}
where $\mathcal{F}(\cdot)$ is the cross-entropy function and $\boldrm{y}^{\textrm{xM}}_t$ is the point-wise cross-modal pseudo-label frame from Equation~(\ref{Eq: Pseudo-Labels}), which is also regarded as our cross-modal prediction for evaluation. $\Tilde{\theta}^{\textrm{m}}_t$ and $\theta^{\textrm{m}}_t$ are the model parameters of teacher and student models, respectively, where $\lambda_s$ is the momentum update coefficient. $\lambda_{\textrm{xM}}$ is pre-defined loss coefficient. Our adaptation process is summarized in Algo.~\ref{Algo: PsCode}

\subsection{Interactive Test-Time Adaptation}
\begin{figure}[t]
    \centering
    \includegraphics[width=\linewidth]{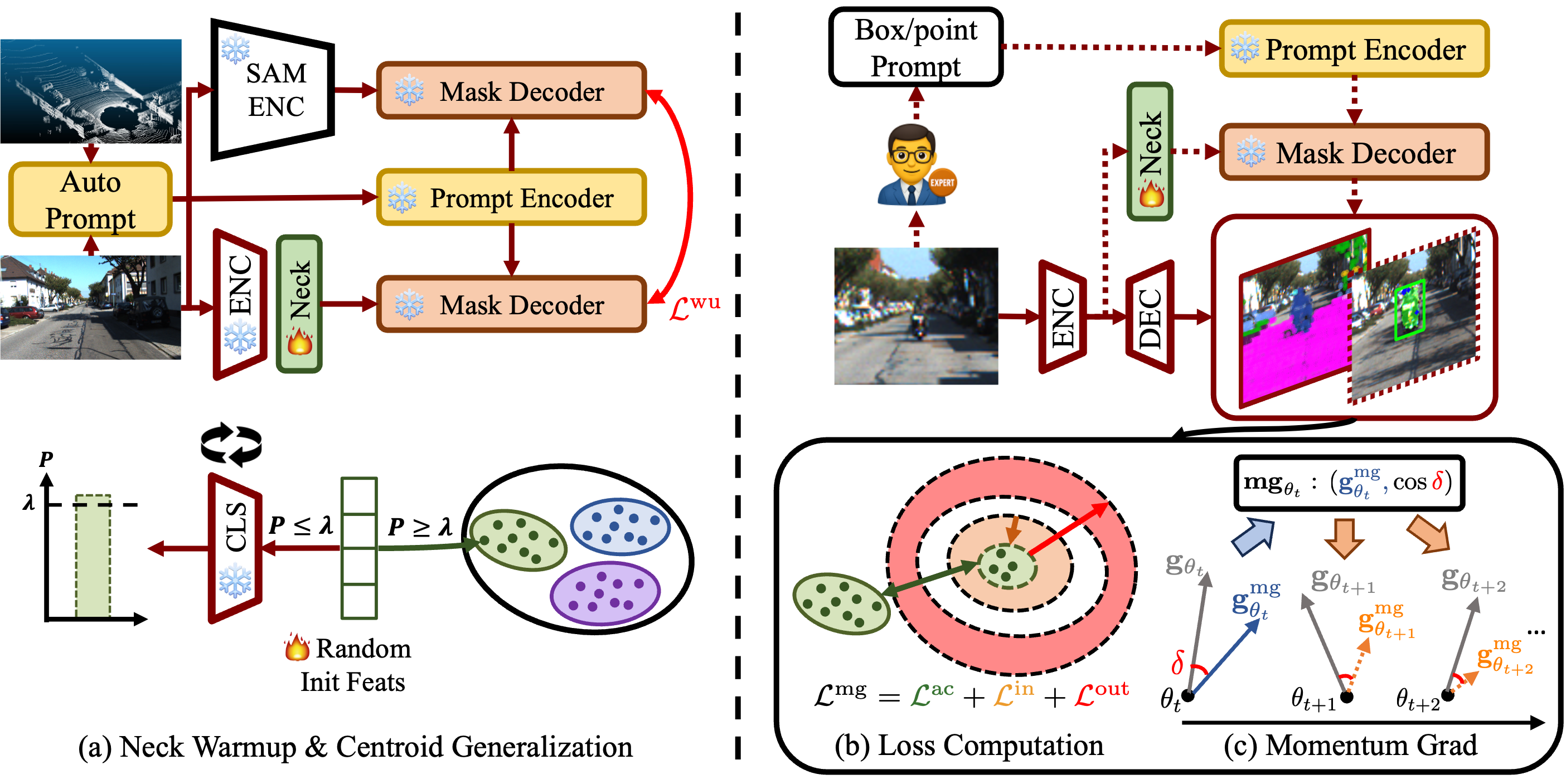}
    \caption{Overall pipeline of ITTA. Before the online adaptation (left), a promptable branch is first created by inserting a light-weight bottleneck after the encoder (ENC) of the 2D network, and then warm up only the bottleneck by distilling the knowledge from interactive visual foundation models (\eg, SAM~\citep{kirillov2023segment}). Meanwhile, class-wise feature centroids are generated by feature inversion using the pre-trained classifier. During the online adaptation process (right), when receiving human prompts, the pre-trained promptable branch decodes a corresponding instance mask, based on which we will capture this valuable knowledge by \textcolor{MidnightBlue}{recording} its loss and gradient, and \textcolor{Orange}{reuse} it to update the network in the following iterations.}
    \label{fig: ch6_itta}
\end{figure}

Existing TTA and MM-TTA methods perform online adaptation based on self-refinement, whose effectiveness varies significantly across different classes. The less frequent foreground classes are substantially more vulnerable to domain shifts than the dominant background classes, leading to more severe error accumulation during self-refinement online. To address this limitation, we proposed ITTA, a flexible MM-TTA add-on that provides an interface to capture efficient human prompts as feedback to better facilitate the online adaptation of vulnerable foreground classes. Specifically, our intuition is to leverage the emerging interactive segmentation~\citep{huang2023interformer,kirillov2023segment} as a portal for receiving efficient human prompts of foreground objects (\eg, points and boxes), and effectively capture this human knowledge to enhance the online adaptation. As shown in Fig.~\ref{fig: ch6_itta}, before the online adaptation, we first warm up a newly introduced branch for prompt processing and generate class-wise feature centroids (Section~\ref{subsec:ch6_warmup}). During the online adaptation process, the warmed-up promptable branch can receive a human prompt and output a decode instance mask, which facilitates the adaptation of the corresponding semantic class (Section~\ref{subsec: ch6_momentum_grad}).

\subsubsection{Prior to Interactive Test-Time Adaptation}\label{subsec:ch6_warmup}
\textbf{Promptable Branch and Warm up.} To incorporate interactive segmentation with TTA methods for full segmentation, one intuitive solution is to apply the existing interactive segmentation algorithm in parallel directly, but it brings an extra computational burden. Since semantic segmentation and interactive segmentation are two closely related tasks, we argue that their encoder are mostly shareable, motivating us to design a light-weight promptable branch\footnote{Structures are detailed in the appendices.} on top of the semantic segmentation network. Specifically, by decomposing the 2D network as $\phi^{\mathrm{2D}}(\cdot) = f^\mathrm{2D}(g^\mathrm{2D}(\cdot))$ where $g^\mathrm{2D}(\cdot)$ and $f^\mathrm{2D}(\cdot)$ denote the 2D encoder and decoder, respectively, the promptable branch predicts the prompt-related mask as:
\begin{gather}
    \textbf{f}         = \psi^\mathrm{n}(g^\mathrm{2D}(\textbf{x}^\mathrm{2D})),\ \
    \tilde{\textbf{f}} = \psi^\mathrm{p}(\chi),\ \ 
    \textbf{p}^\mathrm{p} = \psi^\mathrm{m}(\textbf{f}, \tilde{\textbf{f}}), \label{Eq: ch6_mask_predict}
\end{gather}
where $\psi^\mathrm{n}$ is the light-weight bottleneck consisting of several 2D convolutional layers to extract promptable image embedding $\textbf{f}$. $\psi^\mathrm{p}$ and $\psi^\mathrm{m}$ are the prompt encoder and mask decoder that output the prompt embedding $\tilde{\textbf{f}}$ and predicted mask probability $\textbf{p}^\mathrm{p}$, respectively. Here, we leverage the same prompt encoder and mask decoder from the existing visual foundation model SAM~\citep{kirillov2023segment}.

Prior to deploying the promptable branch, the newly inserted lightweight bottleneck is warmed up to extract the interactive image embedding in the source domain, where the training data is retrieved as promptable individual instances in each image through 3D point clustering and point-pixel correspondences. Specifically, given a scan of multi-modal input and ground-truth labels denoted as $(\textbf{x}^{\mathrm{2D}}, \textbf{x}^{\mathrm{3D}}, \textbf{y})$ and a pre-defined foreground class of interest $\mathcal{C} = \{c_i\}_{i=1}^{N_\mathcal{C}}$, the foreground instance and its prompts are generated as:
\begin{gather}
    \tilde{\textbf{x}}^{\mathrm{3D}} = \{\textbf{x}_i^{\mathrm{3D}} | \textbf{y}_i = c\}, \ \ \textbf{y}^\mathrm{inst} = \texttt{DBSCAN}(\tilde{\textbf{x}}^{\mathrm{3D}}),\\
    \hat{y} \sim \mathrm{Unif}(\mathrm{set}(\textbf{y}^\mathrm{inst})), \ \ \tilde{\textbf{x}}^\mathrm{pmt} = \texttt{Proj}(\{\tilde{\textbf{x}}^\mathrm{3D}_m | \textbf{y}^\mathrm{inst}_m = \hat{y}\}),
\end{gather} where $c$ is a random instance class sampled from $\mathcal{C}$ ($c \sim \mathrm{Unif}(\mathcal{C})$, likewise for $\hat{y}$). $\textbf{y}^{\mathrm{inst}}$ contains the geometric cluster ID predicted by DBSCAN~\citep{ester1996density} and $\texttt{Proj}(\cdot)$ denotes the 3D-to-2D projection between sensors. $\tilde{\textbf{x}}^\mathrm{pmt}$ therefore contain all valid 2D point prompts for instance $\hat{y}$.

Given the extracted instance and prompts in each image, the warm-up process is designed to simulate the interaction with human experts, where the 2D networks are trained to generate a similar predicted mask as the interactive visual foundation model (\eg, SAM~\citep{kirillov2023segment}) using the same prompts and image input. Specifically, given the number of all valid point prompts and the pre-defined lower and upper bound of interactions (i.e., clicks) denoted as $\rho_\mathrm{min}, \rho_\mathrm{max}$, respectively, the number of clicks is sampled as $\rho \sim \mathcal{U}_\mathbb{Z}(\mathrm{min}(\rho_\mathrm{min}, \|\tilde{\textbf{x}}^\mathrm{pmt}\|), \mathrm{min}(\rho_\mathrm{max}, \|\tilde{\textbf{x}}^\mathrm{pmt}\|))$. For each round $j \in [1, \rho]$, the input of the prompt encoder are formulated as: 
\begin{gather}
    \chi_{(j)} = \{\textbf{x}_{(j)}^{\mathrm{pmt}}, \textbf{x}^{\mathrm{box}}, \textbf{p}^\mathrm{p}_{(j-1)}\},
\end{gather}
where $\chi_{(j)}$ is the $j$-round of prompt input. $\textbf{x}_{(j)}^{\mathrm{pmt}}$ denotes $j$ prompting points and $\textbf{x}^{\mathrm{box}}$ represents the 2D bounding box of instance derived from $\tilde{\textbf{x}}^\mathrm{pmt}$. $\textbf{p}^\mathrm{p}_{(j-1)}$ is the predicted mask from the previous iteration, which equals the 0-matrix if $j = 0$. For each round $j$, a new prompting point is generated based on the clicking simulation strategy~\citep{huang2023interformer} and appended to $\textbf{x}_{(j-1)}^{\mathrm{pmt}}$ of the previous round to update prompting points of this round $\textbf{x}_{(j)}^{\mathrm{pmt}}$. After $\rho$ rounds of interaction, the image and prompt input $\chi_{(\rho)}$ are fed into SAM to generate the mask prediction denoted as $\tilde{\textbf{y}}^\mathrm{p}$, which is applied to supervise the $\rho$-round predicted mask of our promptable branch with a  warmup loss:
\begin{gather}
    \mathcal{L}_{\mathrm{wu}} = FL(\textbf{p}^\mathrm{p}, \tilde{\textbf{y}}^\mathrm{p}; \gamma,\alpha), \label{eq: ch6_focal}
\end{gather}
where $FL(\cdot, \cdot\,; \gamma, \alpha)$ denotes the normalized focal loss~\citep{lin2017focal} with focusing and balanced parameters as $\gamma$ and $\alpha$, respectively. The warmup loss $\mathcal{L}_{\mathrm{wu}}$ is used to update the bottleneck $\psi^\mathrm{n}$ with all other parameters remaining frozen. 

\textbf{Class-wise Centroid Generation.} Sharing a similar merit with previous clustering-based TTA methods~\citep{su2023revisiting,Cao_2023_ICCV}, we aim to collect the class-wise feature centroids as prior knowledge of the source domain. To better cope with the source-free MM-TTA setting without accessing the source domain data, we propose generating class-wise centroids by feature inversion using pre-trained classifiers. Specifically, given the last-layer linear classifier of each modality, we first randomly initialize $N_f$ features from the normal distribution for each $c \in \mathcal{C}$. The features are then updated by optimizing the predicted scores of the frozen classifier. After all scores exceed the pre-defined threshold $\lambda$, the mean and covariance of these features are stored as the class-wise centroid for the following online adaptation process, denoted as $\mu^\mathrm{m}_c$ and $\Sigma^\mathrm{m}_c$ ($\mathrm{m} \in \{\mathrm{2D}, \mathrm{3D}\}$), respectively.

\subsubsection{Online Optimization Objectives and Momentum Grad}\label{subsec: ch6_momentum_grad} 
In MM-TTA, we assume human prompts only occasionally occur to avoid overcompromising the adaptation efficiency online. This scarcity of human feedback poses a critical challenge: how to maximally leverage such precious but rare interventions throughout the adaptation process. To tackle this challenge, our ITTA introduces a two-fold strategy. First, it leverages a feature-based clustering objective (Fig.~\ref{fig: ch6_itta}(b)) that directly advances feature alignment with the class-wise feature centroid. Second, to extend the impact of human feedback even beyond its availability, our momentum gradient (Fig.~\ref{fig: ch6_itta}(c)) records the gradient regarding this clustering objective as valuable human knowledge, which is further reused as regularization when human feedback is unavailable. In ITTA, we assume the existence of student and teacher models as in Section~\ref{sec:Frame-wise Preds}. Given a pair of multi-modal inputs with human prompts, the probability mask $\tilde{\mathbf{p}}^\mathrm{p}$ is computed as in Equation~(\ref{Eq: ch6_mask_predict}) using the 2D teacher $\tilde{\phi}^\mathrm{2D}(\cdot)$. Besides utilizing $\tilde{\mathbf{p}}^\mathrm{p}$ to update pseudo-labels, we further introduce three additional components to enhance the impact of human feedback.

\textbf{Mask Refinement Objective.}
To mitigate the impact of domain discrepancy on the promotable branch, we propose a mask refinement objective to adapt the promptable branch online. Specifically, we regard the multi-prompt mask $\tilde{\mathbf{p}}^\mathrm{p}$ output by the teacher model as pseudo-labels to supervise the masks predicted by the student model with a single point prompt, formulating an objective as:
\begin{equation}
    \mathcal{L}^\mathrm{p} = \lambda_\mathrm{p} FL(\mathbf{p}^\mathrm{p},\mathbf{M}^\mathrm{p} ; \gamma, \alpha),
\end{equation}
where $\mathbf{M}^\mathrm{p}$ is the pseudo-labels derived by thresholding $\tilde{\mathbf{p}}^\mathrm{p}$ (we empirically follow the same threshold as SAM~\citep{kirillov2023segment}). $\mathbf{p}^\mathrm{p}$ is the student's predicted logits of the instance mask with the first point prompt only, while $\lambda_\mathrm{p}$ is the pre-defined coefficient.

\textbf{Feature-based Clustering Objective.}
Given high-level features of the predicted instance mask as the anchor, we aim to mitigate the discrepancy it shares with the corresponding class-wise centroids. To further propagate this prior knowledge to the same-class predictions within the current scan, the cluster-aware alignment is promoted by compacting the feature inliers around anchors and rejecting outliers through repulsive forces. Formally, given the instance mask $\textbf{M}^\mathrm{p}$ from thresholding $\tilde{\mathbf{p}}^\mathrm{p}$ and normalized high-level features $\textbf{f}^\mathrm{m}$ from the student network, the inlier and outlier masks $\textbf{M}^\mathrm{in}$ and $\textbf{M}^\mathrm{out}$ are first computed as: 
\begin{gather}
    \textbf{f}^\mathrm{a} = \mathrm{avg}(\textbf{M}^\mathrm{p} \otimes \textbf{f}^\mathrm{m}), \ \
    \textbf{s}            = \mathrm{exp}(
        \langle\textbf{f}^\mathrm{a}, \textbf{M}' \otimes \textbf{f}^\mathrm{m} \rangle),\\
    \textbf{M}^\mathrm{in}  = \mathbbm{1}_{\textbf{s} \geq \tau^\mathrm{in}}, \ \
    \textbf{M}^\mathrm{out} = \mathbbm{1}_{\textbf{s} < \tau^\mathrm{out}},
\end{gather}
where $\textbf{M}'$ is the feature mask selecting those with the same predicted class as the anchor, and $\langle\cdot\rangle$ denotes the element-wise computation of cosine similarity. $\tau^\mathrm{in}$ and $\tau^\mathrm{out}$ represent the pre-defined filtering threshold for inliers and outliers, respectively. Our feature-based clustering objective can then be formulated as:
\allowdisplaybreaks
\begin{equation}
    \mathcal{L}^\mathrm{mg} = \underbrace{\lambda_\mathrm{ac} \langle \textbf{f}^\mathrm{a},\mu^\mathrm{m}_c \rangle}_{\mathcal{L}^\mathrm{ac}} + \underbrace{\frac{\lambda_\mathrm{cls}}{\sum\textbf{M}^\mathrm{in}}\sum(\textbf{M}^\mathrm{in}\otimes\textbf{s})}_{\mathcal{L}^\mathrm{in}} + \underbrace{\frac{\lambda_\mathrm{cls}}{\sum\textbf{M}^\mathrm{out}}\sum(-\textbf{M}^\mathrm{out}\otimes\textbf{s})}_{\mathcal{L}^\mathrm{out}}, \label{eq: ch6_mg_loss}
\end{equation}
where $\mathcal{L}^\mathrm{ac}$ is the cosine similarity between the average anchor and corresponding class-wise centroid. $\mathcal{L}^\mathrm{in}$ and $\mathcal{L}^\mathrm{out}$ are the sub-objectives to achieve compact clustering with inliners and rejecting outliers as shown in Fig.~\ref{fig: ch6_itta}(b), respectively. $\lambda_\mathrm{ac}$ and $\lambda_\mathrm{cls}$ represent their corresponding loss coefficients.

\textbf{Momentum Gradient.}
While the aforementioned clustering objectives help instantly rectify the class-specific domain shift, they are only available when human feedback is received. To further extend this valuable regularization, we regard the gradient corresponding to the clustering objective $\mathcal{L}_\mathrm{mg}$ as a prior, which can be captured and reused to influence the long-term updating direction more broadly. Taking an arbitrary time step $t$ with human prompts received as an example, given the overall loss of any TTA method and the clustering objective denoted as $\mathcal{L}_t$ and $\mathcal{L}^\mathrm{mg}_t$, respectively, we record the gradient information for any trainable parameter $\theta_t$: 
\begin{gather}
    \textbf{mg}_{\theta_t} = \{
    \underbrace{\textbf{g}_{\theta_t}^\mathrm{mg}}_{\mathrm{gradient}},
    \underbrace{\langle\textbf{g}_{\theta_t}^\mathrm{mg}, \textbf{g}_{\theta_t}\rangle}_{\mathrm{relative\ angle}}
    \},\label{Eq: ch6_store_mg}
\end{gather}
where $\textbf{g}_{\theta_t}^\mathrm{mg} = \nabla_{\theta_t}\mathcal{L}^\mathrm{mg}_t$ and $\textbf{g}_{\theta_t} = \nabla_{\theta_t}\mathcal{L}_t$ are the gradient of $\mathcal{L}^\mathrm{mg}_t$ and $\mathcal{L}_t$ with respect to $\theta_t$ computed during the backpropagation. The momentum gradient $\textbf{mg}_{\theta_t}$ encompasses both the magnitude and the relative angle between the overall gradient, which can be viewed as guidance from human feedback that rectifies the biased optimization direction.
For the consecutive iteration $k = t+\Delta t$, the momentum gradient is reused by preserving the same relative angle with the overall optimization direction $\textbf{g}_{\theta_k}$ to achieve a similar regularization effect in a decaying manner, formulated as:
\begin{gather}
    \textbf{u}_k = \textbf{g}^\mathrm{mg}_{\theta_t} - \frac{\textbf{g}_{\theta_k}^\top \textbf{g}^\mathrm{mg}_{\theta_t}}{\|\textbf{g}_{\theta_k}\|^2} \textbf{g}_{\theta_k}, \ \
    \hat{\textbf{u}}_k = \frac{\textbf{u}_k}{\|\textbf{u}_k\|},                                                                                                                    \\
    \textbf{g}^\mathrm{mg}_{\theta_k} = \lambda^\mathrm{mg}_k\|\textbf{g}^\mathrm{mg}_{\theta_t}\|
    (\cos(\delta) \textbf{g}_{\theta_k} + \sin(\delta)\hat{\textbf{u}}_k),\label{eq: ch6_grad_disentangle}
\end{gather}
where $\hat{\textbf{u}}_k$ is the unit vector orthogonal to $\textbf{g}_{\theta_k}$ within the same plane of $\textbf{g}_{\theta_k}$ and $\textbf{g}^\mathrm{mg}_{\theta_t}$ and $\lambda^\mathrm{mg}_k = (\gamma^\mathrm{mg})^{(k-t)}$ is the scaling factor with a decaying ratio of $\gamma^\mathrm{mg}$. In practice, we set a maximum valid window of $\Delta t_\mathrm{max}$ of the momentum gradient (\ie, $\lambda^\mathrm{mg}_k=0$, if $k-t > \Delta t_\mathrm{max}$). $\delta$ is the relative angle derived from Equation~\ref{Eq: ch6_store_mg} as $\delta = \arccos(\textbf{g}_{\theta_t}^\mathrm{mg}, \textbf{g}_{\theta_t})$. The momentum gradient $\textbf{g}^\mathrm{mg}_{\theta_k}$ is added to $\textbf{g}_{\theta_k}$ to achieve regularization.

\section{Experimental Results}
In this section, we present our thorough experimental results on three different benchmarks. Details of benchmarks, backbones, and settings are first presented in Section~\ref{sec:Exp Settings}. The main results are then illustrated in Section~\ref{sec:Overall Results}, followed by detailed ablation studies and qualitative results in Section~\ref{sec:Ablation Studies}.

\subsection{Benchmarks and settings}\label{sec:Exp Settings}
\textbf{Benchmark Details}. To thoroughly investigate the effectiveness of Latte, we conduct our experiments on 3 different MM-TTA benchmarks, including (i) \textbf{USA-to-Singapore (U-to-S)}, (ii) \textbf{A2D2-to-SemanticKITTI (A-to-K)}, and (iii) \textbf{Synthia-to-SemanticKITTI (S-to-K)}. Specifically, U-to-S is developed from NuScenes-LiDARSeg~\citep{caesar2020nuscenes} dataset, where the domain gap is mainly credited to the infrastructure difference between countries. Same as~\citep{jaritz2022cross}, the source domain and the target domain data are selected by filtering country keywords based on the data recording description in NuScenes. Different from previous UDA methods~\citep{jaritz2022cross,peng2021sparse} utilizing self-defined class mappings, we follow the official mapping in NuScenes-LiDARSeg, forming a 16-class segmentation benchmark. The remaining two benchmarks follow a similar setting as in~\citep{shin2022mm}, where the domain gap of A-to-K~\citep{geyer2020a2d2,behley2019semantickitti} lies in different LiDAR mounting positions and image resolutions, while the one of S-to-K~\citep{behley2019semantickitti} lies in different patterns between synthetic and real data for both modalities. For A-to-K, we adopt the same class-mapping as in~\citep{jaritz2022cross, shin2022mm}, which leads to a 10-class benchmark, while we design our class-mapping for S-to-K since its original class map has not been revealed previously~\citep{shin2022mm}, forming a 9-class benchmark.

In addition to the standard MM-TTA benchmarks with stationary target domains, we further extend our comparison to its more challenging variant, named Multi-Modal Continuous Test-Time Adaptation (MM-CTTA), where the target domain is continuously changing with different weather conditions or illumination conditions~\citep{wang2022continual, Cao_2023_ICCV}. Specifically, we leverage the two MM-CTTA benchmarks for 3D segmentation proposed in~\citep{Cao_2023_ICCV}, including \textbf{SemanticKITTI-to-Synthia (K-to-S)} and \textbf{SemanticKITTI-to-Waymo (K-to-W)}, where the former one is a challenging synthetic-to-real benchmark with large domain discrepancy, while the latter is less challenging but closer to real-life deployment. Both benchmarks are preprocessed and organized following the identical protocol as in~\citep{Cao_2023_ICCV}. \textit{More details are presented in our appendices.}

\textbf{Baseline methods}. Previous TTA and MM-TTA methods are included in comparison with Latte++. For TTA methods, we compare Latte++ with point-wise pseudo-labels with filtering (PsLabel), TENT~\citep{wang2020tent}, ETA~\citep{niu2022efficient}, and SAR~\citep{niu2023towards} with and without restoring (denoted as SAR-rs and SAR, respectively). Latte~\citep{cao2025reliable}, MMTTA~\citep{shin2022mm}, as well as its online version of xMUDA~\citep{jaritz2022cross} (including pure xMUDA and pseudo-label version $\mathrm{xMUDA}_\mathrm{PL}$) are additionally included as the previous SOTA methods for comparison. Considering that EATA~\citep{niu2022efficient} with Fisher regularizer requires pre-access to the target domain samples, we compare with ETA, an official variant of EATA without this regularizer for a fair comparison. Since our extended evaluation involves a new challenging scenario, MM-CTTA, two SOTA CTTA methods (CoTTA~\citep{wang2022continual} and CoMAC~\citep{Cao_2023_ICCV}) based on multiple augmented predictions are additionally introduced to justify our effectiveness and efficiency under different MM-TTA settings. We further present a performance lower bound by directly testing with source pre-trained models (\ie, Source only) and an upper bound by online adapting networks with ground-truth labels from the target domain for one epoch (\ie, Oracle TTA). To testify to the effectiveness of ITTA, we present the performance comparison of its integration with four commonly used MM-TTA methods, including $\mathrm{xMUDA}_\mathrm{PL}$, MMTTA, Latte, and Latte++. 

\textbf{Implementation details}. With the rapid progress in 2D and 3D segmentation, we employ all baseline methods, Latte++, and ITTA variants with more recent SegFormer~\citep{xie2021segformer} (SegFormer-B1) and SPVCNN~\citep{tang2020searching} (SPVCNN-cr1) as their 2D and 3D backbones, respectively, to investigate the improvement of existing TTA methods when combined with advanced networks. The pre-training procedures on the source domain follow a similar setting as in~\citep{jaritz2022cross}, except for 2D SegFormer, which is trained with a base learning rate of 6E-5 with an AdamW optimizer and Poly scheduler as in~\citep{xie2021segformer}. For Latte, we maintain the same learning rate and optimizer as in the pre-training stage, while the scheduler is disabled. For our voxelization, we leverage the off-the-shelf SLAM algorithm KISS-ICP~\citep{vizzo2023kiss} to generate poses and utilize a window size $w_\mathrm{t}$ of 3 with a voxel size of 0.2m, while the quantile $\alpha$ and coefficient $\lambda_{\mathrm{xM}}$ are set to 0.9 and 0.3, respectively. $\lambda_s$ is empirically set to 0.99 as in~\citep{song2023ecotta,Cao_2023_ICCV}. The window sizes of Latte++ $\mathbf{w}_{\mathrm{t}}$ are set to $[3, 5]$.

In terms of ITTA, we empirically set the classes of interest $C$ as $C=\{\texttt{pedestrian},\texttt{bicycle}\}$ since they possess the most inferior class-wise IoU compared to others. The warmup iteration is set to 10,000 iterations with $\rho_\mathrm{min}=1$ and $\rho_\mathrm{max}=10$. $\gamma$ and $\alpha$ in Equation~\ref{eq: ch6_focal} are set to 2.0 and 0.5, respectively. For the class-wise centroid generation, we randomly initialize five batches of 1024 random features and conduct feature inversion till their prediction scores exceed 0.8. During the online adaptation process, we empirically use the random points and bounding boxes to simulate the online human prompt. The prompt generation is identical to our promptable branch pre-training procedures with an activating ratio of $p_I = 0.01$ and the adjusted $\rho_\mathrm{min} = 5$ to encourage more informative point prompts. $\tau_\mathrm{in}$ and $\tau_\mathrm{out}$ are pre-defined as 0.9 and 0.7, respectively, while the loss coefficients are set as $\lambda_\mathrm{p} = 0.01, \lambda_\mathrm{ac} = 1.0, \lambda_\mathrm{cls} = 1.0$. For the momentum gradient, the decaying ratio $\gamma^\mathrm{mg}$ is set as 0.9 with $\Delta t_\mathrm{max} = 10$.

For all methods, we strictly follow the \textit{one-pass} protocol to first evaluate the predictions of the input batch and then update the networks. For most baseline and Latte/Latte++, we follow previous works~\citep{wang2020tent,niu2022efficient} to update only the trainable parameters in normalization layers, while CoTTA and CoMAC are used to update all trainable parameters as stated in their original protocols~\citep{wang2022continual,Cao_2023_ICCV}. For all baselines, we conduct a parameter search and report their best results. We utilize the mean intersection over union (mIoU) as our evaluation metric. All experiments except ITTA-related ones are conducted with PyTorch on a single RTX 3090. ITTA-related experiments are tested with RTX 5090 for multi-click prompting with a larger GPU RAM, and we testify that this hardware change has a trivial impact on our results.

\begin{table*}[t]
    \centering
    \setlength{\tabcolsep}{4pt}
    \caption{Performance (mIoU) comparison on MM-TTA benchmarks. Latte++ outperforms all previous SOTA methods on the cross-modal metric (xM) across three benchmarks. Here ``MM'' denotes whether the multi-modal interaction is used or not. ``Avg'' presents the average xM performance across three benchmarks. ``Inf. Time'' is computed as the average inference time of three benchmarks. Digits in \textcolor{MidnightBlue}{blue} and \textcolor{Orange}{orange} indicate the \textcolor{MidnightBlue}{\textbf{improve}} and \textcolor{Orange}{\textbf{decrease}} brought by ITTA, respectively.}
    \huge
    \resizebox{\textwidth}{!}{
    \begin{tblr}{ 
        colspec                             = {l c c c c *{12}{c} | c },
        cell{1}{1}                          = {r=2}{l},
        cell{1}{2,3,4,5,9,13,17,18}         = {r=2}{c}, 
        cell{18, 21, 24, 27}{2,3, 4}        = {r=2}{c}, 
        cell{1}{6, 10, 14}                  = {c=3}{c}
        }
    \hline
    \hline
    Method & Prompt & MM & \makecell{Inf. Time\\(ms)} & &
    U-to-S & & & & A-to-K & & & & S-to-K & & & & \SetCell{c, gray9}Avg\\
    \hline
    & & & & & 
    2D & 3D & \SetCell{c, gray9}xM & & 
    2D & 3D & \SetCell{c, gray9}xM & & 
    2D & 3D & \SetCell{c, gray9}xM & & \Tstrut\\
    
    \hline
    Source only  & \xmark & \xmark & 52.7 & & 31.4 & 41.1 &
    \SetCell{c, gray9}43.9 & & 47.4 & 17.9 & \SetCell{c, gray9}44.3 & & 23.4 &
    36.4 & \SetCell{c, gray9}38.2 & & \SetCell{c, gray9}42.1\Tstrut\\
    
    Oracle TTA  & \xmark & \xmark & 184.7 & & 38.7 & 45.5 &
    \SetCell{c, gray9}50.3 & & 49.1 & 61.3 & \SetCell{c, gray9}62.5 & & 36.0 &
    54.5 & \SetCell{c, gray9}54.6 & & \SetCell{c, gray9}55.8\\
    
    \hline
    TENT & \xmark & \xmark & 154.6 & & 36.8 & 36.1 &
    \SetCell{c, gray9}41.1 & & 43.3 & 44.4 & \SetCell{c, gray9}49.1 & & 22.4 &
    35.5 & \SetCell{c, gray9}37.5 & & \SetCell{c, gray9}42.6\Tstrut\\
    
    
    ETA & \xmark & \xmark & 65.5 & & 36.7 & 34.7 &
    \SetCell{c, gray9}43.7 & & 43.2 & 42.5 & \SetCell{c, gray9}49.7 & & 22.4 &
    30.6 & \SetCell{c, gray9}32.9 & & \SetCell{c, gray9}42.1 \\
    
    SAR & \xmark & \xmark & 270.0 & & 36.6 & 37.2 &
    \SetCell{c, gray9}43.9 & & 43.0 & 42.7 & \SetCell{c, gray9}50.3 & & 20.7 &
    31.3 & \SetCell{c, gray9}32.9 & & \SetCell{c, gray9}42.4\\
    
    SAR-rs & \xmark & \xmark & 271.2 & & 36.6 & 35.5 &
    \SetCell{c, gray9}43.9 & & 43.1 & 43.8 & \SetCell{c, gray9}50.1 & & 24.2 &
    25.1 & \SetCell{c, gray9}28.8 & & \SetCell{c, gray9}40.9\\

    TENT+xM & \xmark & \cmark & 155.3 & & 37.1 & 39.7 &
    \SetCell{c, gray9}45.2 & & 43.3 & 47.1 & \SetCell{c, gray9}50.3 & & 23.9 &
    36.5 & \SetCell{c, gray9}37.7 & & \SetCell{c, gray9}44.4\Tstrut\\
    ETA+xM & \xmark & \cmark & 67.0 & & 36.9 & 35.2 &
    \SetCell{c, gray9}44.8 & & 43.1 & 44.2 & \SetCell{c, gray9}50.1 & & 23.7 &
    30.8 & \SetCell{c, gray9}33.1 & & \SetCell{c, gray9}42.7 \\
    SAR+xM & \xmark & \cmark & 270.1 & & 36.9 & 39.4 &
    \SetCell{c, gray9}45.5 & & 43.1 & 46.3 & \SetCell{c, gray9}50.5 & & 22.8 &
    32.8 & \SetCell{c, gray9}34.5 & & \SetCell{c, gray9}43.5 \\
    SAR-rs+xM & \xmark & \cmark & 272.1 & & 36.9 & 36.8 &
    \SetCell{c, gray9}45.0 & & 43.1 & 45.6 & \SetCell{c, gray9}50.5 & & 23.8 &
    30.7 & \SetCell{c, gray9}33.1 & & \SetCell{c, gray9}42.9 \\
    
    xMUDA & \xmark & \cmark & 168.0 & & 19.4 & 22.9 &
    \SetCell{c, gray9}24.2 & & 13.1 & 33.0 & \SetCell{c, gray9}32.2 & & 12.2 &
    14.8 & \SetCell{c, gray9}14.1 & & \SetCell{c, gray9}23.5\\
    
    CoTTA & \xmark & \cmark & 989.6 & & 37.0 & 34.5 &
    \SetCell{c, gray9}43.7 & & 46.0 & 42.2 & \SetCell{c, gray9}50.5 & & 30.7 &
    35.1 & \SetCell{c, gray9}35.6 & & \SetCell{c, gray9}43.3\\
    
    CoMAC & \xmark & \cmark & 995.7 & & 37.3 & 35.6 &
    \SetCell{c, gray9}42.7 & & 43.5 & 42.8 & \SetCell{c, gray9}51.5 & & 25.3 &
    30.2 & \SetCell{c, gray9}32.4 & & \SetCell{c, gray9}42.2\\
    
    PsLabel & \xmark & \cmark & 183.3 & & \textbf{37.7} & 35.8 &
    \SetCell{c, gray9}41.8 & & 46.7 & 44.3 & \SetCell{c, gray9}50.0 & & 29.2 &
    33.3 & \SetCell{c, gray9}35.3 & & \SetCell{c, gray9}42.4\\
    
    \hline

    xMUDA\textsubscript{PL} & \xmark & \cmark &
    168.7 & & 36.2 & 38.2 & \SetCell{c, gray9}43.0 & & 42.8 & 48.3 &
    \SetCell{c, gray9}50.9 & & 30.9 & 34.1 & \SetCell{c, gray9}36.3 & &
    \SetCell{c, gray9}43.4\Tstrut\\
    \textbf{I-}$\mathbf{xMUDA}_\mathbf{PL}$ & 
    \cmark & \cmark & - & & 36.5 & 40.9 &
    \SetCell{c, gray9}45.5 & & 45.6 & 48.3 & \SetCell{c, gray9}52.5 & & 30.1 &
    35.8 & \SetCell{c, gray9}38.6 & & \SetCell{c, gray9}45.5\Tstrut \\
    \textit{Improv.$\uparrow$}& & & & & \improve{0.3} & \improve{2.7} &
    \SetCell{c, gray9}\improve{2.2} & & \improve{2.8} & \improve{0.0} &
    \SetCell{c, gray9}\improve{1.6} & & \decrease{0.8} & \improve{1.7} &
    \SetCell{c, gray9}\improve{2.3} & &
    \SetCell{c, gray9}\improve{2.1}\Bstrut\\

    \hline
    MMTTA & \xmark & \cmark & 184.3 & & 37.2 &
    \underline{41.5} & \SetCell{c, gray9}45.4 & & 44.5 & 51.7 &
    \SetCell{c, gray9}53.7 & & 27.5 & 35.1 & \SetCell{c, gray9}35.5 & &
    \SetCell{c, gray9}44.9\Tstrut\\
    \textbf{I-MMTTA}
    & \cmark & \cmark & - & &
    37.1 & \textbf{42.4} & \SetCell{c, gray9}\textbf{46.5} & & 48.5 &
    \textbf{53.2} & \SetCell{c, gray9}\underline{56.5} & & 27.7 & 37.1 &
    \SetCell{c, gray9}37.6 & & \SetCell{c, gray9}46.9\Tstrut \\
    \textit{Improv.$\uparrow$}& & & & & \decrease{0.1} & \improve{0.9} &
    \SetCell{c, gray9}\improve{1.1} & & \improve{4.0} & \improve{1.5} &
    \SetCell{c, gray9}\improve{2.8} & & \improve{0.2} & \improve{2.0} &
    \SetCell{c, gray9}\improve{2.1} & &
    \SetCell{c, gray9}\improve{2.0}\Bstrut\\

    \hline
    \textbf{Latte} & \xmark & \cmark & 221.1 & & 37.5 &
    41.0 & \SetCell{c, gray9}46.2 & & 46.0 & \underline{52.5} &
    \SetCell{c, gray9}54.2 & & 33.3 & 39.3 & \SetCell{c, gray9}41.7 & &
    \SetCell{c, gray9}47.3\Tstrut\\
    \textbf{I-Latte}
    & \cmark & \cmark & - & &
    37.2 & 41.2 & \SetCell{c, gray9}46.3 & & \textbf{49.8} & 52.1 &
    \SetCell{c, gray9}\textbf{57.5} & & \textbf{33.7} & \textbf{40.5} &
    \SetCell{c, gray9}\textbf{44.0} & &
    \SetCell{c, gray9}\textbf{49.3}\Tstrut\\
    \textit{Improv.$\uparrow$}& & & & & \decrease{0.3} & \improve{0.2} &
    \SetCell{c, gray9}\improve{0.1} & & \improve{3.8} & \decrease{0.4} &
    \SetCell{c, gray9}\improve{3.3} & & \improve{0.4} & \improve{1.2} &
    \SetCell{c, gray9}\improve{2.3} & &
    \SetCell{c, gray9}\improve{2.0}\Bstrut\\

    \hline
    \textbf{Latte++}
    & \xmark & \cmark & 242.2 & & \underline{37.6} & 41.0 &
    \SetCell{c, gray9}46.3 & & 46.4 & \textbf{53.0} & \SetCell{c, gray9}55.1 &
    & 33.5 & 39.9 & \SetCell{c, gray9}42.5 & &
    \SetCell{c, gray9}\underline{48.0}\Tstrut\\
    \textbf{I-Latte++}
    & \cmark & \cmark & - & &
    37.2 & 41.1 & \SetCell{c, gray9}\underline{46.4} & & \underline{49.7} &
    52.1 & \SetCell{c, gray9}\textbf{57.5} & & \underline{33.6} &
    \underline{40.3} & \SetCell{c, gray9}\underline{43.9} & &
    \SetCell{c, gray9}\textbf{49.3}\Tstrut\\
    \textit{Improv.$\uparrow$}& & & & & \decrease{0.4} & \improve{0.1} &
    \SetCell{c, gray9}\improve{0.1} & & \improve{3.3} & \decrease{0.9} &
    \SetCell{c, gray9}\improve{2.4} & & \improve{0.1} & \improve{0.4} &
    \SetCell{c, gray9}\improve{1.4} & &
    \SetCell{c, gray9}\improve{1.3}\Bstrut\\
    \hline
    \hline
    \end{tblr}
    }
    \label{Tab: Overall Results}
\end{table*}

\subsection{Overall Results}\label{sec:Overall Results}
\subsubsection{Results on MM-TTA benchmarks}
Table~\ref{Tab: Overall Results} presents the overall results on three benchmarks compared with previous SOTA methods. In terms of cross-modal predictions, by leveraging consistency checks from various time windows, Latte++ achieves a relative performance gap of approximately $2\%$ on U-to-S and $2.6\%$ on A-to-K compared to MMTTA, respectively. It also surpasses Latte and becomes the SOTA MM-TTA method without human interaction. On the other hand, when combined with ITTA, the performance of existing MM-TTA methods and Latte++ improves significantly, leading to a gap of more than $1.3\%$ on the average cross-modal predictions (Avg). Additionally, one can notice that ITTA+ methods claim the best performance across 8 out of 11 metrics. This justifies the benefit of introducing effortless human feedback during the online adaptation process and the effectiveness of our proposed method. Another interesting observation is that almost all cross-modal methods outperform other single-modal TTA methods, \eg, even the simplest PsLabel can outperform recent TTA methods like SAR~\citep{niu2023towards}. This reveals the non-trivial benefits of cross-modal learning for TTA with segmentation. To provide a fair comparison, we thus supplement the results with existing TTA methods combined with a widely used cross-modal learning scheme xMUDA~\citep{jaritz2022cross} as in Table~\ref{Tab: Overall Results}. Specifically, the cross-modal prediction consistency loss from~[16] is included for each TTA method as their additional optimization objective with a coefficient of 0.1. Although the cross-modal learning scheme brings an average relative improvement of 1.0-4.0\%, the performance gap between existing TTA methods and Latte++ is still non-trivial, which justifies the superiority of our cross-modal learning scheme in Latte++. Note that we do not calculate the inference time per scan for ITTA+ methods here, as it is challenging to quantify the time consumed during prompt providing. We leave this for discussion in future work.

\begin{table*}[th]
    \centering
    \setlength{\tabcolsep}{4pt}
    \caption{Cross-modal performance (mIoU) comparison on K-to-S benchmarks. Here we report the softmax-average mIoU of 2D and 3D predictions. Sequences improved or decreased by ITTA are highlighted in Digits in \textcolor{MidnightBlue}{blue} and \textcolor{Orange}{orange}, respectively.}
    \resizebox{.94\textwidth}{!}{
    \huge
    \begin{tblr}{ 
        colspec                   = {l | c c| *{12}{c} | c },
        cell{1, 26}{1}                = {c=3}{c}, 
        cell{2}{1,2,3}            = {r=2}{l}, 
        cell{1, 2, 26}{4}         = {c=13}{c}, 
        cell{15, 18, 21, 24, 28, 31, 34, 37}{2,3} = {r=2}{c} 
        }
    \hline
    \hline
    Time & & & $t \ \ \xrightarrow{\hspace*{27em}}$\\
    \hline
    Methods & Prompt & MM & SemanticKITTI-to-Synthia (K-to-S)\\
    \hline
    & & & \rotatebox[origin=c]{75}{01-Spring} &
    \rotatebox[origin=c]{75}{02-Summer} & \rotatebox[origin=c]{75}{04-Fall} &
    \rotatebox[origin=c]{75}{05-Winter} & \rotatebox[origin=c]{75}{01-Dawn} &
    \rotatebox[origin=c]{75}{02-Night} & \rotatebox[origin=c]{75}{04-Sunset} &
    \rotatebox[origin=c]{75}{05-W-night} & \rotatebox[origin=c]{75}{01-Fog} &
    \rotatebox[origin=c]{75}{02-S-Rain} & \rotatebox[origin=c]{75}{04-R-Night} &
    \rotatebox[origin=c]{75}{05-Rain} & \SetCell{c, gray9} \makecell{xM \\
    Avg}\Tstrut\\
    
    \hline
    
    Source & \xmark & \xmark &
    48.0 & 35.1 & 43.0 & 36.8 & 46.3 & 29.9 & 41.4 & 33.4 & 49.3 & 26.4 & 36.8 &
    29.4 & \SetCell{c, gray9} 38.0\Tstrut\\
    
    Oracle& \xmark & \xmark &
    50.0 & 47.0 & 56.0 & 51.6 & 54.9 & 55.1 & 58.3 & 53.8 & 56.0 & 53.0 & 57.7 &
    48.1 & \SetCell{c, gray9} 53.5\Tstrut\\
    
    \hline[dashed]
    
    TENT &  
    \xmark & \xmark &
    45.6 & 33.8 & 42.4 & 38.2 & 44.7 & 32.1 & 41.3 & 35.2 & 45.8 & 27.5 & 32.7 &
    32.3 & \SetCell{c, gray9} 37.6\Tstrut\\
    
    ETA & \xmark & \xmark &
    45.1 & 32.7 & 42.5 & 36.8 & 43.9 & 30.0 & 41.6 & 32.4 & 39.9 & 26.3 & 31.3 &
    27.4 & \SetCell{c, gray9} 35.8\\

    SAR & \xmark & \xmark &
    45.1 & 32.9 & 42.3 & 37.6 & 44.6 & 31.9 & 42.1 & 34.9 & 46.0 & 27.8 & 33.1 &
    31.4 & \SetCell{c, gray9} 37.5\Tstrut\\

    SAR-rs & \xmark & \xmark &
    45.0 & 32.9 & 42.4 & 37.2 & 44.1 & 30.7 & 41.9 & 33.6 & 44.6 & 27.1 & 32.7 &
    29.5 & \SetCell{c, gray9} 28.8\Tstrut\\

    CoTTA & \xmark & \xmark &
    \underline{47.3} & 36.4 & 46.6 & 40.7 & 41.7 & 40.5 & 49.4 & 41.4 & 45.0 &
    40.8 & 44.2 & 40.2 & \SetCell{c, gray9}42.8\Tstrut\\

    xMUDA & \xmark & \cmark &
    45.6 & 24.0 & 31.2 & 19.5 & 25.9 & 16.6 & 23.6 & 18.6 & 25.3 & 14.4 & 19.7 &
    16.0 & \SetCell{c, gray9} 23.4\Tstrut\\

    PsLabel & \xmark & \cmark &
    45.4 & 34.8 & 46.1 & 42.4 & 46.3 & 42.4 & 49.7 & 43.9 & 46.3 & 40.0 & 43.6 &
    42.5 & \SetCell{c, gray9} 43.6\Tstrut\\

    CoMAC & \xmark & \cmark &
    \textbf{47.4} & 37.2 & 47.0 & 44.3 & 46.5 & 37.2 & 48.5 & 42.2 &
    \underline{49.1} & 24.2 & 28.3 & 18.5 & \SetCell{c, gray9}39.2\Tstrut\\
    
    \hline[dashed]
    xMUDA\textsubscript{PL} & \xmark & \cmark &
    45.5 & 35.1 & 46.8 & 43.5 & 47.4 & 43.9 & 51.2 & 45.8 & 47.9 & 40.6 & 44.5 &
    \underline{43.7} & \SetCell{c, gray9} 44.7\Tstrut\\
    \textbf{I-xMUDA\textsubscript{PL}} & \cmark & \cmark &
    \SetCell{bg=bluegray}45.6 & \SetCell{bg=bluegray}35.7 &
    \SetCell{bg=bluegray}47.9 & \SetCell{bg=bluegray}45.3 &
    \SetCell{bg=bluegray}47.7 & \SetCell{bg=bluegray}45.9 &
    \SetCell{bg=bluegray}53.1 & \SetCell{bg=bluegray}47.3 &
    \SetCell{bg=orggray}47.7 & \SetCell{bg=bluegray}42.6 &
    \SetCell{bg=bluegray}46.7 & \SetCell{bg=bluegray}\textbf{45.7} & \SetCell{c,
    gray9} 45.9\Tstrut\\
    \textit{Improv.}$\uparrow$ &
    & & \improve{0.1} & \improve{0.6} & \improve{1.1} & \improve{1.8} &
    \improve{0.5} & \improve{2.0} & \improve{1.9} & \improve{1.5} &
    \decrease{0.2} & \improve{2.0} & \improve{2.2} & \improve{2.0} & \SetCell{c,
    gray9} \improve{1.2}\Bstrut\\
    \hline[dashed]

    MMTTA & \xmark & \cmark &
    46.2 & 36.4 & 48.4 & 45.7 & 47.3 & 42.5 & 52.0 & 46.6 & \textbf{49.5} & 37.7
    & 41.4 & 41.4 & \SetCell{c, gray9} 44.2\Tstrut\\
    \textbf{I-MMTTA} & \cmark & \cmark &
    \SetCell{bg=bluegray}46.4 & \SetCell{bg=bluegray}37.7 &
    \SetCell{bg=bluegray}49.9 & \SetCell{bg=bluegray}\underline{47.7} &
    \SetCell{bg=bluegray}47.5 & \SetCell{bg=bluegray}45.5 &
    \SetCell{bg=bluegray}53.8 & \SetCell{bg=bluegray}48.7 &
    \SetCell{bg=orggray}49.0 & \SetCell{bg=bluegray}\underline{39.6} &
    \SetCell{bg=bluegray}42.9 & \SetCell{bg=orggray}38.9 & \SetCell{c, gray9}
    45.6\Tstrut\\
    \textit{Improv.}$\uparrow$ &
    & & \improve{0.3} & \improve{1.8} & \improve{2.3} & \improve{1.5} &
    \improve{0.2} & \improve{3.0} & \improve{1.8} & \improve{2.1} &
    \decrease{0.5} & \improve{1.9} & \improve{1.5} & \decrease{2.5} &
    \SetCell{c, gray9} \improve{1.4}\Bstrut\\
    \hline[dashed]

    \textbf{Latte} & \xmark & \cmark &
    46.6 & 41.3 & 51.5 & 46.6 & 47.3 & 48.5 & 53.8 & 48.3 & 47.7 & 42.2 & 44.3 &
    37.2 & \SetCell{c, gray9}46.3\Tstrut\\
    \textbf{I-Latte} & \cmark & \cmark &
    \SetCell{bg=bluegray}\underline{47.3} & \SetCell{bg=bluegray}\textbf{43.3} &
    \SetCell{bg=bluegray}\underline{52.5} & \SetCell{bg=bluegray}\textbf{47.8} &
    \SetCell{bg=bluegray}\underline{48.5} & \SetCell{bg=bluegray}\textbf{51.0} &
    \SetCell{bg=bluegray}\textbf{54.9} & \SetCell{bg=bluegray}\textbf{48.9} &
    \SetCell{bg=orggray}47.2 & \SetCell{bg=bluegray}\textbf{46.8} &
    \SetCell{bg=bluegray}\underline{46.9} & \SetCell{bg=bluegray}42.0 &
    \SetCell{c, gray9} \textbf{48.1}\Tstrut\\
    \textit{Improv.}$\uparrow$ & 
    & & \improve{0.7} & \improve{2.0} & \improve{1.0} & \improve{1.2} &
    \improve{1.2} & \improve{2.5} & \improve{1.1} & \improve{0.6} &
    \decrease{0.5} & \improve{4.6} & \improve{2.6} & \improve{4.8} & \SetCell{c,
    gray9} \improve{1.8}\Bstrut\\
    \hline[dashed]

    \textbf{Latte++} & \xmark & \cmark &
    47.0 & 41.5 & 51.9 & 46.9 & 47.5 & 49.0 & 54.2 & 48.6 & 48.0 & 43.2 & 45.7 &
    38.8 & \SetCell{c, gray9}46.9\Tstrut\\
    \textbf{I-Latte++} & \cmark & \cmark &
    \SetCell{bg=bluegray}\underline{47.3} &
    \SetCell{bg=bluegray}\underline{43.2} & \SetCell{bg=bluegray}\textbf{52.6} &
    \SetCell{bg=bluegray}\textbf{47.8} & \SetCell{bg=bluegray}\textbf{48.6} &
    \SetCell{bg=bluegray}\underline{50.9} &
    \SetCell{bg=bluegray}\underline{54.4} &
    \SetCell{bg=bluegray}\underline{48.8} & \SetCell{bg=orggray}47.5 &
    \SetCell{bg=bluegray}45.7 & \SetCell{bg=bluegray}\textbf{47.3} &
    \SetCell{bg=bluegray}41.0 & \SetCell{c, gray9}\underline{47.9}\Tstrut\\
    \textit{Improv.}$\uparrow$ & 
    & & \improve{0.3} & \improve{1.7} & \improve{1.3} & \improve{0.9} &
    \improve{1.1} & \improve{1.9} & \improve{0.2} & \improve{0.2} &
    \decrease{0.5} & \improve{2.5} & \improve{1.8} & \improve{2.2} & \SetCell{c,
    gray9} \improve{1.0}\Bstrut\\

    \hline
    Time & & & $t \ \ \xleftarrow{\hspace*{27em}}$\\
    \hline
    xMUDA\textsubscript{PL} & \xmark & \cmark &
    46.3 & 43.4 & 50.7 & 44.0 & 46.2 & 39.7 & 45.9 &
    36.7 & 44.7 & 28.6 & 32.6 &
    \underline{27.2} & \SetCell{c, gray9}40.5\Tstrut\\
    \textbf{I-xMUDA\textsubscript{PL}} & \cmark & \cmark &
    \SetCell{bg=orggray}46.0 & \SetCell{bg=bluegray}45.7 & 
    \SetCell{bg=bluegray}53.0 & \SetCell{bg=bluegray}45.9 & 
    \SetCell{bg=bluegray}46.3 & \SetCell{bg=bluegray}41.6 &
    \SetCell{bg=bluegray}\underline{48.1} & \SetCell{bg=bluegray}\underline{38.9} & 
    \SetCell{bg=orggray}44.5 & \SetCell{bg=bluegray}\textbf{29.9} & 
    \SetCell{bg=bluegray}\textbf{33.1} & \SetCell{bg=bluegray}\textbf{27.3} & 
    \SetCell{c, gray9}\underline{41.7}\Tstrut\\
    \textit{Improv.}$\uparrow$ & 
    & & 
    \decrease{0.3} & \improve{2.3} & 
    \improve{2.3} & \improve{1.9} &
    \improve{0.1} & \improve{1.9} & 
    \improve{2.2} & \improve{2.2} &
    \decrease{0.2} & \improve{1.3} & 
    \improve{0.5} & \improve{0.1} & 
    \SetCell{c,gray9} \improve{1.2}\Bstrut\\

    \hline[dashed]

    MMTTA & \xmark & \cmark &
    \textbf{48.9} & 46.9 & 
    51.5 & 46.7 &
    \textbf{46.9} & 40.2 & 
    44.8 & 34.0 & 
    42.4 & 24.0 & 
    30.6 & 26.0 &
    \SetCell{c, gray9} 40.2\Tstrut\\
    \textbf{I-MMTTA} & \cmark & \cmark &
    \SetCell{bg=orggray}\underline{48.4} & \SetCell{bg=bluegray}\underline{48.7} & 
    \SetCell{bg=bluegray}52.3 & \SetCell{bg=bluegray}\textbf{49.0} & 
    \SetCell{bg=orggray}46.1 &\SetCell{bg=bluegray}43.3 &
    \SetCell{bg=bluegray}46.5 & \SetCell{bg=bluegray}36.6 & 
    \SetCell{bg=bluegray}42.4 & \SetCell{bg=bluegray}24.9 & 
    \SetCell{bg=bluegray}31.1 & \SetCell{bg=orggray}25.6 & 
    \SetCell{c, gray9}41.2\Tstrut\\
    \textit{Improv.}$\uparrow$ & 
    & & 
    \decrease{0.5} & \improve{1.8} & 
    \improve{0.8} & \improve{2.3} &
    \decrease{0.8} & \improve{3.1} & 
    \improve{1.7} & \improve{2.6} &
    \improve{0.0} & \improve{0.9} & 
    \improve{0.5} & \decrease{0.4} & 
    \SetCell{c,gray9} \improve{1.0}\Bstrut\\

    \hline[dashed]

    \textbf{Latte}& \xmark & \cmark &
    47.9 & 46.1 & 
    51.6 & 45.1 &
    46.5 & 42.7 & 
    47.1 & 38.4 &
    \underline{46.1} & 28.3 & 
    32.8 & \underline{27.2} & 
    \SetCell{c,gray9} 41.6\Tstrut\\
    \textbf{I-Latte} & \cmark & \cmark &
    \SetCell{bg=orggray}47.7 & \SetCell{bg=bluegray}\textbf{49.2} & 
    \SetCell{bg=bluegray}\textbf{53.4} & \SetCell{bg=bluegray}46.6 & 
    \SetCell{bg=bluegray}\underline{46.7} &\SetCell{bg=bluegray}\textbf{45.0} &
    \SetCell{bg=bluegray}\textbf{48.6} & \SetCell{bg=bluegray}\textbf{39.9} & 
    \SetCell{bg=orggray}46.0 & \SetCell{bg=bluegray}\underline{29.1} & 
    \SetCell{bg=bluegray}\underline{32.9} & \SetCell{bg=bluegray}\underline{27.2} & 
    \SetCell{c, gray9}\textbf{42.7}\Tstrut\\
    \textit{Improv.}$\uparrow$ & 
    & & 
    \decrease{0.2} & \improve{3.1} & 
    \improve{1.8} & \improve{1.5} &
    \improve{0.2} & \improve{2.3} & 
    \improve{1.5} & \improve{1.5} &
    \decrease{0.1} & \improve{0.8} & 
    \improve{0.1} & \improve{0.0} & 
    \SetCell{c,gray9} \improve{1.1}\Bstrut\\

    \hline[dashed]

    \textbf{Latte++}& \xmark & \cmark &
    47.9 & 46.0 & 
    51.6 & 45.1 &
    46.5 & 42.7 & 
    47.1 & 38.4 &
    \textbf{46.2} & 28.3 & 
    32.8 & \underline{27.2} & 
    \SetCell{c, gray9}41.6\Tstrut\\
    \textbf{I-Latte++} & \cmark & \cmark &
    \SetCell{bg=orggray}47.5 & \SetCell{bg=bluegray}\textbf{49.2} & 
    \SetCell{bg=bluegray}\underline{53.2} & \SetCell{bg=bluegray}\underline{46.7} & 
    \SetCell{bg=bluegray}46.5 &\SetCell{bg=bluegray}\underline{44.9} &
    \SetCell{bg=bluegray}\textbf{48.6} & \SetCell{bg=bluegray}\textbf{39.9} & 
    \SetCell{bg=orggray}46.0 & \SetCell{bg=bluegray}\underline{29.1} & 
    \SetCell{bg=bluegray}\textbf{33.1} & \SetCell{bg=orggray}27.1 & 
    \SetCell{c, gray9}\textbf{42.7}\Tstrut\\
    \textit{Improv.}$\uparrow$ & 
    & & 
    \decrease{0.4} & \improve{3.2} & 
    \improve{1.6} & \improve{1.6} &
    \improve{0.0} & \improve{2.2} & 
    \improve{1.5} & \improve{1.5} &
    \decrease{0.2} & \improve{0.8} & 
    \improve{0.3} & \decrease{0.1} & 
    \SetCell{c,gray9} \improve{1.1}\Bstrut\\
    
    \hline
    \hline
    \end{tblr}
    }
    \label{Table:K2S_xM_Results}
\end{table*}

\begin{table*}[th!]
    \centering
    \setlength{\tabcolsep}{4pt}
    \caption{Cross-modal performance (mIoU) comparison of K-to-W. Here we report the softmax-average mIoU of 2D and 3D predictions. Sequences improved or decreased by ITTA are highlighted in Digits in \textcolor{MidnightBlue}{blue} and \textcolor{Orange}{orange}, respectively.}
    \resizebox{.9\textwidth}{!}{
    \huge
    \begin{tblr}{ colspec                         = {l | c c | *{10}{c} | c},
        cell{1, 26}{1}                                = {c=3}{c},
        cell{2}{1, 2, 3}                          = {r=2}{l}, 
        cell{1, 2, 26}{4}                         = {c=11}{c}, 
        cell{15, 18, 21, 24, 28, 31, 34, 37}{2,3} = {r=2}{c} }
    \hline
    \hline
    Time & & & $t \ \ \xrightarrow{\hspace*{24em}}$\\
    \hline
    Methods & Prompt & MM & SemanticKITTI-to-Waymo (K-to-W)\\
    \hline
    & & & \rotatebox[origin=c]{75}{D-O-1} & \rotatebox[origin=c]{75}{D-P-1} &
    \rotatebox[origin=c]{75}{D-S-1} & \rotatebox[origin=c]{75}{DD-1} &
    \rotatebox[origin=c]{75}{N-1} & \rotatebox[origin=c]{75}{D-O-2} &
    \rotatebox[origin=c]{75}{D-P-2} & \rotatebox[origin=c]{75}{D-S-2} &
    \rotatebox[origin=c]{75}{DD-2} & \rotatebox[origin=c]{75}{N-2} & \SetCell{c,
    gray9} \makecell{xM \\ Avg}\Tstrut\\
    
    \hline
    
    Source & 
    \xmark & \xmark & 43.5 & 45.9 & 46.1 & 42.8 & 38.3 & 43.6 & 48.1 & 44.9 &
    45.1 & 37.1 & \SetCell{c, gray9} 43.5\Tstrut\\
    
    Oracle& 
    \xmark & \xmark & 46.0 & 49.7 & 49.0 & 47.8 & 46.3 & 51.4 & 53.3 & 49.9 &
    51.1 & 47.2 & \SetCell{c, gray9} 49.2\Tstrut\\
    
    \hline[dashed]
    
    TENT &  
    \xmark & \xmark & 43.9 & 35.7 & 42.4 & 41.2 & 39.0 & 42.1 & 48.0 & 38.9 &
    41.7 & 38.7 & \SetCell{c, gray9} 42.2\Tstrut\\
    
    ETA & 
    \xmark & \xmark & 43.9 & 46.6 & 43.3 & 42.1 & 38.5 & 44.7 & 50.2 & 42.0 &
    44.5 & 39.0 & \SetCell{c, gray9} 43.5\\

    SAR & 
    \xmark & \xmark & 44.0 & 46.5 & 43.5 & 42.0 & 39.3 & 44.4 & 50.0 & 41.6 &
    44.2 & 39.8 & \SetCell{c, gray9} 43.5\Tstrut\\

    SAR-rs & 
    \xmark & \xmark & 44.0 & 46.6 & 43.5 & 41.9 & 38.5 & 44.6 & 50.3 & 41.8 &
    44.1 & 38.4 & \SetCell{c, gray9} 43.4\Tstrut\\

    CoTTA& 
    \xmark & \xmark & \textbf{47.3} & 36.4 & 46.6 & 40.7 & 41.7 & 40.5 & 49.4 &
    41.4 & 45.0 & 40.8 & \SetCell{c, gray9} 44.2\Tstrut\\

    xMUDA & 
    \xmark & \cmark & 43.5 & 38.9 & 30.9 & 26.4 & 26.2 & 29.8 & 31.6 & 26.8 &
    28.7 & 27.1 & \SetCell{c, gray9} 31.0\Tstrut\\

    PsLabel & 
    \xmark & \cmark & 44.7 & 48.8 & 47.4 & 48.2 & 50.9 & 51.0 & 53.0 & 49.8 &
    53.0 & 53.2 & \SetCell{c, gray9} 50.0\Tstrut\\

    CoMAC& 
    \xmark & \cmark & 44.2 & 45.7 & 44.0 & 41.9 & 35.2 & 41.5 & 42.8 & 40.5 &
    40.4 & 28.2 & \SetCell{c, gray9} 40.4\Tstrut\\
    
    \hline[dashed]

    xMUDA\textsubscript{PL} & 
    \xmark & \cmark & 44.6 & 48.4 & 47.5 & 48.1 & 51.3 & 51.3 & 53.0 & 50.3 &
    53.4 & \underline{53.8} & \SetCell{c, gray9}50.2\Tstrut\\
    \textbf{I-xMUDA\textsubscript{PL}} & 
    \cmark & \cmark & \SetCell{bg=bluegray}45.3 &
    \SetCell{bg=bluegray}\underline{49.1} & \SetCell{bg=bluegray}\textbf{48.8} &
    \SetCell{bg=bluegray}\textbf{50.5} & \SetCell{bg=bluegray}\textbf{52.8} &
    \SetCell{bg=bluegray}\underline{52.5} & \SetCell{bg=bluegray}53.3 &
    \SetCell{bg=bluegray}\underline{51.9} & \SetCell{bg=bluegray}54.2 &
    \SetCell{bg=orggray}53.6 & \SetCell{c, gray9}\underline{51.2}\\
    \textit{Improv.}$\uparrow$ & & & \improve{0.7} & \improve{0.7} &
    \improve{1.3} & \improve{2.4} & \improve{1.5} & \improve{1.2} &
    \improve{0.3} & \improve{1.6} & \improve{0.8} & \decrease{0.2} & \SetCell{c,
    gray9}\improve{1.0}\\

    \hline[dashed]

    MMTTA & 
    \xmark & \cmark & 44.1 & 45.2 & 44.6 & 43.8 & 43.1 & 43.8 & 48.0 & 45.4 &
    47.4 & 47.0 & \SetCell{c, gray9} 45.2\Tstrut\\
    \textbf{I-MMTTA}& 
    \cmark & \cmark & \SetCell{bg=orggray}43.9 & \SetCell{bg=bluegray}45.4 &
    \SetCell{bg=bluegray}46.1 & \SetCell{bg=bluegray}45.8 &
    \SetCell{bg=bluegray}47.3 & \SetCell{bg=bluegray}47.3 &
    \SetCell{bg=bluegray}49.2 & \SetCell{bg=bluegray}47.5 &
    \SetCell{bg=bluegray}48.5 & \SetCell{bg=bluegray}47.4 & \SetCell{c,
    gray9}46.8\\
    \textit{Improv.}$\uparrow$ & & & \decrease{0.2} & \improve{0.7} &
    \improve{0.3} & \improve{0.4} & \improve{4.2} & \improve{3.5} &
    \improve{1.2} & \improve{2.1} & \improve{1.1} & \improve{0.6} & \SetCell{c,
    gray9}\improve{1.6}\\

    \hline[dashed]

    \textbf{Latte} & 
    \xmark & \cmark & 44.5 & 48.3 & 47.5 & 47.9 & 50.5 & 51.2 & 53.8 & 50.3 &
    53.8 & 53.7 & \SetCell{c, gray9} 50.1\Tstrut\\
    \textbf{I-Latte} & 
    \cmark & \cmark & \SetCell{bg=bluegray}45.2 & \SetCell{bg=bluegray}49.0 &
    \SetCell{bg=bluegray}47.8 & \SetCell{bg=bluegray}48.5 &
    \SetCell{bg=bluegray}52.1 & \SetCell{bg=bluegray}52.3 &
    \SetCell{bg=bluegray}\underline{54.8} & \SetCell{bg=bluegray}51.1 &
    \SetCell{bg=bluegray}\underline{54.3} & \SetCell{bg=orggray}53.4 &
    \SetCell{c, gray9}50.8\\
    \textit{Improv.}$\uparrow$ & & & \improve{0.7} & \improve{0.7} &
    \improve{0.3} & \improve{0.6} & \improve{1.6} & \improve{1.1} &
    \improve{1.0} & \improve{0.8} & \improve{0.5} & \decrease{0.3} & \SetCell{c,
    gray9}\improve{0.7}\\

    \hline[dashed]
    
    \textbf{Latte++} & 
    \xmark & \cmark & \underline{45.7} & 49.0 & 48.3 & 48.1 & 50.4 & 51.8 & 54.1
    & 50.9 & 54.1 & \textbf{54.3} & \SetCell{c, gray9}50.7\Tstrut\\
    \textbf{I-Latte++} & 
    \cmark & \cmark & \SetCell{bg=orggray}45.4 &
    \SetCell{bg=bluegray}\textbf{49.8} & \SetCell{bg=bluegray}\underline{48.7} &
    \SetCell{bg=bluegray}\underline{49.8} &
    \SetCell{bg=bluegray}\underline{52.7} & \SetCell{bg=bluegray}\textbf{53.0} &
    \SetCell{bg=bluegray}\textbf{55.1} & \SetCell{bg=bluegray}\textbf{52.2} &
    \SetCell{bg=bluegray}\textbf{55.1} & \SetCell{bg=orggray}53.4 & \SetCell{c,
    gray9}\textbf{51.5}\\
    \textit{Improv.}$\uparrow$ & & & \decrease{0.3} & \improve{0.8} &
    \improve{0.4} & \improve{1.7} & \improve{2.3} & \improve{1.2} &
    \improve{1.0} & \improve{1.3} & \improve{1.0} & \decrease{0.9} & \SetCell{c,
    gray9}\improve{0.8}\\

    \hline

    Time & & & $t \ \ \xleftarrow{\hspace*{24em}}$\\

    \hline

    xMUDA\textsubscript{PL} & 
    \xmark & \cmark & 
    48.2 & \underline{49.7} & 
    49.9 & 48.6 & 
    50.9 & 48.9
    & 51.1 & 43.6 & 
    44.8 & 38.8 & 
    \SetCell{c, gray9} 47.4\Tstrut\\
    \textbf{I-xMUDA\textsubscript{PL}} & 
    \cmark & \cmark & 
    \SetCell{bg=orggray}47.5 & \SetCell{bg=orggray}49.3 &
    \SetCell{bg=bluegray}\underline{51.6} & \SetCell{bg=bluegray}\textbf{50.3} & 
    \SetCell{bg=bluegray}\textbf{53.6} & \SetCell{bg=bluegray}\textbf{50.7} & 
    \SetCell{bg=bluegray}51.9 & \SetCell{bg=bluegray}\textbf{45.7} &
    \SetCell{bg=bluegray}\textbf{46.5} & \SetCell{bg=bluegray}\textbf{40.2} & 
    \SetCell{c, gray9}\textbf{48.7}\\
    \textit{Improv.}$\uparrow$ & & & \decrease{0.7} & \decrease{0.4} &
    \improve{1.7} & \improve{1.7} & \improve{2.7} & \improve{1.8} &
    \improve{0.8} & \improve{2.1} & \improve{1.7} & \improve{1.4} & \SetCell{c,
    gray9}\improve{1.3}\\

    \hline[dashed]

    MMTTA & 
    \xmark & \cmark &
    43.4 & 44.6 &
    44.0 & 40.2 &
    39.9 & 43.8 & 
    47.0 & 40.2 & 
    39.3 & 37.4 & 
    \SetCell{c, gray9}42.0\Tstrut\\
    \textbf{I-MMTTA} & 
    \cmark & \cmark &
    \SetCell{bg=bluegray}44.3 & \SetCell{bg=bluegray}45.6 &
    \SetCell{bg=bluegray}45.9 & \SetCell{bg=bluegray}42.5 & 
    \SetCell{bg=bluegray}42.4 & \SetCell{bg=bluegray}43.9 & 
    \SetCell{bg=bluegray}47.2 & \SetCell{bg=bluegray}40.2 &
    \SetCell{bg=orggray}37.3 & \SetCell{bg=orggray}35.7 & 
    \SetCell{c, gray9}42.5\\
    \textit{Improv.}$\uparrow$ 
    & & & 
    \improve{0.9} & \improve{1.0} &
    \improve{1.9} & \improve{2.3} & 
    \improve{2.5} & \improve{0.1} &
    \improve{0.2} & \improve{0.0} & 
    \decrease{2.0} & \decrease{1.7} & 
    \SetCell{c,gray9}\improve{0.5}\\

    \hline[dashed]

    \textbf{Latte}& 
    \xmark & \cmark & 
    \underline{48.7} & \textbf{49.9} & 
    50.2 & 48.4 & 
    50.6 & 49.2 & 
    51.7 & 44.0 &
    45.0 & 39.6 & \SetCell{c, gray9}\underline{47.7}\Tstrut\\
    \textbf{I-Latte} & 
    \cmark & \cmark & 
    \SetCell{bg=orggray}48.4 & \SetCell{bg=orggray}49.5 &
    \SetCell{bg=bluegray}\textbf{51.7} & \SetCell{bg=bluegray}\textbf{50.3} &
    \SetCell{bg=bluegray}\underline{53.2} & \SetCell{bg=bluegray}\textbf{50.7} &
    \SetCell{bg=bluegray}\underline{52.0} & \SetCell{bg=bluegray}45.3 &
    \SetCell{bg=bluegray}46.2 & \SetCell{bg=bluegray}40.1 & 
    \SetCell{c,gray9}\textbf{48.7}\\
    \textit{Improv.}$\uparrow$ & & & \decrease{0.3} & \decrease{0.4} &
    \improve{1.5} & \improve{1.9} & \improve{2.6} & \improve{1.5} &
    \improve{0.3} & \improve{1.3} & \improve{1.2} & \improve{0.5} & \SetCell{c,
    gray9}\improve{1.1}\\

    \hline[dashed]

    \textbf{Latte++}& 
    \xmark & \cmark & 
    \textbf{49.1} & \underline{49.7} & 
    50.4 & 48.6 & 
    50.3 & 49.2 & 
    51.2 & 43.8 &
    45.2 & 39.7 & \SetCell{c, gray9}\underline{47.7}\Tstrut\\
    \textbf{I-Latte++} & 
    \cmark & \cmark & 
    \SetCell{bg=orggray}48.3 & \SetCell{bg=orggray}49.5 &
    \SetCell{bg=bluegray}\textbf{51.7} & \SetCell{bg=bluegray}\underline{50.0} &
    \SetCell{bg=bluegray}52.7 & \SetCell{bg=bluegray}\underline{50.3} &
    \SetCell{bg=bluegray}\textbf{52.3 }& \SetCell{bg=bluegray}\underline{45.5} &
    \SetCell{bg=bluegray}\underline{46.4} & \SetCell{bg=bluegray}\underline{40.1} & 
    \SetCell{c,gray9}\textbf{48.7}\\
    \textit{Improv.}$\uparrow$ & & & \decrease{0.8} & \decrease{0.2} &
    \improve{1.3} & \improve{1.4} & \improve{2.4} & \improve{1.1} &
    \improve{1.1} & \improve{1.7} & \improve{1.2} & \improve{0.4} & \SetCell{c,
    gray9}\improve{1.0}\\
    
    \hline
    \hline
    \end{tblr}
    }
    \label{Table:K2W_xM_Results}
\end{table*}

\begin{figure*}[t]
    \subfloat{\includegraphics[width=.47\textwidth]{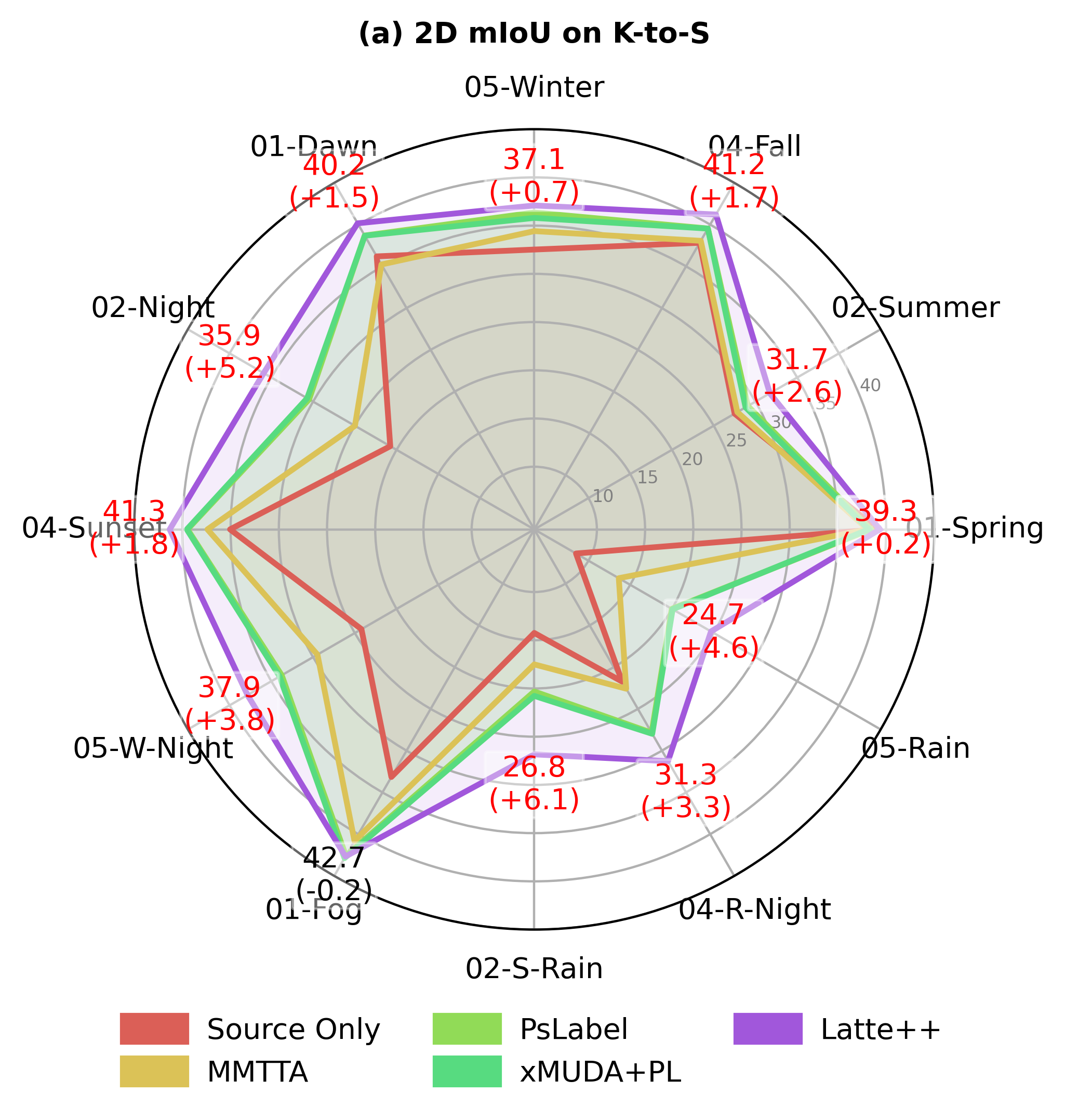}\label{fig: LP_K2S_2D}}
    \subfloat{\includegraphics[width=.47\textwidth]{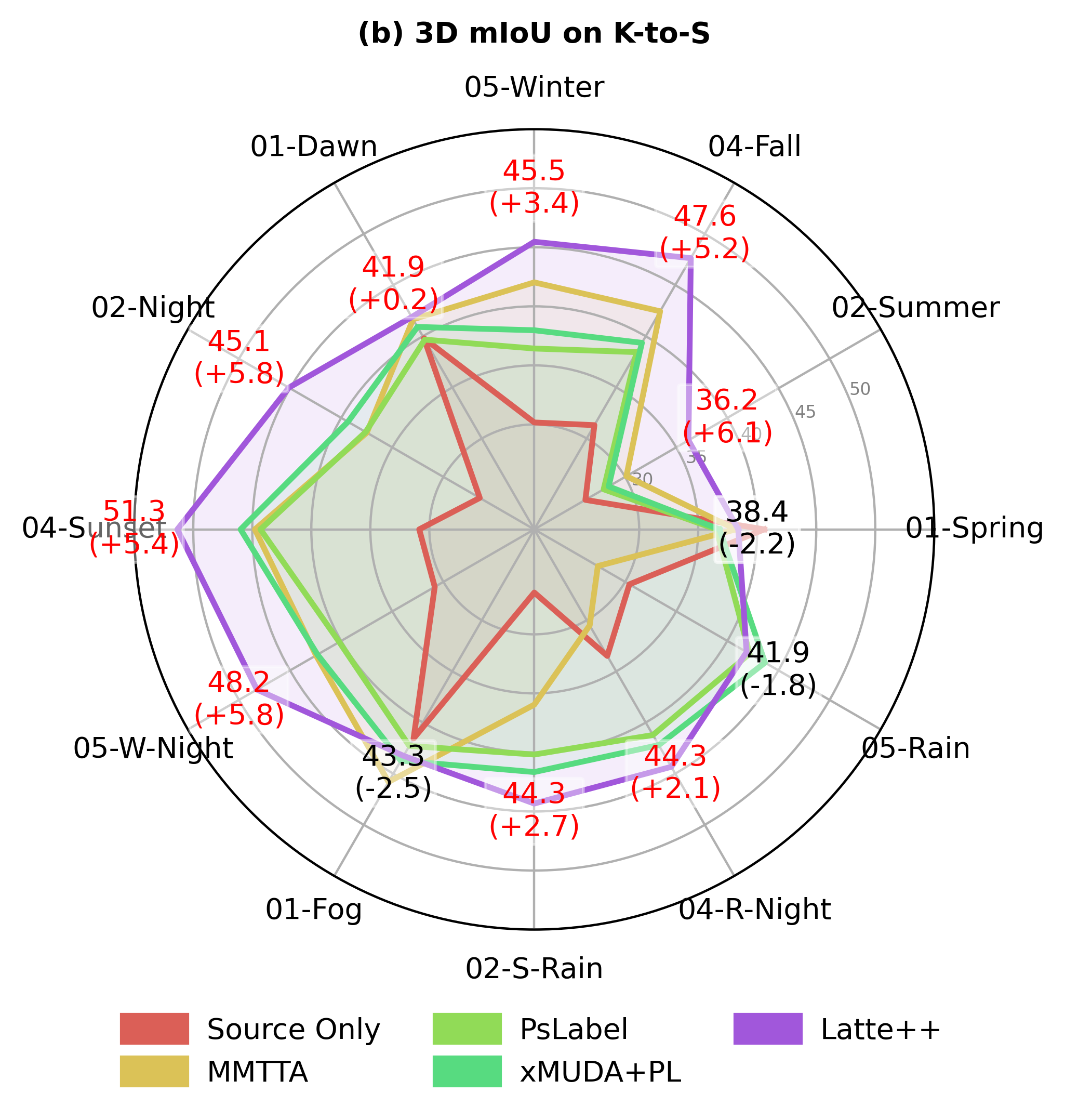}\label{fig: LP_K2S_3D}}\\
    \subfloat{\includegraphics[width=.45\textwidth]{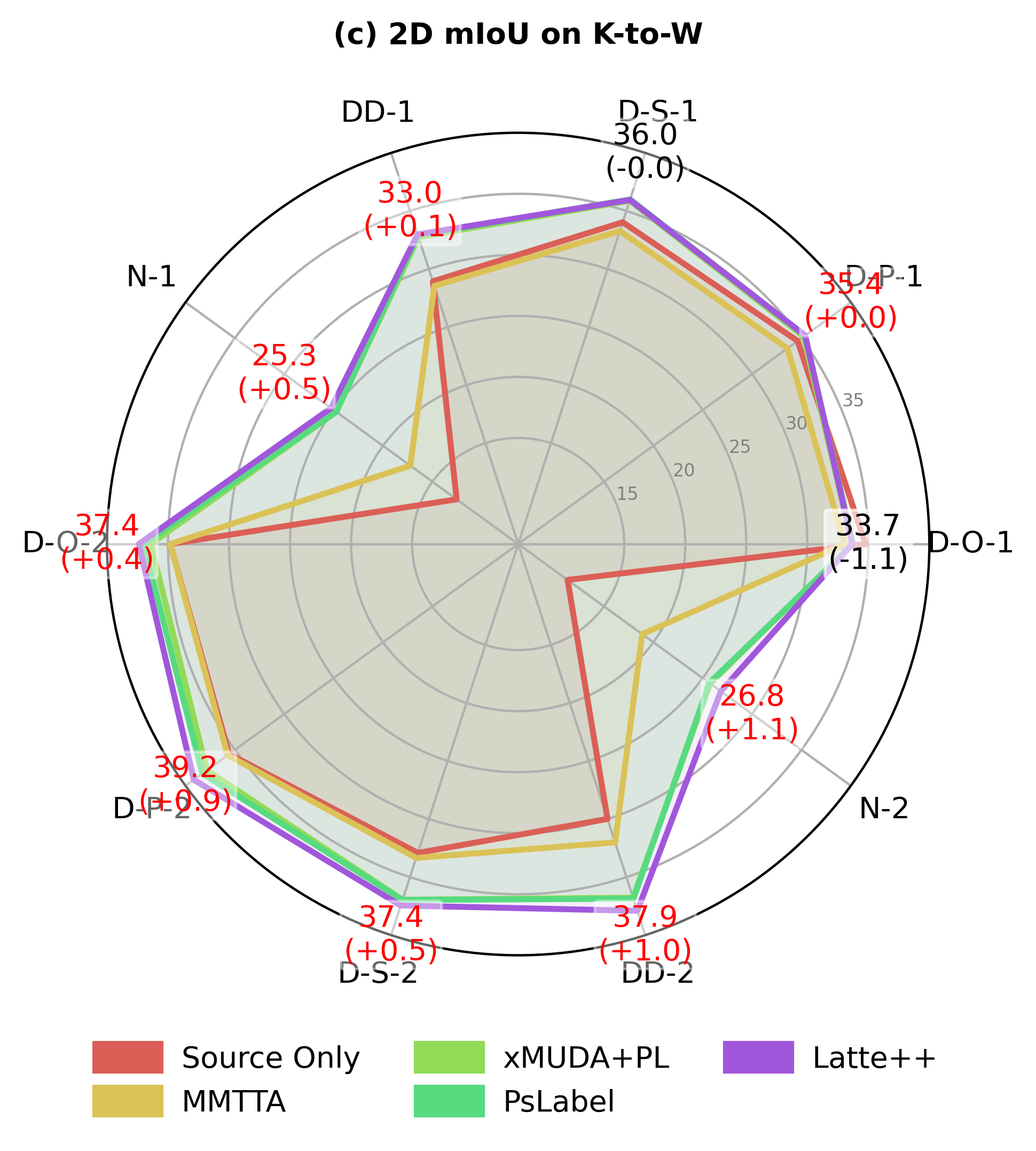}\label{fig: LP_K2W_2D}}
    \subfloat{\includegraphics[width=.45\textwidth]{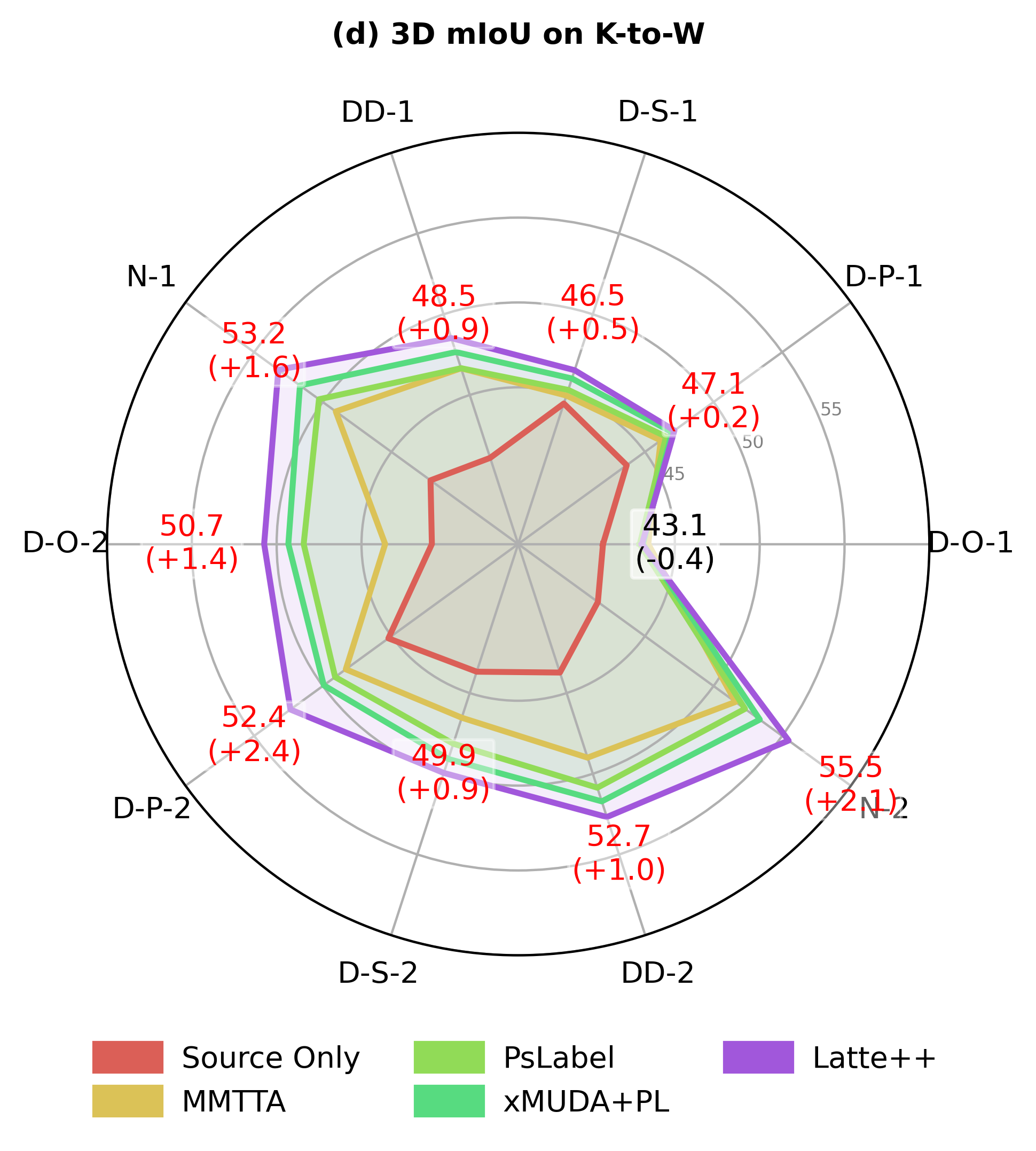}\label{fig: LP_K2W_3D}}
    \caption{Single modal predictions comparison (mIoU) on K-to-S and K-to-W. We report the performance of Latte++ for each sequence (highlighted in \textcolor{red}{red} if any improvement), followed by the absolute gaps compared to the previous sequence-wise second highest performance in the round bracket.}\label{fig: MM-CTTA_single_modal}
    \hspace{-10pt}
\end{figure*}

\subsubsection{Results on MM-CTTA benchmarks}
We further evaluate the performance of Latte++ and ITTA on MM-CTTA benchmarks with continuously changing target domains. As shown in Table~\ref{Table:K2S_xM_Results}, when testing on K-to-S, which initially possesses a more challenging synthetic-to-real domain discrepancy, both Latte and Latte++ yield noticeable relative improvements of $3.6\%$ and $4.9\%$ on the average cross-modal segmentation performance compared to the existing SOTA method xMUDA\textsubscript{PL}, respectively. For sequence-wise performance, they outperform the single-scan refinement method MMTTA by at least $5.0\%$ on 8/12 sequences. When evaluated on the less challenging K-to-W as in Table~\ref{Table:K2W_xM_Results}, Latte performs competitively with a trivial relative gap of less than $0.2\%$ compared to xMUDA\textsubscript{PL}, while Latte++ achieves SOTA performance of $50.7\%$ by leveraging the more informative multi-window aggregation. Similar to K-to-S, both Latte and Latte++ surpass the sequence-wise performance of MMTTA on all sequences in K-to-W, leading to a significant overall improvement of $5.5\%$. As for ITTA, all four MM-TTA methods achieve an average improvement of about $1.0\%$ on K-to-S. For all tested baseline methods, introducing ITTA brings consistent sequence-wise improvement on more than 10/12 sequences. When testing on K-to-W, a similar consistent gap can be observed, where ITTA brings improvements on more than 8 out of 10 sequences for all MM-TTA methods. Following the experimental settings in CoMAC~\cite{Cao_2023_ICCV}, we conduct additional runs with Latte/Latte++ and existing SOTA methods (xMUDA\textsubscript{PL}, MMTTA, PsLabel) by reversing the sequence playback order of these two benchmarks for a more thorough comparison. 

\begin{table*}[t]
    \centering
    \setlength{\tabcolsep}{4pt}
    \caption{Class-wise performance (mIoU) comparison on S-to-K. Accuracy improved or decreased by ITTA is highlighted in Digits in \textcolor{MidnightBlue}{blue} and \textcolor{Orange}{orange}, respectively.}
    \resizebox{.87\linewidth}{!}{
    \huge
    \begin{tblr}{ colspec              = {c| l | *{9}{c} | c }, cell{2, 5, 8,
        11}{1} = {r=3}{c} }
    \hline
    \hline
    Base TTA & Method & \rotatebox[origin=c]{90}{Car} &
    \rotatebox[origin=c]{90}{Bike} & \rotatebox[origin=c]{90}{Person} &
    \rotatebox[origin=c]{90}{Road} & \rotatebox[origin=c]{90}{Sidewalk} &
    \rotatebox[origin=c]{90}{Building} & \rotatebox[origin=c]{90}{Nature} &
    \rotatebox[origin=c]{90}{Pole} & \rotatebox[origin=c]{90}{Others} &
    \SetCell{c, gray9} Avg \\
    
    \hline

    $\mathrm{xMUDA_{PL}}$ & Base & 70.6 & 26.1 &  1.8 & 75.7 & 38.8 & 39.4 &
    46.0 & 27.7 & 0.6 & \SetCell{c, gray9}36.3\\
    & \textbf{I-Base} & \SetCell{bg=bluegray}74.3 & \SetCell{bg=bluegray}27.7 &
    \SetCell{bg=bluegray}8.3 & \SetCell{bg=bluegray}78.3 &
    \SetCell{bg=bluegray}40.1 & \SetCell{bg=bluegray}41.7 &
    \SetCell{bg=bluegray}49.9 & \SetCell{bg=orggray}26.1 &
    \SetCell{bg=bluegray}0.9 & \SetCell{c, gray9}38.6\\
    & \textit{Improve.}$\uparrow$ & \improve{3.7} & \improve{1.6} &
    \improve{6.5} & \improve{2.6} & \improve{1.3} & \improve{2.3} &
    \improve{3.9} & \decrease{1.6} & \improve{0.3} & \SetCell{c,
    gray9}\improve{2.3}\\
    
    \hline
    
    MMTTA & Base & 66.0 & 24.5 & 9.3 & 52.2 & 26.4 & 56.3 & 59.4 & 25.5 & 0.2 &
    \SetCell{c, gray9} 35.5\\
    & \textbf{I-Base} & \SetCell{bg=bluegray}67.6 & \SetCell{bg=bluegray}31.9 &
    \SetCell{bg=bluegray}10.0 & \SetCell{bg=bluegray}54.3 &
    \SetCell{bg=bluegray}28.0 & \SetCell{bg=bluegray}57.5 &
    \SetCell{bg=bluegray}61.7 & \SetCell{bg=bluegray}27.5 &
    \SetCell{bg=bluegray}0.2 & \SetCell{c, gray9}37.6\\
    & \textit{Improve.}$\uparrow$ & \improve{1.6} & \improve{7.4} &
    \improve{0.7} & \improve{1,1} & \improve{1.6} & \improve{1.2} &
    \improve{2.3} & \improve{2.0} & \improve{0.0} & \SetCell{c,
    gray9}\improve{2.1}\\

    \hline
    
    Latte & Base & 71.8 & 27.2 & 11.1 & 76.9 & 39.0 & 55.8 & 62.0 & 31.0 & 0.2 &
    \SetCell{c, gray9}41.7\\
    & \textbf{I-Base} & \SetCell{bg=bluegray}73.2 & \SetCell{bg=bluegray}38.8 &
    \SetCell{bg=bluegray}15.6 & \SetCell{bg=bluegray}77.4 &
    \SetCell{bg=bluegray}39.3 & \SetCell{bg=bluegray}56.5 &
    \SetCell{bg=bluegray}62.7 & \SetCell{bg=bluegray}32.4 &
    \SetCell{bg=bluegray}0.2 & \SetCell{c, gray9}\textbf{44.0}\\
    & \textit{Improve.}$\uparrow$ & \improve{1.4} & \improve{11.6} &
    \improve{4.5} & \improve{0.5} & \improve{0.3} & \improve{0.7} &
    \improve{0.7} & \improve{1.4} & \improve{0.2} & \SetCell{c,
    gray9}\improve{2.3}\\

    \hline

    Latte++ & Base & 73.8 & 28.0 & 12.1 & 78.2 & 39.7 & 56.4 & 62.5 & 31.7 & 0.2
    & \SetCell{c, gray9}42.5\\
    & \textbf{I-Base} & \SetCell{bg=orggray}73.1 & \SetCell{bg=bluegray}38.1 &
    \SetCell{bg=bluegray}15.5 & \SetCell{bg=orggray}77.4 &
    \SetCell{bg=orggray}39.0 & \SetCell{bg=bluegray}56.5 &
    \SetCell{bg=bluegray}62.5 & \SetCell{bg=bluegray}32.6 &
    \SetCell{bg=bluegray}0.2 & \SetCell{c, gray9}\underline{43.9}\\
    & \textit{Improve.}$\uparrow$ & \decrease{0.7} & \improve{10.1} &
    \improve{3.4} & \decrease{0.8} & \decrease{0.7} & \improve{0.1} &
    \improve{0.0} & \improve{0.9} & \improve{0.0} & \SetCell{c,
    gray9}\improve{1.4}\\
    
    \hline
    \hline
    \end{tblr}
}\label{Tab: S2K_classwise}
\end{table*}

\begin{table*}[t]
    \centering
    \setlength{\tabcolsep}{4pt}
    \caption{Class-wise performance (mIoU) comparison on A-to-K. Accuracy improved or decreased by ITTA is highlighted in Digits in \textcolor{MidnightBlue}{blue} and \textcolor{Orange}{orange}, respectively.}
    \resizebox{.95\linewidth}{!}{
    \huge
    \begin{tblr}{ colspec              = {c| l | *{10}{c} | c }, cell{2, 5, 8,
        11}{1} = {r=3}{c} }
    \hline
    \hline
    Base TTA & Spec. & \rotatebox[origin=c]{90}{Car} &
    \rotatebox[origin=c]{90}{Truck} & \rotatebox[origin=c]{90}{Bike} &
    \rotatebox[origin=c]{90}{Person} & \rotatebox[origin=c]{90}{Road} &
    \rotatebox[origin=c]{90}{Parking} & \rotatebox[origin=c]{90}{Sidewalk} &
    \rotatebox[origin=c]{90}{Building} & \rotatebox[origin=c]{90}{Nature} &
    \rotatebox[origin=c]{90}{Others} & \SetCell{c, gray9} Avg \\

    \hline
    
    $\mathrm{xMUDA_{PL}}$ & Base & 82.4 & 30.5 & 49.2 & 36.2 & 90.7 & 11.3 &
    63.4 & 56.5 & 62.0 & 27.2 & \SetCell{c, gray9}50.9\\

    & \textbf{I-Base} & \SetCell{bg=bluegray}82.7 & \SetCell{bg=bluegray}38.1 &
    \SetCell{bg=bluegray}51.3 & \SetCell{bg=bluegray}41.9 &
    \SetCell{bg=bluegray}90.9 & \SetCell{bg=bluegray}14.0 &
    \SetCell{bg=bluegray}63.6 & \SetCell{bg=bluegray}58.7 &
    \SetCell{bg=orggray}57.1 & \SetCell{bg=orggray}26.8 & \SetCell{c,
    gray9}52.5\\

    & \textit{Improv.}$\uparrow$ & \improve{0.3} & \improve{7.6} & \improve{2.1}
    & \improve{5.7} & \improve{0.1} & \improve{2.7} & \improve{0.2} &
    \improve{2.2} & \decrease{4.9} & \decrease{0.4} & \SetCell{c,
    gray9}\improve{1.6} \\

    \hline
    
    MMTTA & Base & 78.1 & 45.3 & 47.2 & 40.7 & 90.0 &  6.1 & 63.6 & 57.3 & 80.7
    & 27.9 & \SetCell{c, gray9}53.7\\

    & \textbf{I-Base} & \SetCell{bg=bluegray}80.1 & \SetCell{bg=bluegray}47.9 &
    \SetCell{bg=bluegray}52.1 & \SetCell{bg=bluegray}48.0 &
    \SetCell{bg=bluegray}90.9 &  \SetCell{bg=bluegray}7.4 &
    \SetCell{bg=bluegray}65.3 & \SetCell{bg=bluegray}59.7 &
    \SetCell{bg=bluegray}82.2 & \SetCell{bg=bluegray}31.4 & \SetCell{c,
    gray9}\underline{56.5}\\

    & \textit{Improv.}$\uparrow$ & \improve{2.0} & \improve{2.6} & \improve{4.9}
    & \improve{7.3} & \improve{0.9} & \improve{1.3} & \improve{1.7} &
    \improve{2.4} & \improve{1.5} & \improve{3.5} & \SetCell{c,
    gray9}\improve{1.6} \\

    \hline
    
    Latte & Base & 80.1 & 43.7 & 48.1 & 36.7 & 90.0 &  9.1 & 63.0 & 59.2 & 77.4
    & 34.3 & \SetCell{c, gray9}54.2\\

    & \textbf{I-Base} & \SetCell{bg=bluegray}81.5 & \SetCell{bg=bluegray}48.8 &
    \SetCell{bg=bluegray}50.5 & \SetCell{bg=bluegray}46.8 &
    \SetCell{bg=bluegray}90.8 & \SetCell{bg=bluegray}13.7 &
    \SetCell{bg=bluegray}65.4 & \SetCell{bg=bluegray}60.3 &
    \SetCell{bg=bluegray}80.3 & \SetCell{bg=bluegray}36.9 & \SetCell{c,
    gray9}\textbf{57.5}\\

    & \textit{Improv.}$\uparrow$ & \improve{1.4} & \improve{5.1} & \improve{2.4}
    & \improve{10.1} & \improve{0.8} & \improve{4.6} & \improve{2.4} &
    \improve{1.1} & \improve{2.9} & \improve{2.6} & \SetCell{c,
    gray9}\improve{3.3} \\

    \hline
    
    Latte++ & Base & 80.4 & 44.8 & 48.9 & 39.1 & 90.1 &  9.3 & 63.4 & 59.8 &
    79.3 & 35.6 & \SetCell{c, gray9}55.1\\

    & \textbf{I-Base} & \SetCell{bg=bluegray}81.5 & \SetCell{bg=bluegray}48.7 &
    \SetCell{bg=bluegray}51.0 & \SetCell{bg=bluegray}47.2 &
    \SetCell{bg=bluegray}90.7 & \SetCell{bg=bluegray}13.6 &
    \SetCell{bg=bluegray}65.2 & \SetCell{bg=bluegray}60.3 &
    \SetCell{bg=bluegray}79.9 & \SetCell{bg=bluegray}36.8 & \SetCell{c,
    gray9}\textbf{57.5}\\

    & \textit{Improv.}$\uparrow$ & \improve{1.1} & \improve{3.9} & \improve{2.1}
    & \improve{8.1} & \improve{0.6} & \improve{4.3} & \improve{1.8} &
    \improve{0.5} & \improve{0.6} & \improve{1.2} & \SetCell{c,
    gray9}\improve{2.4} \\
    
    \hline
    \hline
    \end{tblr}
}\label{Tab: A2k_classwise}
\end{table*}

Despite the non-trivial universal performance degeneration caused by the more challenging setting (\ie, TTA begins with the sequences with large initial domain gaps, such as 05-Rain and N-2), the gap between Latte/Latte+ and the existing SOTA TTA/MMTTA methods remains consistent on both benchmarks, which further justifies the effectiveness of our method. It is worth noting that the performance gap between Latte and Latte++ becomes rather trivial when reversing the sequence order, mainly because there exists an increasing number of consistently incorrect predictions due to the large initial domain discrepancy. In this case, utilizing multiple time windows has a limited contribution to stabilizing the intra-modal prediction noise, which leads to similar performance of Latte and Latte++ when facing significant initial domain gaps in MM-CTTA. 

\subsection{Ablation Studies and Visualziation}\label{sec:Ablation Studies}
In this section, we investigate the effectiveness of each component in Latte and Latte++ by thorough ablation studies and qualitative results. 

\subsubsection{MM-CTTA single-modal predictions}
Besides cross-modal performance, single-modal predictions are also crucial metrics that reflect whether models from different modalities effectively amplify each other. To justify that Latte++ can achieve better cross-modal learning compared to others, we present the sequence-wise single-modal prediction comparison with SOTA methods for both MM-CTTA benchmarks. As shown in Fig.~\ref{fig: MM-CTTA_single_modal}, Latte++ prevails on most sequences in terms of either 2D (11/12 sequences) or 3D (9/12 sequences) predictions on the challenging K-to-S with a relative improvement of $1.1\%$ and $1.6\%$ on average 2D and 3D performance, respectively. On less challenging K-to-W, although the performance gap becomes narrower compared to K-to-S, a consistent sequence-wise improvement can still be observed on Latte++ in either 2D or 3D performance, which justifies that Latte++ achieves better cross-modal attending compared to existing methods.

\subsubsection{MM-TTA class-wise predictions with ITTA}
One of our main arguments is that utilizing ITTA can alleviate the error accumulation on imbalanced classes. To validate this argument, we provide the class-wise performance comparison between four MM-TTA methods with or without ITTA on A-to-K (Table~\ref{Tab: A2k_classwise}) and S-to-K (Table~\ref{Tab: S2K_classwise}). As shown in Table~\ref{Tab: A2k_classwise}, the introduction of ITTA brings significant improvements on the class of interest on all MM-TTA methods, leading to class-wise enhancements of more than $2.1\%$ and $5.7\%$ on ``Bike'' and ``Person'', respectively. A similar observation can be found on S-to-K, where the largest class-wise improvement extends to more than $10\%$. Furthermore, ITTA not only benefits the classes of interest but also enhances the discriminability of others, which demonstrates the effectiveness of introducing human feedback in MM-TTA. In terms of Latte++, it improves the performance of almost all semantic classes on both benchmarks compared to Latte, which demonstrates the superiority of multi-window aggregation.

\subsubsection{Effectiveness of cross-modal learning}
The main optimization objectives of Latte contain three parts, including ST-entropy-based cross-modal consistency (both 2D-to-3D and 3D-to-2D) and cross-entropy loss between predictions and cross-modal pseudo-labels. We justify the effectiveness by using different combinations of these components as in Table~\ref{tab:Component Effectiveness}. As shown from experiments No. 1 to No. 6, disabling any part of Latte causes performance decreases of more than $1.6\%$ relatively on the cross-modal prediction (xM), which proves the effectiveness of each component we included. A special case is No. 5 on U-to-S, where disabling ST $\mathcal{L}^{\textrm{xM}}_{23}$ only leads to a minor $0.1\%$ drop in accuracy. This is mainly because the 2D performance is already close to saturation ($37.4$ vs. $38.7$ of Oracle TTA) without the guidance of 3D predictions. In this case, introducing 3D information to 2D is less effective compared to the other benchmarks. 

For ITTA, our ablation studies are conducted with Latte (I-Latte) as shown in Table~\ref{tab: MG_Component Effectiveness}. It can be concluded that our momentum grad plays an essential role in capturing useful knowledge from human feedback, whose absence causes more than $1\%$ drop on A-to-K and S-to-K. The proposed objectives also contribute to the performance gain to a certain extent. Another core design of ours is the disentanglement of momentum gradients. While using simple summation instead has a limited impact on U-to-S and A-to-K, it causes a noticeable decrease in accuracy on S-to-K, probably because the gradient direction on S-to-K is less stable between iterations compared to others, while using our disentanglement can effectively mitigate this instability.

\begin{table*}[t]
    \centering
    \huge
    \setlength{\tabcolsep}{4pt}
    \caption{Ablation studies about the effectiveness of cross-modal learning. When disabling $\boldrm{y}^{\textrm{xM}}$, we utilize the average Softmax of all modalities as the pseudo-labels as a replacement. In experiment No. 8, we replace the ST entropy (ST ety) with point-wise entropy in Equation~(\ref{Eq: Pseudo-Labels}) and set $w_{\textrm{v}}^{\textrm{m}}$ in Equation~(\ref{Eq: ST xM Cons}) to 0.5. Here $\mathcal{L}^{\textrm{xM}}_{32}$ and $\mathcal{L}^{\textrm{xM}}_{23}$ refer to the former half and the latter half in Equation~\ref{Eq: ST xM Cons}, respectively. For Latte++, $\dag$ and $\ddag$ indicate using $\tau_d^{\mathrm{m}}$ for entropy and reference prediction weighting as in Equation~\ref{Eq: multi-window init weights}, respectively.}
    \resizebox{\textwidth}{!}{
    \begin{tblr}{ 
        colspec             = {c | c | c | *{6}{c} | *{11}{c}},
        cell{1}{1-9}        = {r=2}{c}, 
        cell{4}{1}          = {r=9}{c},
        cell{13}{1}         = {r=5}{c},
        cell{1}{13, 17}     = {r=2}{c},
        cell{1}{10, 14, 18} = {c=3}{c} 
    }
    \hline
    \hline
    Method & No. & Description & 
    $\boldrm{y}^{\textrm{xM}}$ & 
    $\mathcal{L}^{\textrm{xM}}_{23}$ &
    $\mathcal{L}^{\textrm{xM}}_{32}$ & 
    $\tau^{\mathrm{b}}_{d}$&
    ${\tau^{\mathrm{m}}_{d}}_\dag$&
    ${\tau^{\mathrm{m}}_{d}}_\ddag$&
    U-to-S & & & & 
    A-to-K & & & &
    S-to-K & & \\
    \hline
    & & & & & & & & & 2D & 3D & xM & & 2D & 3D & xM & & 2D & 3D & xM\\
    \hline
    - & 0 & Source only & 
    -      & -      & -      & -      & -      & -           
    & 31.4 & 41.1 & 43.9 & & 47.4 & 17.9 & 44.3 & & 23.4 & 36.4 & 38.2\Tstrut\\
    \hline
    \rotatebox[origin=c]{90}{Latte} & 1 & only ST ety label &
    \cmark &        &        &        &        &     
    & 36.8 & 34.5 & 43.7 & & 43.3 & 42.2 & 49.9 & & 22.4 & 30.6 & 33.3\Tstrut\\
    & 2 & only ST 2D$\rightarrow$3D &
           & \cmark &        &        &        &          
    & 32.5 & 34.5 & 41.0 & & 41.7 & 42.2 & 48.2 & & 27.7 & 30.6 & 32.9  \\
    
    & 3 & only ST 3D$\rightarrow$2D &
           &        & \cmark &        &        & 
    & 36.8 & 39.6 & 42.6 & & 43.3 & 50.2 & 48.4 & & 22.4 & 26.7 & 25.6  \\
    
    \cline{2-20}
    
    & 4 & ST ety label + ST 2D$\rightarrow$3D &
    \cmark & \cmark &         &        &        &
    & \textbf{37.5} & 35.8 & 41.0 & & 43.6 & 47.5 & 49.8 & & 31.0 & 36.0 & 38.4\Tstrut\\
    
    & 5 & ST ety label + ST 3D$\rightarrow$2D &
    \cmark &        & \cmark  &        &        &  
    & 37.4 & 40.9 & 45.9 & & \textbf{46.2} & 52.1 & 53.4 & & 32.5 & 38.7 & 39.7  \\
    
    & 6 & ST 2D$\leftrightarrow$3D &
           & \cmark & \cmark  &        &        & 
    & 36.8 & 40.3 & 43.8 & & 44.0 & 47.0 & 51.2 & & 30.4 & 35.6 & 37.7  \\
    
    \cline{2-20}
    
    & 7 & w/o percentile in Equation~(8) &
    \cmark & \cmark & \cmark &        &        &
    & 37.4 & 40.3 & 45.4 & & 45.7 & 51.4 & 53.2 & & 32.9 & 39.0 & \textbf{41.7}\Tstrut\\
    
    & 8 & Point ety vs. ST ety & 
    \cmark & \cmark & \cmark &        &        &
    & 37.0 & 40.6 & 45.1 & & 45.1 & 51.4 & 52.7 & & 31.6 & 36.7 & 38.2  \\
    
    \cline{2-20}
    
    & 9 & Latte &
    \cmark & \cmark & \cmark &        &        &
    & 37.4 & \textbf{41.0} & \textbf{46.0} & & 46.1 & \textbf{52.6} & \textbf{54.3} & & \textbf{33.2} & \textbf{39.3} & 41.6\Tstrut\\
    
    \hline

    \rotatebox[origin=c]{90}{Latte++} & 10 & Disable $\tau^{\mathrm{b}}_{d}$ in Equation~(\ref{Eq: multi-window init weights}) 
    & \cmark & \cmark & \cmark &        & \cmark & \cmark
    & 37.5 & 41.0 & 46.3 & & 46.2 & 52.8 & 54.8 & & 33.2 & 39.6 & 42.1\Tstrut\Bstrut\\
    
    \cline{2-20}
    
    & 11 & Disable $\tau_d^\mathrm{m}$ for $E^\mathrm{m}_{i,k}$ in Equation~(\ref{Eq: multi-window init weights}) 
    & \cmark & \cmark & \cmark & \cmark &        & \cmark
    & 37.4 & 40.8 & 46.1 & & 46.1 & 52.7 & 54.7 & & 33.1 & 39.5 & 41.9\Tstrut\\
    
    & 12 & Disable $\tau_d^\mathrm{m}$ for $\bar{\boldrm{p}}^\mathrm{m}_{\mathrm{r}}$ in Equation~(\ref{Eq: multi-window init weights}) 
    & \cmark & \cmark & \cmark & \cmark & \cmark &
    & 37.3 & 40.9 & 46.0 & & 46.2 & 52.7 & 54.8 & & 33.3 & 39.4 & 41.8\\
    
    & 13 & Disable $\tau_d^\mathrm{m}$ for both in Equation~(\ref{Eq: multi-window init weights}) 
    & \cmark & \cmark & \cmark & \cmark &        &
    & 37.3 & 40.6 & 45.8 & & 46.1 & 52.4 & 54.5 & & 32.8 & 39.1 & 41.5\\
    \cline{2-20}

    & 14 & Latte++ 
    & \cmark & \cmark & \cmark & \cmark & \cmark & \cmark
    & \textbf{37.6} & \textbf{41.0} & \textbf{46.3} & & \textbf{46.4} & \textbf{53.0} & \textbf{55.1} & & \textbf{33.5} & \textbf{39.9} & \textbf{42.5}\\
    
    \hline
    \hline
    \end{tblr}
    }
    \label{tab:Component Effectiveness}
\end{table*}

\begin{table*}[t]
    \centering
    \huge
    \setlength{\tabcolsep}{8pt}
    \caption{Ablation studies about components of ITTA. Here we conduct all experiments with Latte (\ie, I-Latte)}
    \resizebox{.9\textwidth}{!}{
    \begin{tblr}{ colspec            = {c | c | *{4}{c} | *{11}{c}},
        cell{1}{1-6}       = {r=2}{c}, cell{1}{10, 14}     = {r=2}{c},
        cell{1}{7, 11, 15} = {c=3}{c} }
    \hline
    \hline
    No. & Description & $\mathcal{L}^\mathrm{p}$ & $\mathcal{L}^\mathrm{ac}$ &
    $\mathcal{L}^\mathrm{in}$ & $\mathcal{L}^\mathrm{out}$ & U-to-S & & & &
    A-to-K & & & & S-to-K\\

    \hline

    & & & & & & 2D & 3D & xM & & 2D & 3D & xM & & 2D & 3D & xM\Tstrut\\

    \hline

    1 & Disable momentum grad & \cmark &  &  &  & 37.1 & 41.1 & 46.1 & & 49.3 &
    51.4 & 56.5 & & 33.5 & 40.1 & 42.6\Tstrut\\

    2 & Disable prompt refinement $\mathcal{L}^\mathrm{p}$ & & \cmark  & \cmark
    & \cmark & \textbf{37.2} & 41.1 & \textbf{46.3} & & 49.7 & \textbf{52.1} &
    57.3 & & \textbf{33.7} & 40.1 & 43.8\Tstrut\\

    3 & Disable anchor loss $\mathcal{L}^\mathrm{ac}$ & \cmark  &   & \cmark  &
    \cmark & 37.1 & 41.1 & 46.2 & & \textbf{49.8} & 51.9 & 57.0 & & 33.6 & 40.0
    & 43.4\Tstrut\\

    4 & Disable both $\mathcal{L}^\mathrm{in}$ and $\mathcal{L}^\mathrm{out}$ &
    \cmark  & \cmark &   & & 37.1 & \textbf{41.2} & 46.0 & & 49.7 & 51.8 & 56.9
    & & 33.6 & 40.1 & 43.4\Tstrut\\
    
    \hline

    5 & Disable decaying ($\gamma^\mathrm{mg}=1$) & \cmark & \cmark  & \cmark  &
    \cmark  
    & 37.1 & 41.1 & \textbf{46.3} & & 49.6 & 52.0 & 57.2 & & 33.5 & 40.3 &
    43.7\Tstrut\\

    6 & Disable disentangle in Equation~\ref{eq: ch6_grad_disentangle} & \cmark
    & \cmark & \cmark & \cmark & \textbf{37.2} & \textbf{41.2} & \textbf{46.3} &
    & \textbf{49.8} & 51.9 & 57.4 & & 33.4 & 40.4 & 43.6\Tstrut\\

    \hline
    
    - & I-Latte & \cmark & \cmark & \cmark & \cmark & \textbf{37.2} &
    \textbf{41.2} & \textbf{46.3} & & \textbf{49.8} & \textbf{52.1} &
    \textbf{57.5} & & \textbf{33.7} & \textbf{40.5} & \textbf{44.0}\Tstrut\\
    
    \hline
    \hline
    \end{tblr}
    }
    \label{tab: MG_Component Effectiveness}
\end{table*}

\begin{figure}[t]
    \centering
    \includegraphics[width=\linewidth]{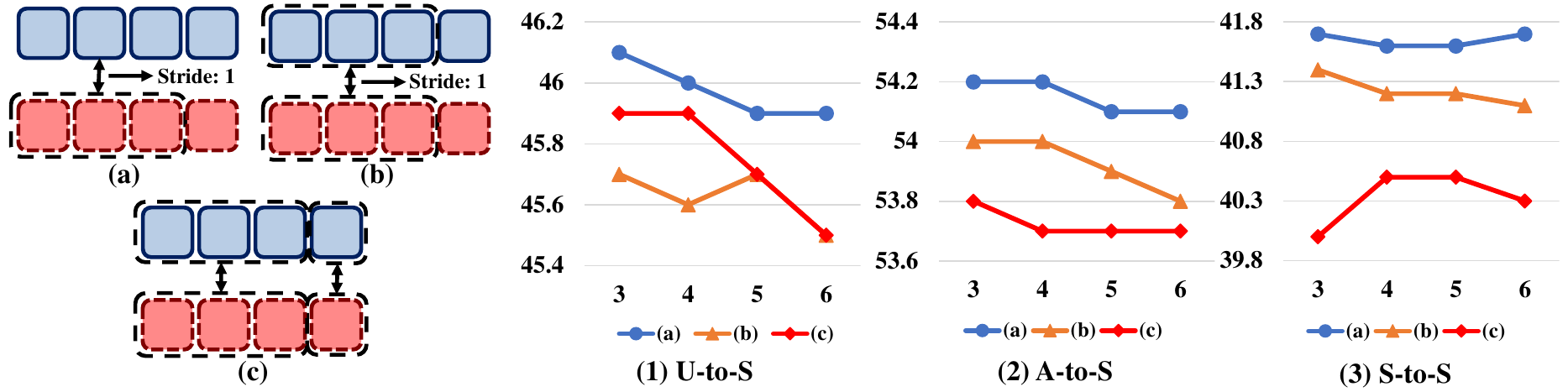}
    \caption{Ablation studies of different frame aggregation mechanisms. Besides (a) Latte's sliding-window aggregation, we test two different aggregation methods, including (b) replacing single frame student predictions with multiple frames in Equation~(\ref{Eq: ST Voxel Reference}) and (c) adding non-overlapping aggregation to (b). Results show that our sliding-window aggregation can better evaluate the local consistency, which leads to more consistent improvement.}
    \label{fig:Ablation Frame Aggregation}
\end{figure}

\begin{figure}[t]
    \centering
    \subfloat{\includegraphics[width=\linewidth]{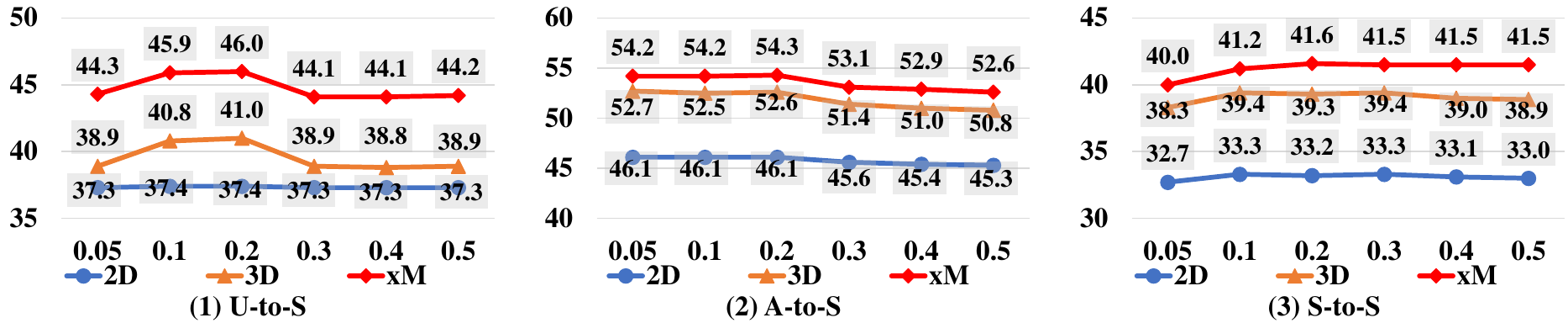}}\\
    \subfloat{\includegraphics[width=\linewidth]{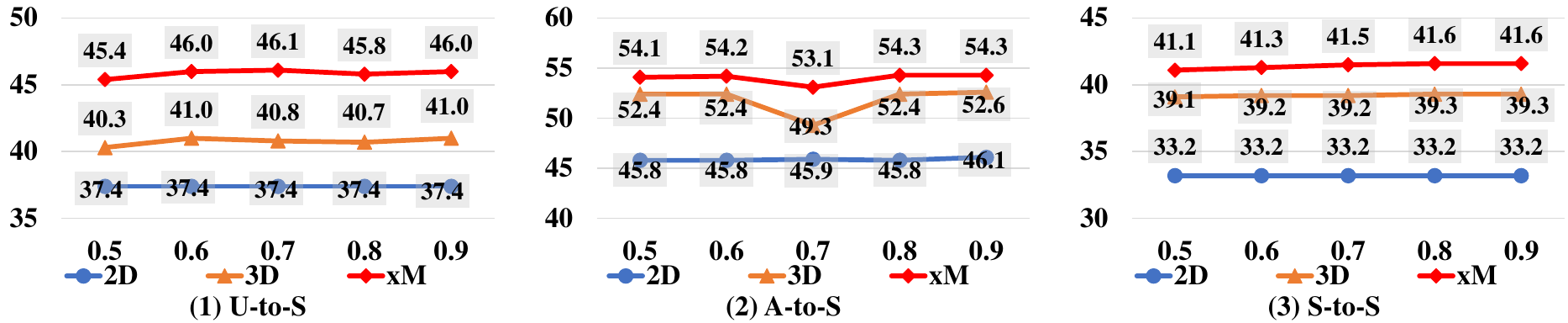}}
    \caption{Ablation studies of various ST voxel sizes (upper row) and filtering percentile $\alpha$ in Equation~(8) (lower row).}
    \label{fig:Ablation Voxel Size}
\end{figure}

\begin{table*}[t]
    \centering
    \setlength{\tabcolsep}{4pt}
    \caption{Performance (mIoU) of Latte++ using different window sizes.}
    \label{Tab: Latte++_window_size}
    \resizebox{.7\linewidth}{!}{
    \huge
    \begin{tblr}{
        colspec = {l| c | *{11}{c}},
        cell{1}{3, 7, 11} = {c=3}{c},
        cell{1}{1} = {r=2}{l},
        cell{1}{2, 6, 10} = {r=2}{c}
    }
    \hline
    \hline
    No. & Window Sizes & U-to-S & & & & A-to-K & & & & S-to-K & &\Tstrut\\
    \hline
    & & 2D & 3D & xM & & 2D & 3D & xM & & 2D & 3D & xM\Tstrut\\
    \hline
    1 & [1, 3] 
    & \textbf{37.6} & \textbf{41.0} & 46.2 &
    & \textbf{46.4} & 52.9 & 54.9 &
    & \textbf{33.5} & 39.7 & 42.3\Tstrut\\

    2 & [1, 3, 5] 
    & \textbf{37.6} & 40.9 & 46.2 &
    & \textbf{46.4} & 52.9 & 54.9 &
    & \textbf{33.5} & 39.8 & 42.4\\

    3 & [2, 3] 
    & \textbf{37.6} & 40.4 & 46.0 &
    & 46.2 & 52.7 & 54.8 &
    & 33.0 & 39.0 & 42.3\\

    4 & [3, 4] 
    & \textbf{37.6} & \textbf{41.0} & 46.1 &
    & \textbf{46.4} & 52.9 & 55.0 &
    & 33.2 & 39.8 & 42.3\\

    \hline

    5 & [3] 
    & 37.4 & \textbf{41.0} & 46.0 &
    & 46.1 & 52.6 & 54.3 &
    & 33.2 & 39.3 & 41.6\Tstrut\\
    
    6 & [3, 5] 
    & \textbf{37.6} & \textbf{41.0} & \textbf{46.3} &
    & \textbf{46.4} & \textbf{53.0} & \textbf{55.1} &
    & \textbf{33.5} & \textbf{39.9} & \textbf{42.5}\\

    \hline
    \hline
    
    \end{tblr}
}
\end{table*}

Besides optimization objects, we further testify two important components of Latte, including ST-entropy-based weighting in Equation~(\ref{Eq: ST xM Cons}) and Equation~(\ref{Eq: Pseudo-Labels}) as well as percentile filtering in Equation~(8). Specifically, disabling percentile filtering as in No. 7 leads to a relative accuracy drop of more than $0.5\%$ across all benchmarks, while utilizing point-wise entropy instead of our ST entropy as in No. 8 relatively degenerates the performance by $2.0\%$, which justifies that ST entropy better estimates the prediction reliability compared to the point-wise entropy. 

In terms of Latte++, we mainly validate the effectiveness of two components, including the initial weight $\tau_d^{\mathrm{b}}$ and multi-window weight $\tau_d^{\mathrm{m}}$ (Equation~\ref{Eq: multi-window init weights}). Specifically, we disable any of these adaptive weights by replacing them with 0.5. As shown in No. 10-12, disabling either $\tau_d^{\mathrm{b}}$ or $\tau_d^{\mathrm{m}}$ leads to a noticeable drop in both single-modal and cross-modal prediction accuracy, while those Latte++ variants can still surpass vanilla Latte consistently in most cases. However, when disabling $\tau_d^{\mathrm{m}}$ for both ST entropy and reference prediction in Equation~\ref{Eq: multi-window init weights} in No. 13, Latte++ achieves inferior performance compared to Latte, which justifies the effectiveness of assigning adaptive weights based on ST entropy of different time windows.

\subsubsection{Different frame aggregation mechanisms}
One of the core designs of Latte and Latte++ is the sliding-window frame aggregation, which is designed to capture temporally local correspondences and estimate their prediction consistency. To justify the aforementioned claim, we compare our sliding-window aggregation with two aggregation variants as shown in Fig.~\ref{fig:Ablation Frame Aggregation}. Specifically, variant (b) regards all student prediction frames rather than a single frame within the time window as the query in Equation~(\ref{Eq: ST Voxel Reference}), while variant (c) further utilizes a non-overlapping sliding strategy. According to Fig.~\ref{fig:Ablation Frame Aggregation}, our sliding-window aggregation surpasses the other two aggregation variants under all sizes of time windows across three benchmarks, which justifies the effectiveness of our method. The inferior performance of variants (b) and (c) could be due to two factors. Firstly, aggregating student predictions could be an inferior option since student predictions tend to be less stable, and aggregating multiple frames can therefore lead to noisy queries for cross-modal learning. Secondly, non-overlapping aggregation can result in a miss-checking area on the boundary of each time window, causing unrepresentative entropy estimation for some ST voxels. Another observation is that utilizing a smaller sliding window can usually yield better performance since temporal local consistency can be captured.

For Latte++, we test the performance of Latte++ when utilizing different combinations of window sizes as shown in Table~\ref{Tab: Latte++_window_size}. Among all alternatives, the combination of $[3,5]$ achieves SOTA performance among all evaluation metrics, and introducing a window size of $1$ brings a trivial difference (No. 6 vs. No. 2). Note that compared to vanilla Latte which utilizes a single pre-defined time window for consistency check, Latte++ with an arbitrary combination of different windows (No. 1-4, and No. 6) brings non-trivial and consistent improvement across three MM-TTA benchmarks. Especially on more challenging A-to-K and S-to-K, the relative gap on cross-modal predictions between Latte++ with arbitrary time window combinations and vanilla Latte consistently exceeds $1.0\%$, which validates the effectiveness of aggregating ST entropy under different time windows. 

\begin{figure}[t]
    \centering
    \subfloat{\includegraphics[width=\textwidth]{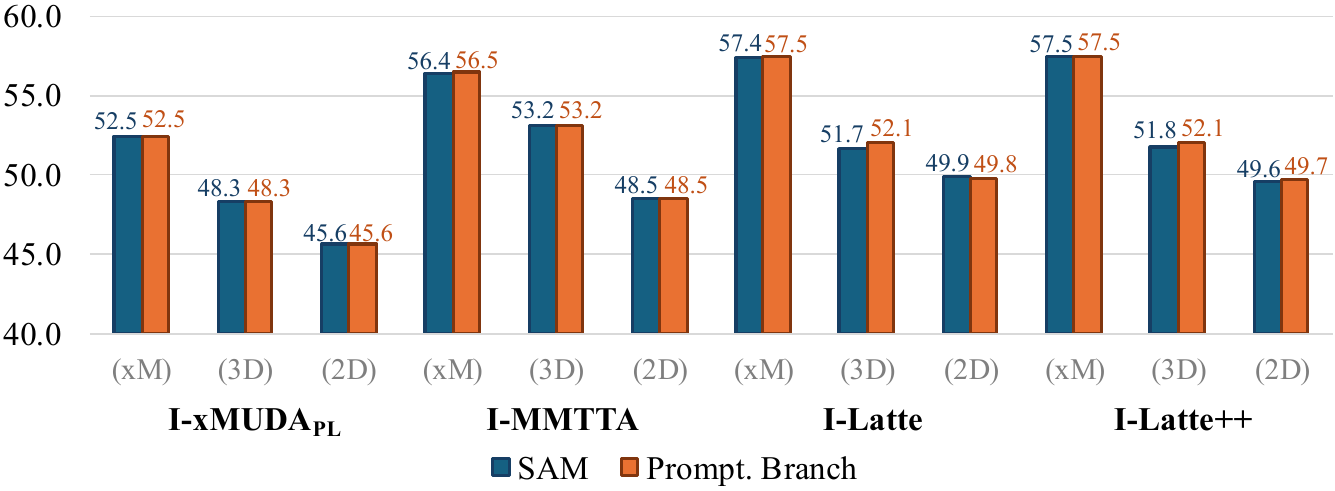}}\\
    \subfloat{\includegraphics[width=\textwidth]{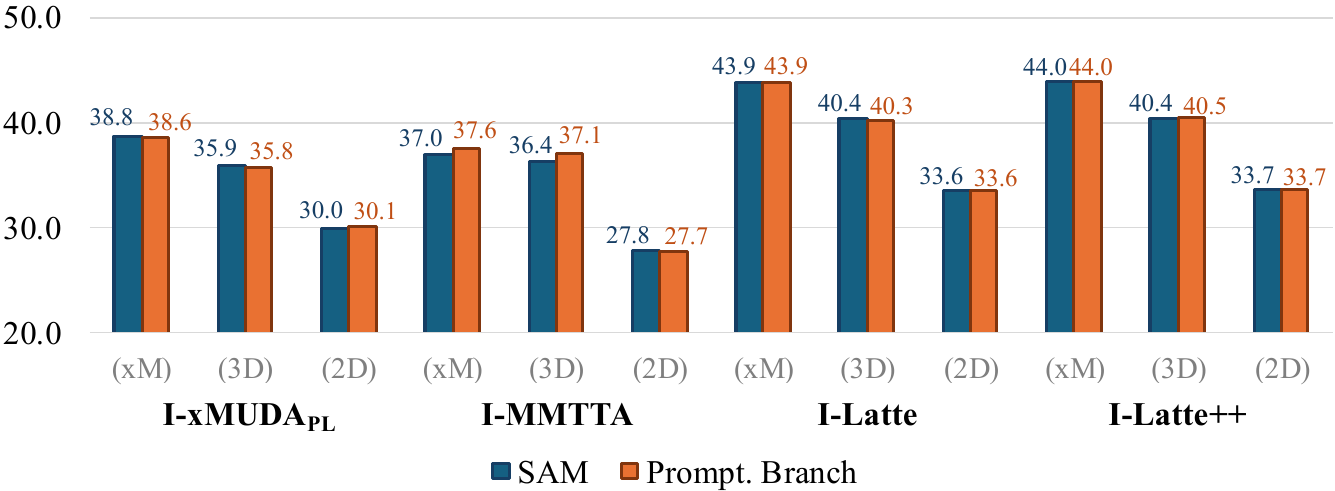}}
    \caption{Ablation studies on the effectiveness of our promptable branch. The upper row contains results from A-to-K, and the lower row contains results from S-to-K.}
    \label{fig: ch6_ab_sam_vs_branch}
\end{figure}

\begin{figure*}[t]
    \centering
    \subfloat{\includegraphics[width=.9\textwidth]{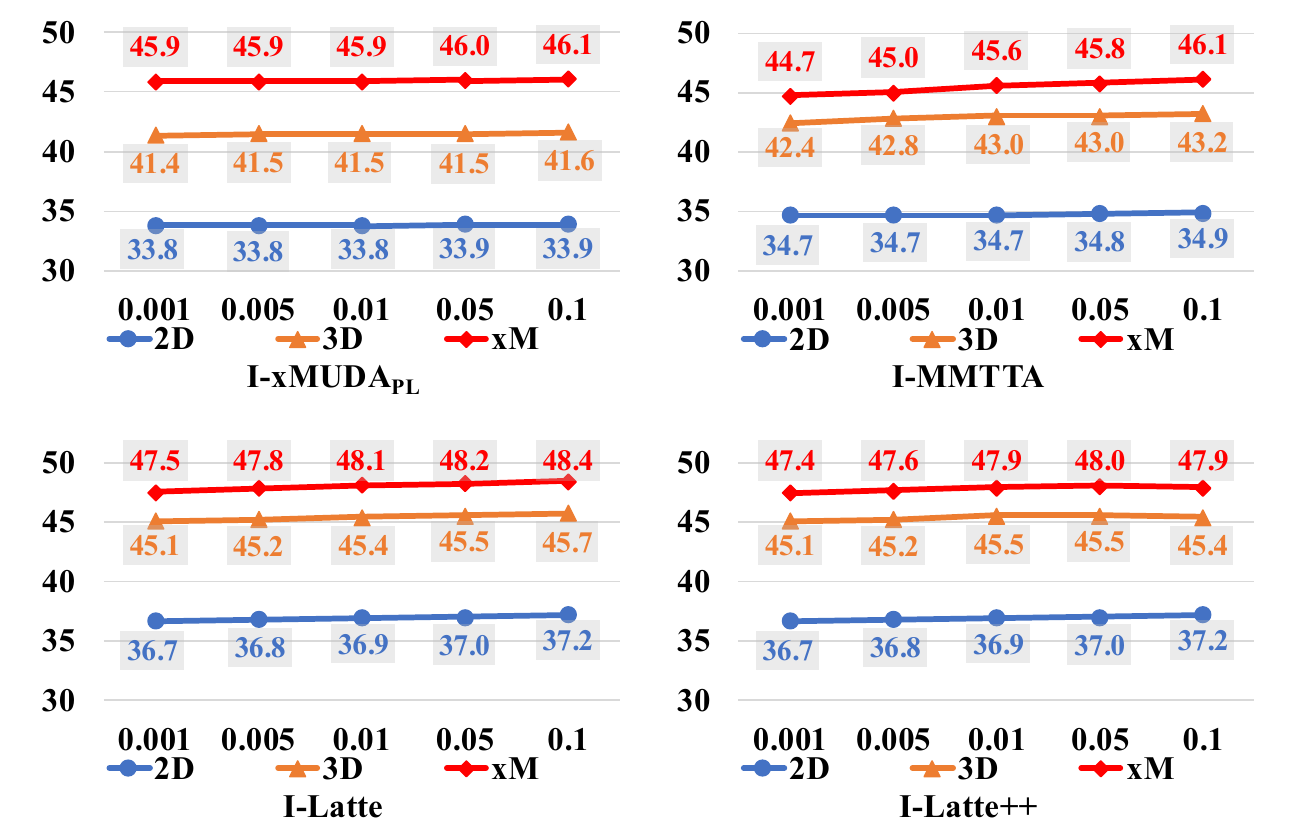}}\\
    \subfloat{\includegraphics[width=.9\textwidth]{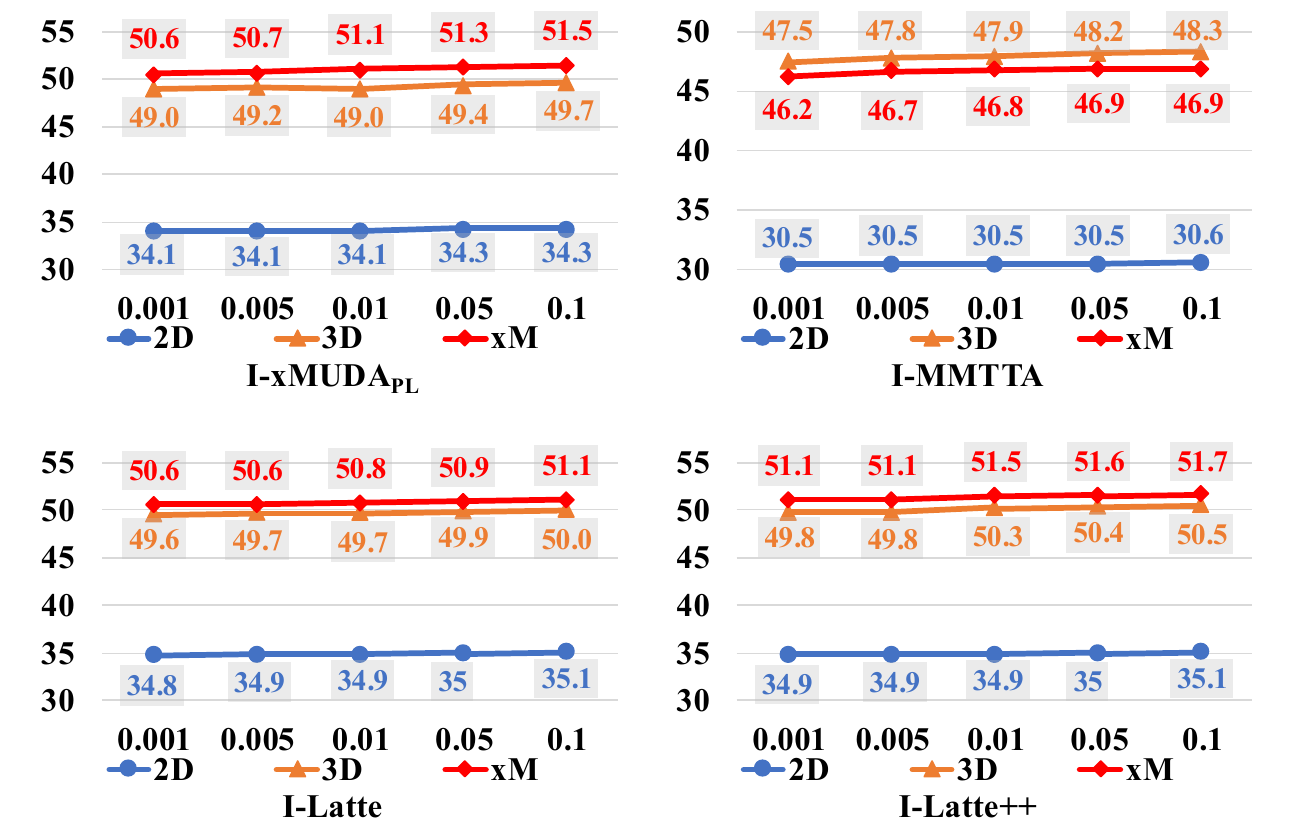}}
    \caption{Ablation studies on the effectiveness of interactive frequency. The upper two rows are results from K-to-S, and the lower two rows are from K-to-W.}
    \label{fig: ch6_ab_inte_freq}
\end{figure*}

\subsubsection{SAM vs. Prompptable Branch}
Our core argument is that a simple promptable branch is sufficient for receiving human feedback in our task. To verify this argument, we compare ITTA with its variant whose promptable branch is replaced by SAM as illustrated in Fig.~\ref{fig: ch6_ab_sam_vs_branch}. Surprisingly, the gap between using our lightweight branch and SAM is very minor, which is less than $\pm 0.3$ across all metrics and base TTA methods. This is mainly because we use the interactive segmentation in our pipeline for closed-set categories and specify the input to autonomous driving-related scenes. This illustrates that our design achieves satisfactory results on instance identification without costly computational overhead.

\begin{figure*}[t]
    \centering
    \includegraphics[width=\linewidth]{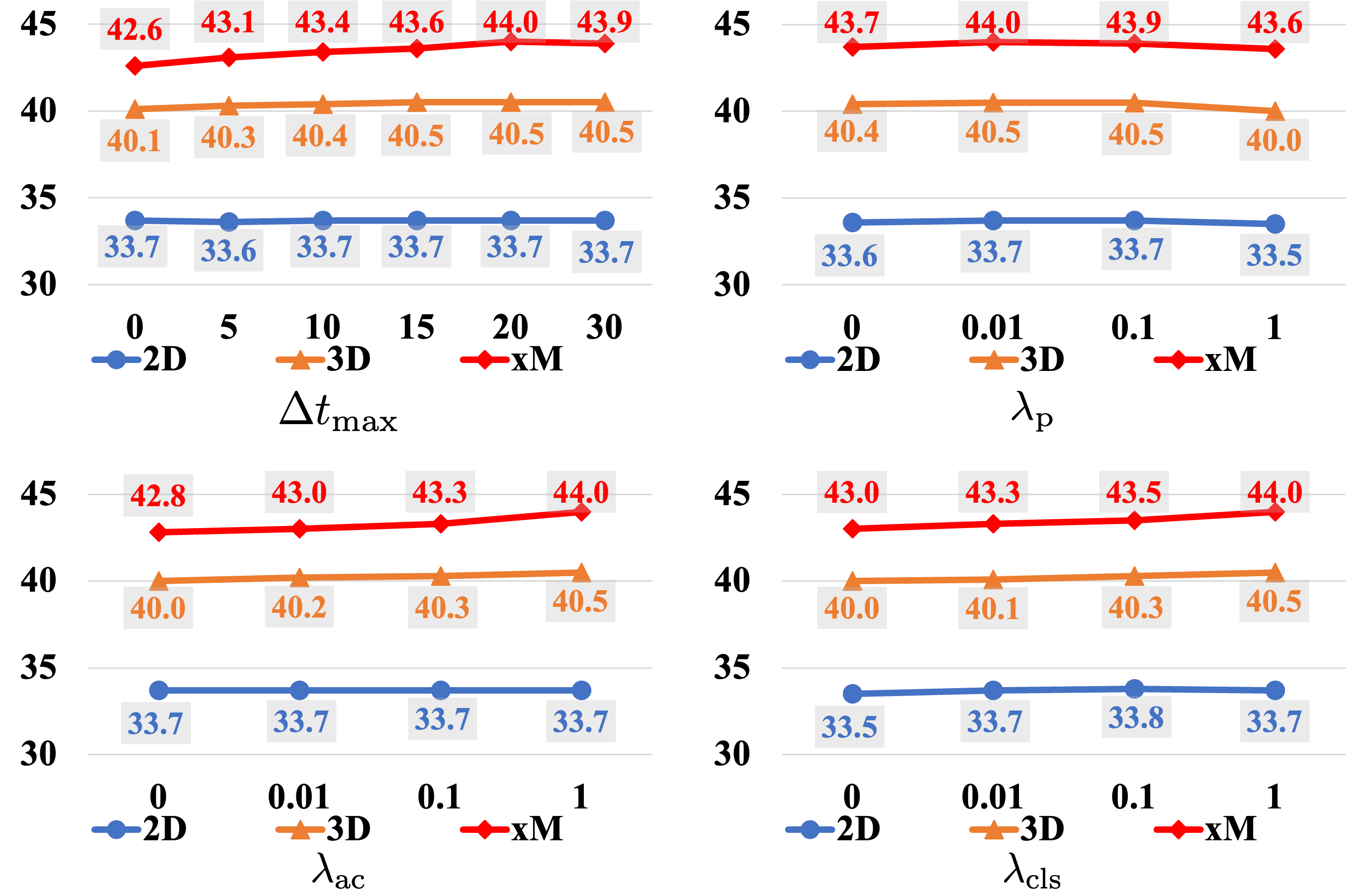}
    \caption{Parameter analysis with I-Latte on S-to-K.}
    \label{fig: ch6_param_analysys}
\end{figure*}

\subsubsection{Interactive Frequency}
The amount of interaction is a crucial factor in ITTA, and we therefore conduct a series of ablation experiments with varying interactive frequencies. As presented in Fig.~\ref{fig: ch6_ab_inte_freq}, all prediction accuracy increases as the number of interactions increases. The gap of cross-modal accuracy between $p_I=0.001$ and $p_I=0.1$ ranges from $0.2\%-1.4\%$ and $0.5\%-0.9\%$ on K-to-S and K-to-W, respectively, while noticeable improvements can be achieved compared to their based MM-TTA methods even when $p_I=0.001$. Practically, we find that $p_I=0.01$ is a good balance between efficiency and effectiveness, and therefore set it as the default value for ITTA.

\subsubsection{Parameter sensitivity analysis.}
To test whether Latte is sensitive to hyperparameter settings, we conduct a sensitivity analysis on Latte across three benchmarks. Here we illustrate the sensitivity analysis of two hyperparameters of Latte, including ST voxel size and filtering percentile $\alpha$ as shown in Fig.~\ref{fig:Ablation Voxel Size}. In terms of ST voxel sizes, one can tell from Fig.~\ref{fig:Ablation Voxel Size} that all benchmarks share the same optimal voxel size of 0.2, where performance drops as the voxel size becomes too large or too small. This is mainly because large voxels could cause boundary ambiguity, while small voxels contain insufficient representative points for evaluation. As for filtering percentile $\alpha$, a non-trivial improvement can be observed on U-to-S and A-to-K when $\alpha \geq 0.8$, while a lower $\alpha$ begins to discard more confident ST voxels, leading to a slight drop in accuracy. On S-to-K, Latte without filtering (\ie, No. 7 in Table~\ref{tab:Component Effectiveness}) performs slightly better on cross-modal predictions with a minor gap of $0.1\%$, while it can also be observed that setting $\alpha$ as 0.9 leads to an improvement of $0.3\%$ on both modality-specific predictions, which justifies the overall effectiveness of our filtering procedures. 

\begin{figure*}[t]
    \centering
    \includegraphics[width=.9\linewidth]{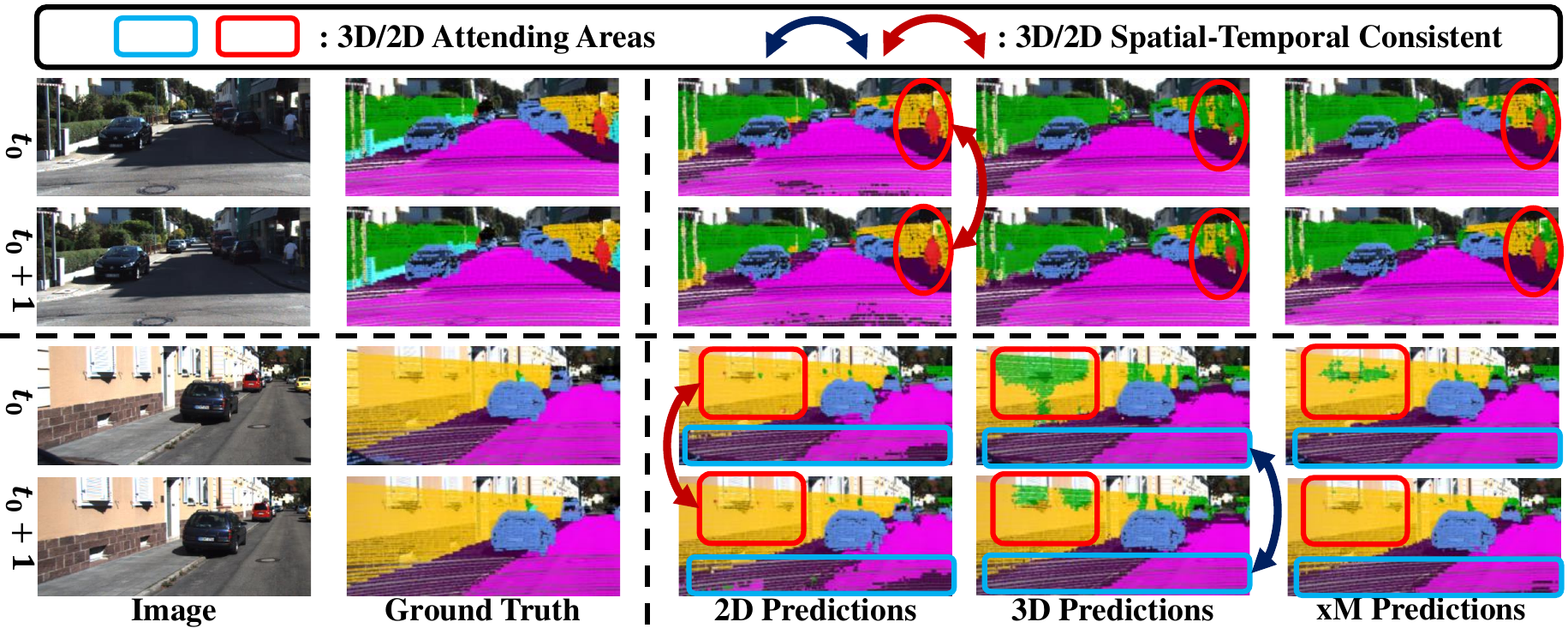}
    \caption{Qualitative results of Latte on S-to-K. Here we visualize modality-specific predictions and cross-modal predictions from Latte during two sets of consecutive frames.}
    \label{fig:Qualitative Comparison}
\end{figure*}

\begin{figure*}[t]
    \centering
    \subfloat{\includegraphics[width=.87\linewidth]{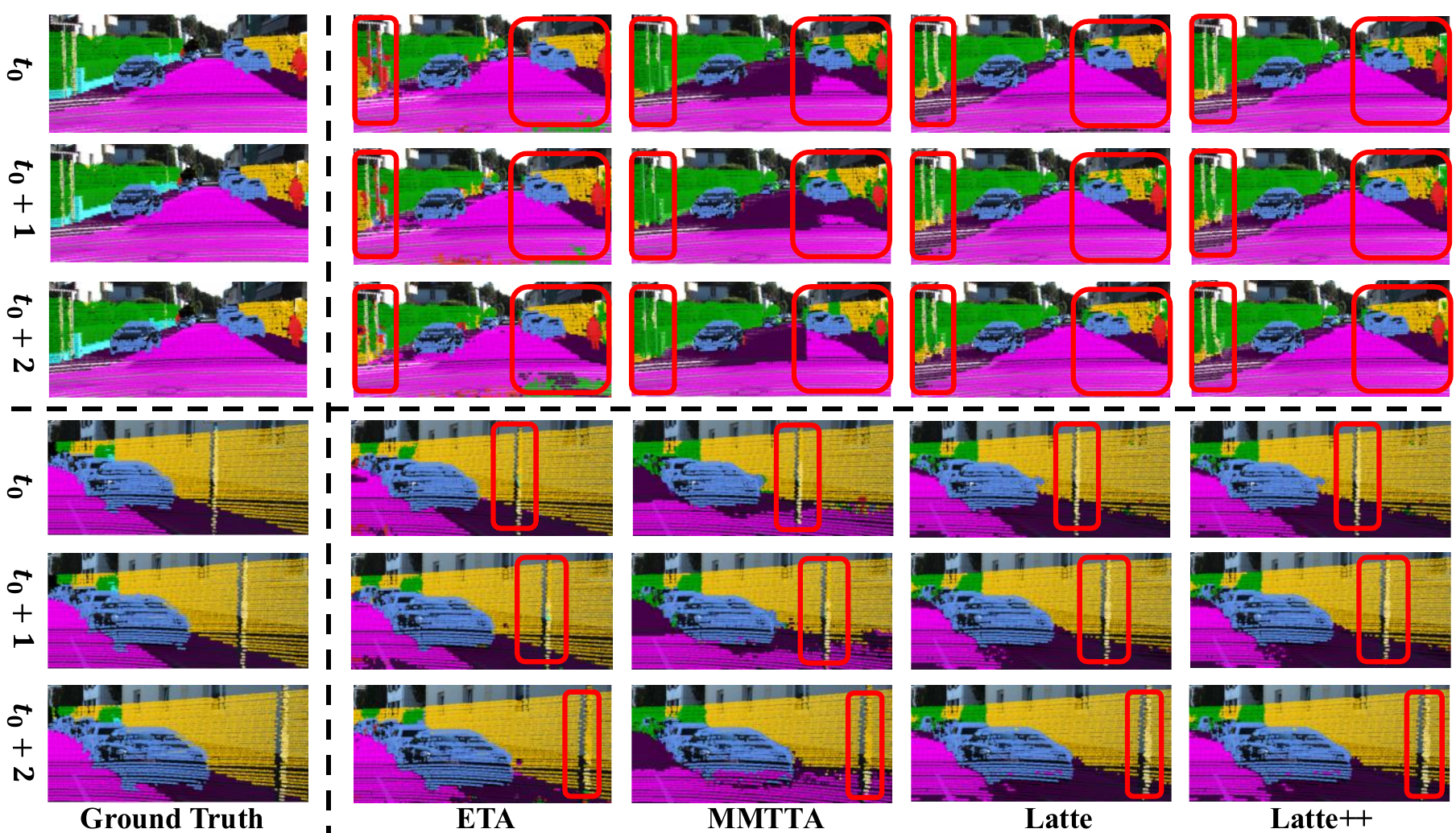}}\\
    \subfloat{\includegraphics[width=.87\linewidth]{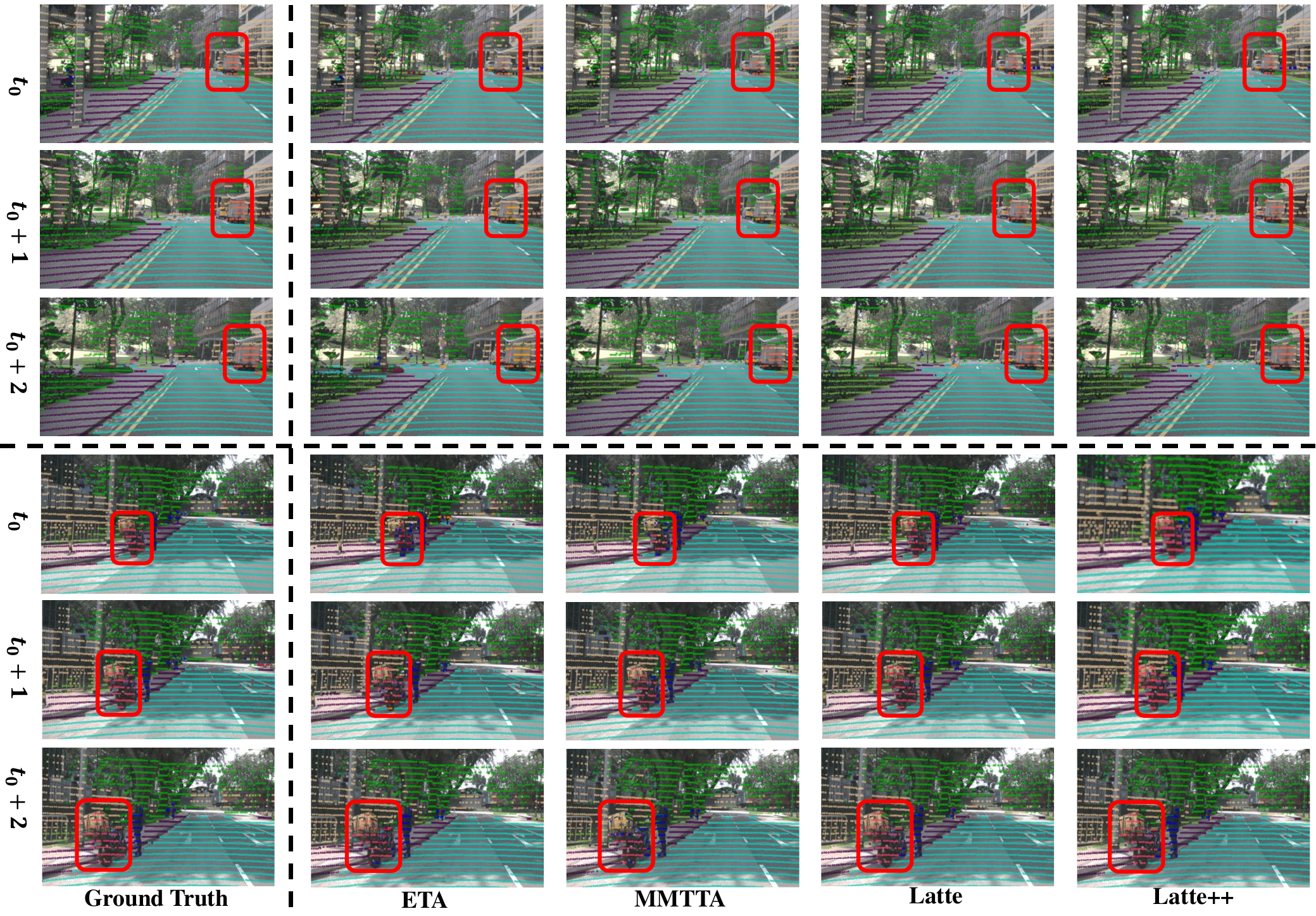}}
    \caption{Qualitative comparison with ETA~\citep{niu2022efficient} and MMTTA~\citep{shin2022mm} on S-to-K. \textcolor{red}{Red} boxes highlight the area where Latte++ produces more accurate predictions. Figure best viewed in color and zoomed in.}
    \label{fig:Supp Viz}
\end{figure*}

\begin{figure*}[t]
    \centering
    \includegraphics[width=\linewidth]{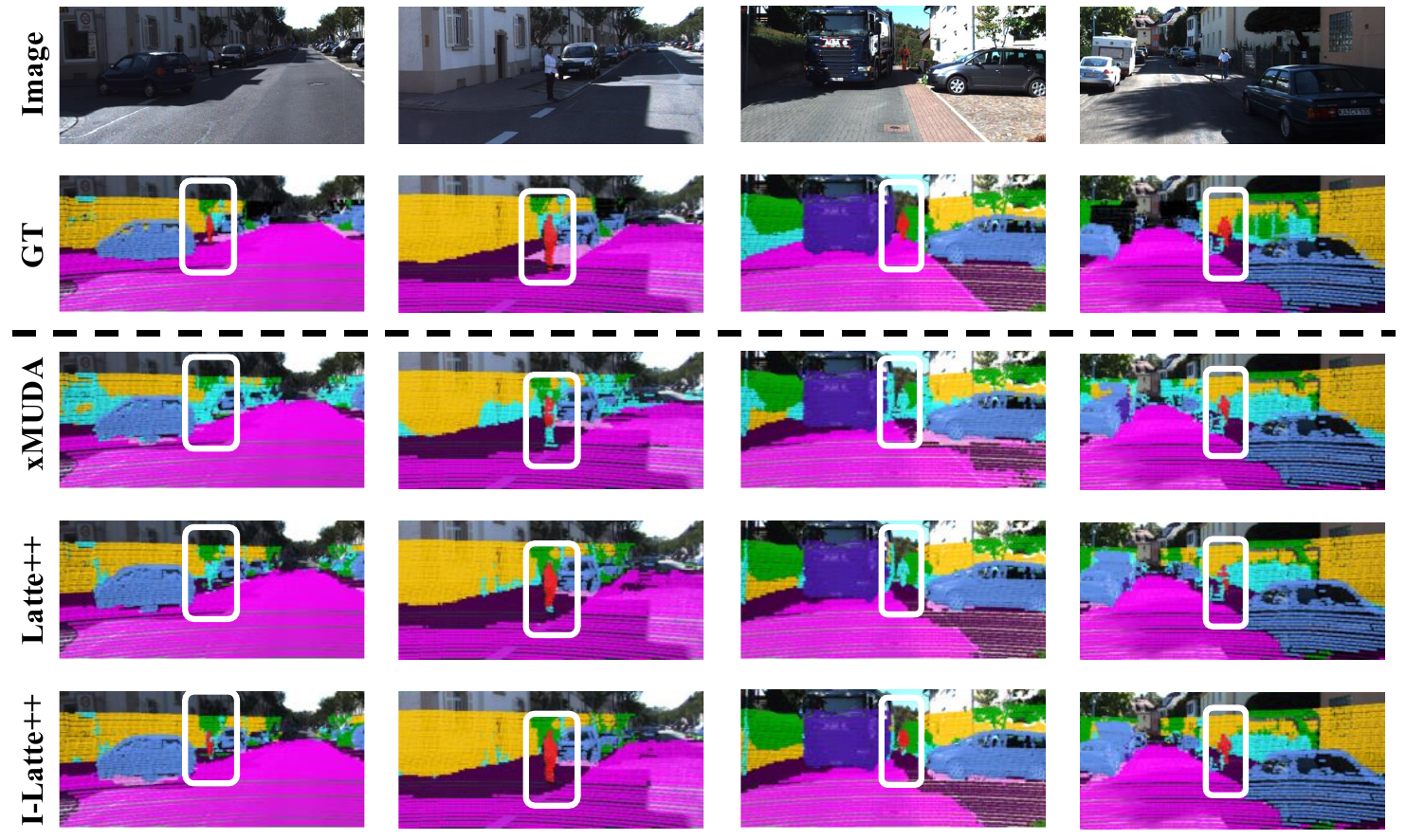}
    \caption{Qualitative comparison of $\mathrm{xMUDA_{PL}}$, Latte++ and I-Latte++. White boxes highlight the object of interest, where I-Latte++ can achieve more accurate segmentation results. Figure best viewed in color and zoomed in.}
    \label{fig:ITTA_viz}
\end{figure*}

In terms of ITTA, four hyperparameters are specifically investigated, including the maximum valid window $\Delta t_\mathrm{max}$ and the loss coefficients $\lambda_\mathrm{p}$, $\lambda_\mathrm{ac}$, and $\lambda_\mathrm{cls}$. As shown in Fig.~\ref{fig: ch6_param_analysys}, for $\Delta t_\mathrm{max}$, it can be observed that utilizing larger $\Delta t_\mathrm{max}$ gradually improves the performance, while it saturates at $\Delta t_\mathrm{max}=20$. As for the loss coefficient,  $\lambda_\mathrm{ac}$ and $\lambda_\mathrm{cls}$ are more significant coefficients compared to $\lambda_\mathrm{p}$ as they directly affect the regularization effect. These results motivate our default setting in ITTA, which we apply to all ITTA-related methods.

\subsubsection{Qualitative results}
Our main motivation is utilizing the prediction consistency of spatial-temporal correspondences to estimate modality reliability and achieve better cross-modal attending. As shown in Fig.~\ref{fig:Qualitative Comparison}, both Latte and Latte++ can effectively attend to the modality with more consistent predictions in spatial-temporal correspondences and therefore improve the cross-modal prediction consistency across time. For instance, Latte and Latte++ successfully attend to the stable 2D predictions of pedestrians in the upper set of consecutive frames, suppressing the spatial-temporal inconsistency from 3D predictions. Similar observations can also be found in the lower sample, where the more consistent 2D predictions of buildings in the red rectangle and 3D predictions of roads in the blue rectangle prevail.

To demonstrate the improvement brought by Latte++, we provide some additional qualitative comparisons between Latte++ and previous SOTA TTA (ETA~\cite{niu2022efficient}) and MM-TTA (MMTTA~\cite{shin2022mm}) methods on U-to-S and S-to-K. As shown in Fig.~\ref{fig:Supp Viz}, the cross-modal predictions from Latte and Latte++ are more accurate compared to ETA and MMTTA (\eg, more accurate pole recognition on the lower set of consecutive frames in Fig.~\ref{fig:Supp Viz}). Furthermore, the predictions from Latte and Latte++ are more consistent across time. For instance, the pedestrians and poles in the red rectangles in S-to-K (upper row), as well as the motorcycle and the truck in U-to-S (lower row), can be consistently recognized by Latte/Latte++, while both ETA and MMTTA suffer from the instability of single-frame predictions. This justifies the effectiveness of Latte++ and the improvement brought by our multi-frame aggregation strategy. In terms of ITTA, we also qualitatively compared I-Latte++ with the vanilla Latte++ in Fig.~\ref{fig:ITTA_viz}, where it can be observed that I-Latte++ can achieve much more accurate segmentation results on objects of interest (\ie, \texttt{pedestrian}). These qualitative comparison justifies the effectiveness of both Latte++ and ITTA.

\section{Conclusion}
In this paper, we propose a MM-TTA method, named Latte++, which leverages multiple time windows for a more accurate estimation of voxel-wise prediction reliability. Considering that existing MM-TTA methods struggle to rectify incorrect predictions over time consistently, we propose ITTA, a flexible add-on that enables existing MM-TTA methods to be aided by effortless human feedback. Extensive experimental results across five benchmarks are provided, justifying the effectiveness of both Latte++ and ITTA across various benchmarks. We also demonstrate that ITTA can significantly improve the existing MM-TTA methods by incorporating scarce human feedback. This observation highlights the potential of integrating human interaction with online adaptation methods, an area worth exploring to extend this paradigm to various tasks.

\section*{Declarations}
\begin{itemize}
    \item Funding: This research is supported by the National Research Foundation, Singapore, under the NRF Medium Sized Centre scheme (CARTIN). Any opinions, findings, and conclusions or recommendations expressed in this material are those of the author(s) and do not reflect the views of the National Research Foundation, Singapore.
    \item Competing interests: The authors have no competing interests to declare that are relevant to the content of this article.
    \item Data availability: The NuScenes dataset is available at \url{https://www.nuscenes.org/nuscenes}. The A2D2 dataset is available at \url{https://www.a2d2.audi/en/download/}. The SemanticKITTI dataset is available at \url{https://semantic-kitti.org/}. The Synthia dataset is available at \url{https://synthia-dataset.net/}. The Waymo dataset is available at \url{https://waymo.com/open/download/}. Benchmark details are available at \url{https://sites.google.com/view/mmcotta?pli=1} and \url{https://github.com/AronCao49/Latte}.
    \item Code availability: Code will be available at \url{https://github.com/AronCao49/Latte-plusplus} upon acceptance.
\end{itemize}

\begin{appendices}

\section{Structure of Promptable Branch}\label{appendix: prompt_branch}
As mentioned in Section~\ref{subsec:ch6_warmup}, the promptable branch is structured by inserting a light-weight bottleneck to map the full segmentation features to the promptable ones. More specifically, we treat the image features to be fed to the final classifier of SegFormer~\citep{xie2021segformer} as the input of our bottleneck, whose structure is detailed as in Table~\ref{tab: promptable structure}. The bottleneck is mainly constructed by two 2D convolutional layers, compressing the feature channels of image features from SegFormer from $4\times C_F$ to $C_F$ ($C_F=256$ in our adopted SegFormer-B1) while maintaining the spatial resolution. When interacting with human prompts in the mask decoder of SAM, the promptable features are bilinearly interpolated to match the desired spatial resolution of the decoder.

\begin{table*}[t]
    \centering
    \huge
    \setlength{\tabcolsep}{8pt}
    \caption{Structure of Promptable Bottleneck. $C_{in}$ and $C_{out}$ represent the input and output channel number of each layer, while $K$, $S$, and $P$ denote the kernel size, stride, and padding of each 2D convolutional layer, respectively.}
    \resizebox{.9\textwidth}{!}{
    \begin{tblr}{ 
        colspec            = {c | c | c | c | c},
        cell{1}{1}         = {r=6}{c}
    }
    \hline
    \hline
    \rotatebox[origin=c]{90}{\makecell{Promptable \\ Bottleneck}} & Input Size & Output Size & Layer Name & Hyper Param.\\
    \hline
    & $(4\times C_F) \times H_F \times W_F$ & $C_F \times H_F \times W_F$ & 
    2D Conv 1 & $C_{in}=4\times C_F, C_{out}=C_F, K=1, S=1$ \\
    
    & $C_F \times H_F \times W_F$ & $C_F \times H_F \times W_F$ & 
    LayerNorm & $C_{in}=C_F, C_{out}=C_F$ \\

    & $C_F \times H_F \times W_F$ & $C_F \times H_F \times W_F$ & 
    GELU & / \\

    & $C_F \times H_F \times W_F$ & $C_F \times H_F \times W_F$ & 
    2D Conv 2 & $C_{in}=C_F, C_{out}=C_F, K=3, S=1, P=1$ \\

    & $C_F \times H_F \times W_F$ & $C_F \times H_F \times W_F$ & 
    LayerNorm & $C_{in}=C_F, C_{out}=C_F$ \\
    
    \hline
    \hline
    \end{tblr}
    }
    \label{tab: promptable structure}
\end{table*}

\begin{table*}[ht]
    \centering
    \caption{Class mapping of the benchmark SemanticKITTI-to-Synthia and SemanticKITTI-to-Waymo.}
    \resizebox{.9\linewidth}{!}{
    \begin{tabular}{m{6em} | m{18em} | m{18em}}
    \hline
    \hline
    S-to-S Class & SemanticKITTI classes & Synthia classes\\
    \hline
    car & car, moving-car, truck, moving-truck & car\\
    bike & bicycle, motorcycle, bicyclist, motorcyclist, moving-bicyclist, moving-motorcyclist & bicycle \\
    person & person, moving-person & pedestrian \\
    road & road, lane-marking, parking &  road, lanemarking \\
    sidewalk & sidewalk & sidewalk \\
    building & building & building \\
    nature & vegetation, trunk, terrain & vegetation \\
    poles & pole & pole \\
    fence & fence & fence \\
    traffic-sign & traffic-sign & traffic-sign \\
    other-objects & other-object & traffic-light \\
    \hline
    S-to-W Class & SemanticKITTI classes & Waymo classes\\
    \hline
    car & car, moving-car, truck, moving-truck & car, truck, other-vehicle, bus\\
    bike & bicycle, motorcycle, bicyclist, motorcyclist, moving-bicyclist, moving-motorcyclist & bicycle, motorcycle, bicyclist, motorcyclist \\
    person & person, moving-person & pedestrian \\
    road & road, lane-marking, parking &  road, lane-marker \\
    sidewalk & sidewalk & sidewalk, curb, walkable \\
    building & building & building \\
    nature & vegetation, terrain & vegetation \\
    poles & pole & pole \\
    trunk & trunk & tree-trunk \\
    traffic-sign & traffic-sign & sign \\
    other-objects & other-object & other-ground, construction-cone, traffic-light \\
    \hline
    \hline
    \end{tabular}
    }
    \vspace{10pt}
    \label{Table:mmctta_mapping}
\end{table*}

\section{MM-CTTA Benchmarks}
Both MM-CTTA benchmarks proposed in CoMAC~\citep{Cao_2023_ICCV}, including SemanticKITTI-to-Synthia (K-to-S) and SemanticKITTI-to-Waymo (K-to-W), are introduced in this work for the main result comparison. We strictly follow the identical preprocessing procedures and split arrangement in CoMAC~\citep{Cao_2023_ICCV}, leading to the class mapping as shown in Table~\ref{Table:mmctta_mapping}.

\end{appendices}

\bibliography{sn-bibliography}

\end{document}